\documentclass[final]{cvpr}

\usepackage{times}
\usepackage{epsfig}
\usepackage{graphicx}
\usepackage{amsmath}
\usepackage{amssymb}

\usepackage{booktabs}
\usepackage{enumitem}
\usepackage[pagebackref=false,breaklinks=true,colorlinks,bookmarks=false]{hyperref}

\def\confYear{CVPR 2021}

\begin{document}

\title{Image Inpainting with External-internal Learning and Monochromic Bottleneck}
\renewcommand{\thefootnote}{\fnsymbol{footnote}}

\author{Tengfei Wang\footnotemark[1] \qquad     Hao Ouyang\footnotemark[1] \qquad   Qifeng Chen\\

The Hong Kong University of Science and Technology\\

}

\maketitle
\pagestyle{empty}
\thispagestyle{empty}
\footnotetext[1]{Equal contribution}

\begin{abstract}
Although recent inpainting approaches have demonstrated significant improvement with deep neural networks,  they still suffer from artifacts such as blunt structures and abrupt colors when filling in the missing regions. To address these issues, we propose an external-internal inpainting scheme with a monochromic bottleneck that helps image inpainting models remove these artifacts.  In the external learning stage, we reconstruct missing structures and details in the monochromic space to reduce the learning dimension.  In the internal learning stage, we propose a novel internal color propagation method with progressive learning strategies for consistent color restoration. Extensive experiments demonstrate that our proposed scheme helps image inpainting models produce more structure-preserved and visually compelling results. Our source code is available at \url{https://github.com/Tengfei-Wang/external-internal-inpainting}.
\end{abstract}

\section{Introduction}

Image inpainting is a task that aims to complete the missing regions of an image with visually realistic and semantically consistent content. Image inpainting can benefit general users in various practical applications, including unwanted object removal from an image, face defect removal, and image editing. While we have witnessed significant progress in image inpainting, inpainting models still suffer from abrupt color artifacts, especially when the missing regions are large. This work will analyze the weaknesses of state-of-the-art inpainting approaches and present a novel framework to improve existing inpainting methods.
 
\begin{figure*}[t]
    \centering
    \begin{tabular}{@{}c@{\hspace{0.6mm}}c@{\hspace{0.5mm}}c@{\hspace{0.5mm}}c@{\hspace{0.5mm}}c@{\hspace{0.5mm}}c@{\hspace{0.5mm}}c@{}}
    \includegraphics[width=0.24\linewidth]{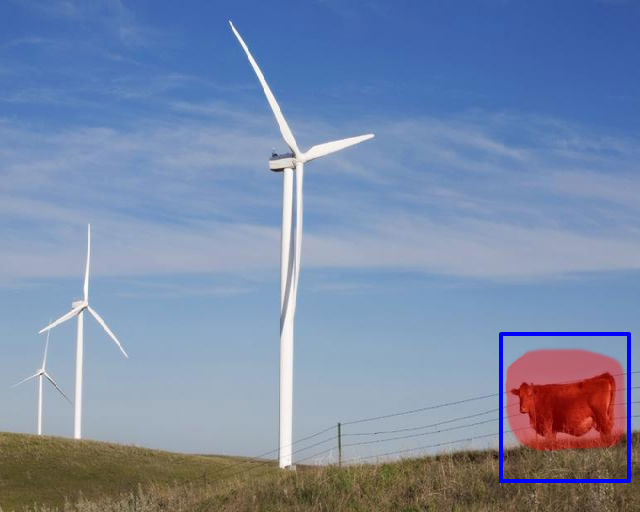}   &   
    \includegraphics[width=0.12\linewidth]{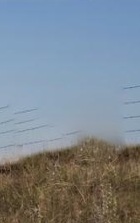}  &    
        \includegraphics[width=0.12\linewidth]{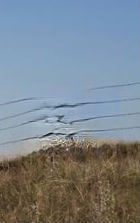}  &  
    \includegraphics[width=0.12\linewidth]{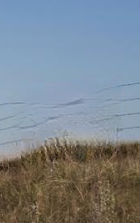}&
    \includegraphics[width=0.12\linewidth]{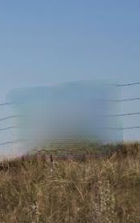}&
    \includegraphics[width=0.12\linewidth]{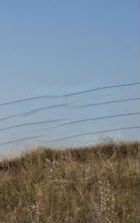}&
    \includegraphics[width=0.12\linewidth]{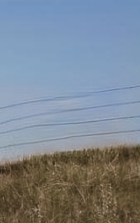}\\   
    \includegraphics[width=0.24\linewidth]{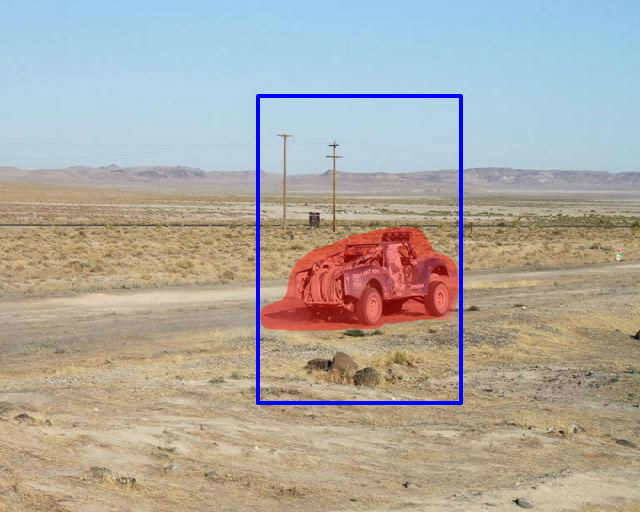}   & 
    \includegraphics[width=0.12\linewidth]{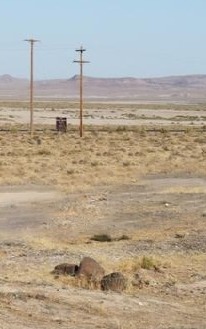}  &  
        \includegraphics[width=0.12\linewidth]{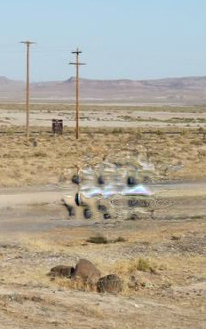}  &  
 \includegraphics[width=0.12\linewidth]{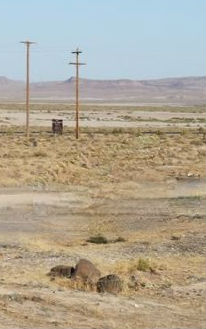}&     
    \includegraphics[width=0.12\linewidth]{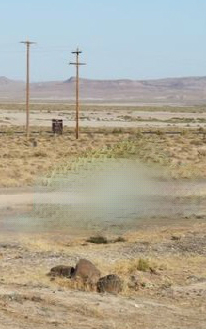}&
    \includegraphics[width=0.12\linewidth]{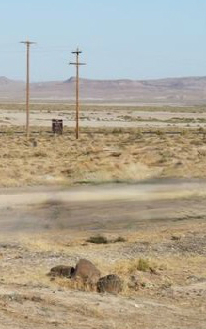}&
    \includegraphics[width=0.12\linewidth]{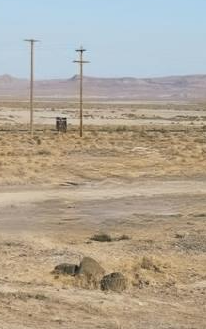}\\     
    \scriptsize{Input}&\scriptsize{PatchMatch~\cite{barnes2009patchmatch}}&\scriptsize{GMCNN~\cite{wang2018image}}&\scriptsize{PartialConv~\cite{liu2018image}}&\scriptsize{EdgeConnect~\cite{nazeri2019edgeconnect}}&\scriptsize{GatedConv~\cite{ yu2019free}}&\scriptsize{Ours}
    \end{tabular}
    \caption{Image inpainting results by traditional and deep learning methods.    Zoom in for details.}
    \label{fig:teaser}
\end{figure*}

State-of-the-art inpainting methods roughly fall into two categories of patch matching by iteratively nearest-neighbor search and deep learning models, with different pros and cons. PatchMatch~\cite{barnes2009patchmatch} is a learning-free method that only utilizes internal statistics of a single image. As shown in Fig.~\ref{fig:teaser}, it generates smooth patterns and colors that are consistent with the non-missing region, but it fails to fill in semantic-aware content. The deep learning based inpainting approaches can learn semantic-aware models by training on large-scale datasets. These approaches have explored coarse-to-fine inpainting models in different fashions. They may first generate edges~\cite{nazeri2019edgeconnect,li2019progressive}, structural information~\cite{ren2019structureflow}, segmentation maps~\cite{song2018spg} or blurry images~\cite{yu2018generative,yu2019free,yi2020contextual}, and then use these intermediate outputs as guidance for filling in details. However, their results still suffer from color and texture artifacts. One of the most common artifacts observed is color bleeding, as shown in Fig.~\ref{fig:teaser}. These methods trained on a large-scale dataset tend to introduce inconsistent colors that do not conform to the color distribution of the test image. On the other hand, we observe that color bleeding artifacts seldom appear in the internal methods.

Based on the observations above, we propose a  robust inpainting method by combining the best of both worlds. We adopt a novel external-internal inpainting scheme with a monochromic bottleneck: first completing the monochromic image via learning externally from large-scale datasets and then colorizing the completed monochrome by learning internally on the single test image. Our proposed method is orthogonal to early inpainting approaches and thus can be easily applied to improve previous learning-based inpainting models for a higher-quality generation. In the external learning stage, by changing the output of the reconstruction network from polychromatic images to monochromic images, we reduce the dimension of the optimization space from $\mathbb{R}
^3$ to  $\mathbb{R}
$, leading to more structure-preserving reconstruction~(Section \ref{4.3}). Models trained in this way also show stronger generalization ability on cross-dataset evaluation. In the colorization stage, motivated by the recent advancement in deep internal learning, we propose a novel internal color propagation approach guided by the completed monochromic bottleneck. However, similar monochromic values can map to different polychromic outputs even in a single image. We, therefore, adopt a progressive restoration strategy for combining both local and global color statistics. Our external-internal learning scheme not only facilitates structure reconstruction but also ensures color consistency. By focusing on the internal color distribution of a single image, we can eliminate abrupt colors and produce a visually pleasing image~(Section~\ref{3.1}). 

 We conduct extensive experiments to evaluate the performance of our method on four public datasets Places2~\cite{zhou2017places}, Paris StreetView~\cite{pathak2016context}, CelebA-HQ~\cite{karras2017progressive} and DTD~\cite{cimpoi14describing}.
We apply our method to different baseline networks (GatedConv~\cite{yu2019free}, EdgeConnect~\cite{nazeri2019edgeconnect}, HiFill~\cite{yi2020contextual} and GMCNN~\cite{wang2018image}), and observe meaningful improvement in terms of structure preservation and color harmonization. Furthermore, we perform model analysis and ablation studies to verify our hypothesis and modifications.  The main contributions of our paper can be summarized as:

\begin{itemize}[noitemsep,topsep=0pt]

  \item To the best of our knowledge, we are the first to introduce an external-internal learning method to deep image inpainting. It learns semantic knowledge externally by training on large datasets while fully utilizes internal statistics of the single test image.
  \item We design a progressive internal color restoration network that achieves outstanding colorization performance in our case. 
  \item We generalize our proposed method to several deep inpainting models and observe clear improvement in terms of visual quality and model generalization ability on multiple datasets.  
\end{itemize}

\begin{figure*}[t]
    \centering
    \includegraphics[width=\linewidth]{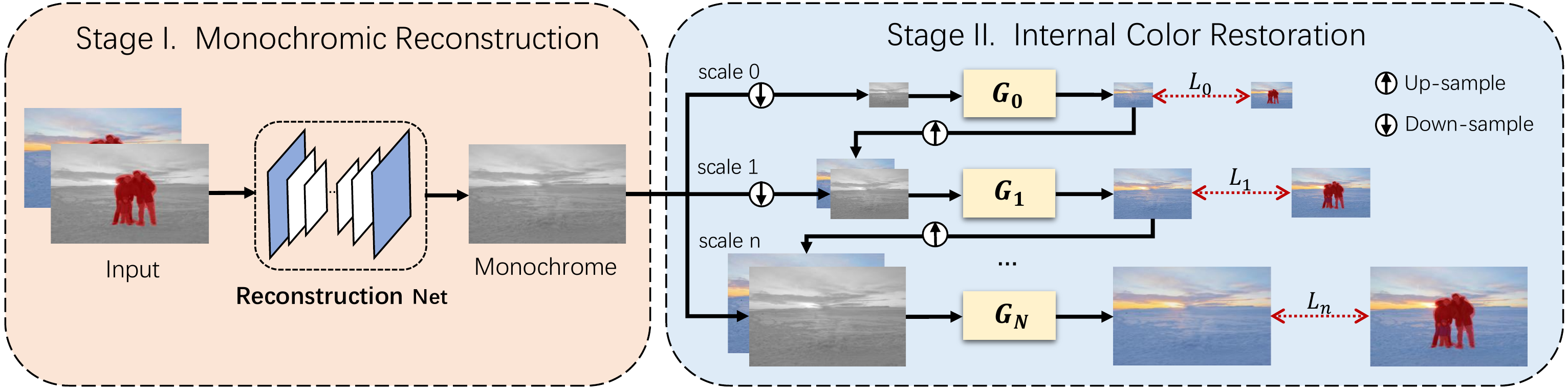}
    \caption{Overview of our external-internal inpainting method. It externally learns to reconstruct  structures in the monochromic space via training on large datasets, while implicitly propagates colors within the single test image via internal learning. The colorization loss  $L_n$ is only calculated on the  unmasked regions.}
    \label{fig:architecture}
\end{figure*}

\section{Related Work}
\subsection{Image Inpainting}
Traditional learning-free image inpainting methods can be roughly divided into two categories: diffusion-based and patch-based methods. Diffusion-based methods~\cite{ashikhmin2001synthesizing,drori2003fragment,ballester2001filling,esedoglu2002digital} propagate neighboring information using techniques such as isophote direction field. These methods perform well on texture data or images with narrow holes while they will fail when the masked region is large or contains meaningful structures. Patch-based methods such as PatchMatch~\cite{barnes2009patchmatch} fill in the missing region by searching the patches outside the hole with a fast nearest neighbor algorithm. However, the pattern in the missing region cannot always be found in the image, and also repetitive patterns tend to appear in the reconstructed image. These methods utilize only the internal information that achieves color consistency but fails in filling in semantic-aware contents. 

The recent development of deep learning has greatly improved the performance of image inpainting, especially in image categories like faces and complex natural scenes. The inpainting model benefits from learning and understanding semantic meanings from large-scale datasets~\cite{deng2009imagenet}. Pathak et al.~\cite{pathak2016context} first proposed context encoders that utilized an encoder-decoder network to extract features and reconstruct the outputs. Iizuka et al.~\cite{iizuka2017globally} used both global and local discriminator, and Yu et al.~\cite{yu2018generative} proposed the contextual attention for retrieving remote features and achieving global coherency. Liu et al.~\cite{liu2018image} applied the partial convolution, and  Yu et al.~\cite{yu2019free}  proposed the gated convolution to overcome the weakness of the vanilla convolution.  Yi et al.~\cite{yi2020contextual}  proposed the contextual residual aggregation module, and Zeng et al.~\cite{zeng2020high} adopted a guided upsampling network for high-resolution image inpainting.

Most recent methods  first predict coarse structure such as edges~\cite{nazeri2019edgeconnect,li2019progressive}, foreground contours~\cite{xiong2019foreground}, structure shape~\cite{ren2019structureflow} and semantic maps~\cite{song2018spg}, and then provide additional prior for guiding the completion of images. These methods show that conducting inpainting in a spatially coarse-to-fine way will benefit the training process. Our method also adopts a similar idea while completing images not only spatially but also in a channel-wise coarse-to-fine way via external-internal learning.

\subsection{Guided Colorization}
User-guided colorization methods focus on local input, such as user strokes or color points. The color is propagated using low-level similarity metrics based on luminance~\cite{levin2004colorization,huang2005adaptive}, textures~\cite{qu2006manga}, and intrinsic distance~\cite{yatziv2006fast}. In addition to local hints, Li et al.~\cite{li2015image} utilized color theme and Chang et al.~\cite{chang2015palette} used color palette for expressing global color control. Zhang et al.~\cite{zhang2017real}  also combined low-level cues along with high-level semantic similarities. Example-based approaches transferred color from a single or multiple reference images to the target image. These approaches, no matter using which techniques (color transfer~\cite{he2017neural,reinhard2001color} or image analogies~\cite{hertzmann2001image}), all focused on finding the correct correspondence between the reference and target images. In our internal colorizarion, we use the monochromic output from the first stage as a conditional input and thus propagate the internal color information from the non-missing region to the missing region. Different from the user-guidance and example-guidance, the guidance in our case is not only extraordinarily dense but also has accurate one-to-one correspondences, which can provide sufficient information both locally and globally. We show in the paper that existing guided colorization methods cannot fully utilize the reliable color information in the non-missing regions.

\subsection{Deep Internal Learning}
Training a deep convolutional neural network on only a single image has shown effectiveness in various image generation tasks such as super-resolution, texture synthesis, and so on~\cite{shocher2018zero,chan2019everybody,zhou2018non,shocher2019ingan,shaham2019singan}.   Ulyanov et al.~\cite{ulyanov2018deep} were the first to utilize deep model as a prior to train image inpainting. They trained a deep model on the non-missing region of a single image from random Gaussian noise and try to propagate similar content information to the missing regions. However, their model fails to generate realistic details in the inpainted area. Shocher et al.~\cite{shocher2019ingan} introduced the InternalGAN for conditional image generation. However, since our ground truth image is only partially available (the non-missing part), it is difficult to apply the adversarial training. Considering our case, we carefully design a progressive deep network for internal image colorization.

\section{Method}

In this section, we first analyze the drawbacks of state-of-the-art image inpainting methods and the motivation of our external-internal learning scheme.  We then present the details of the two stages: external monochromic reconstruction trained on large-scale datasets for generating semantically correct content, and internal color restoration on a single image for propagating the color from non-missing parts to missing regions. The overview architecture is shown in Fig.~\ref{fig:architecture}. The proposed method does not conflict with existing inpainting approaches but instead completes a more coarse-to-fine procedure.

\begin{figure*}[t]
\centering
\begin{tabular}{@{}c@{\hspace{0.3mm}}c@{\hspace{0.3mm}}c@{\hspace{1.2mm}}c@{\hspace{0.3mm}}c@{\hspace{0.3mm}}c@{}}
\includegraphics[width=0.163\linewidth]{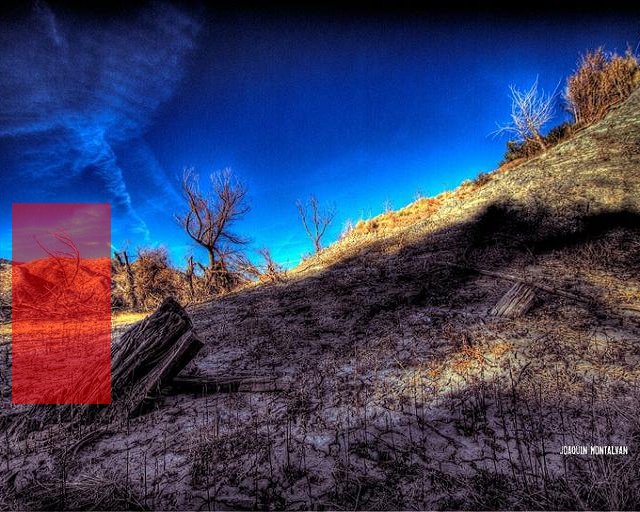}&
\includegraphics[width=0.163\linewidth]{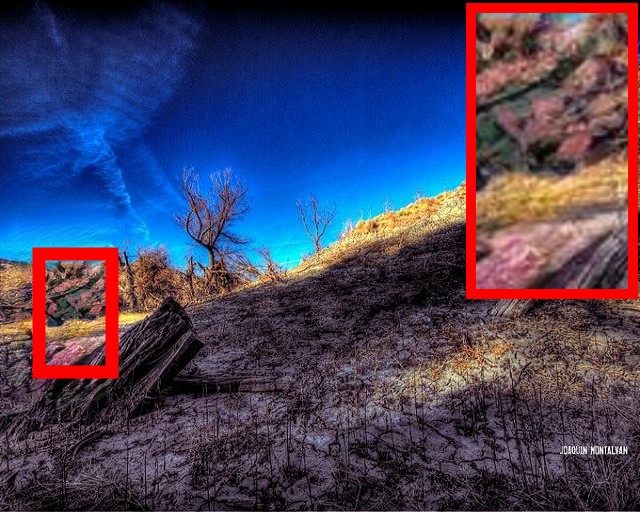}&
\includegraphics[width=0.163\linewidth]{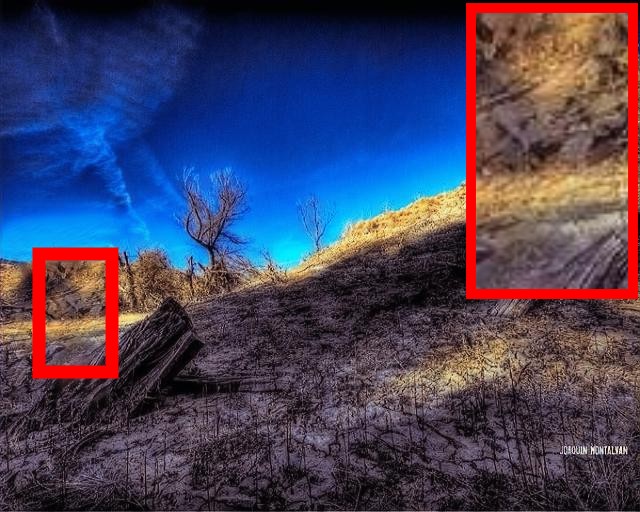}&
\includegraphics[width=0.163\linewidth]{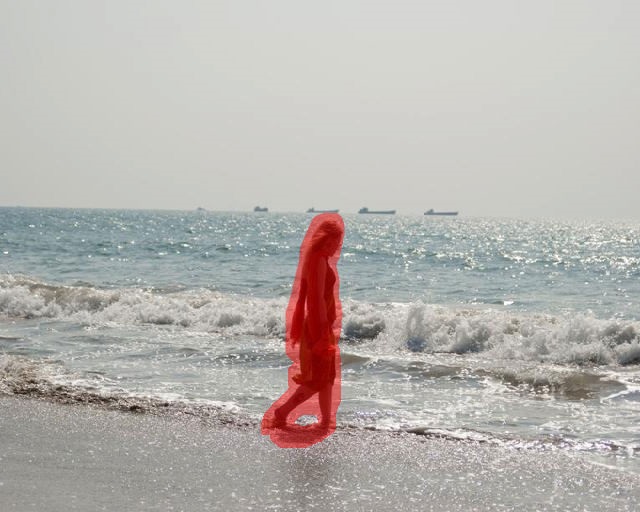}&
\includegraphics[width=0.163\linewidth]{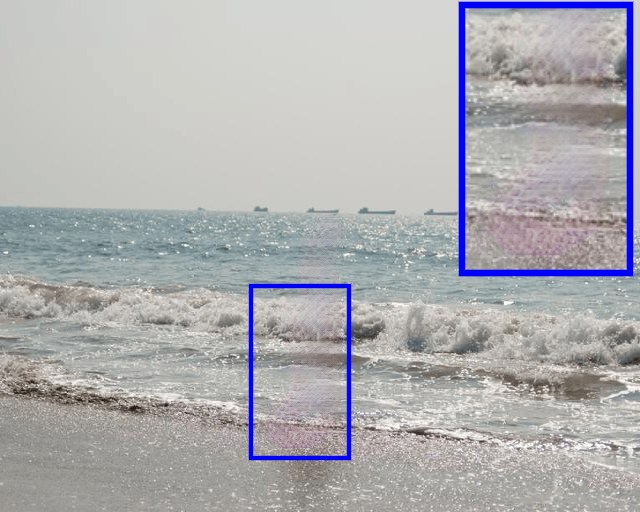}&
\includegraphics[width=0.163\linewidth]{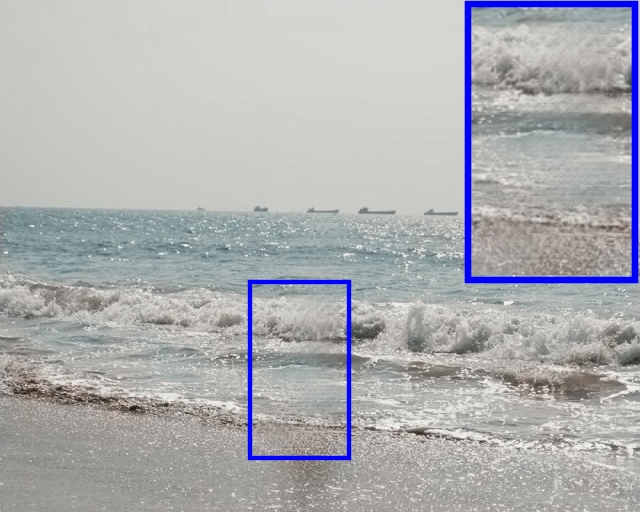}\\
\small{Input}& \small{ GatedConv~\cite{yu2019free} }  & \small{Ours (re-colorized)} &\small{Input}& \small{EdgeConnect~\cite{nazeri2019edgeconnect}}  & \small{Ours (re-colorized)} \\
\end{tabular}
\caption{Re-colorized results of applying our internal colorization method to GatedConv (left) and EdgeConnect (right). Colors of original results are defective and inconsist, while our re-colorized results are visually harmonized.}
\label{fig:recolorize}
\end{figure*}

\begin{figure*}[t]
\centering
\begin{tabular}{@{}c@{\hspace{0.4mm}}c@{\hspace{0.4mm}}c@{\hspace{0.4mm}}c@{\hspace{0.4mm}}c@{}}
\includegraphics[width=0.195\linewidth]{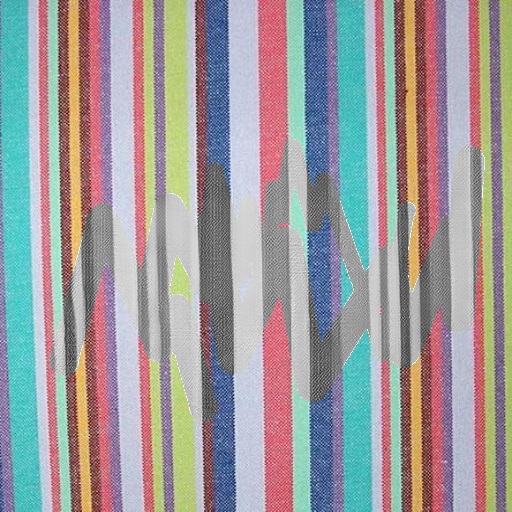}&
\includegraphics[width=0.195\linewidth]{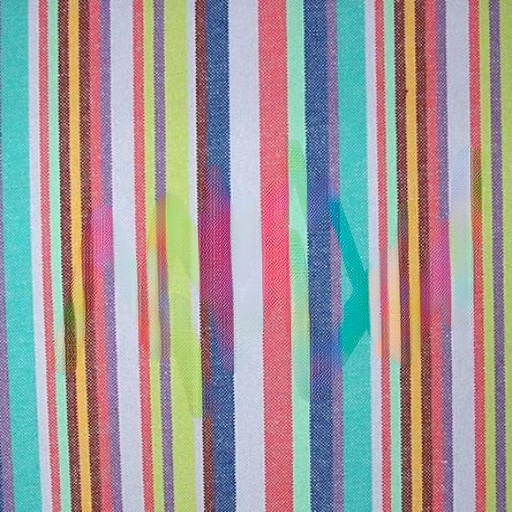}&
\includegraphics[width=0.195\linewidth]{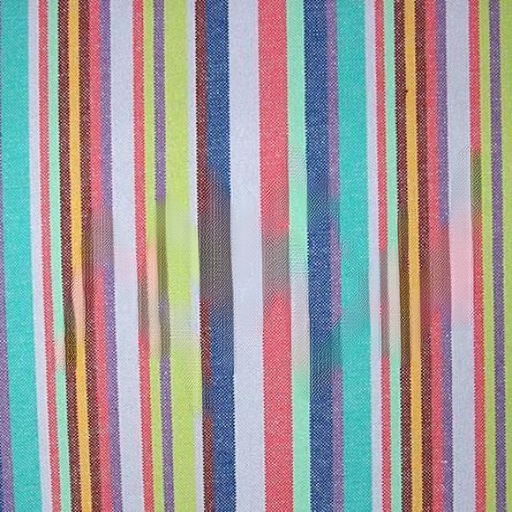}&
\includegraphics[width=0.195\linewidth]{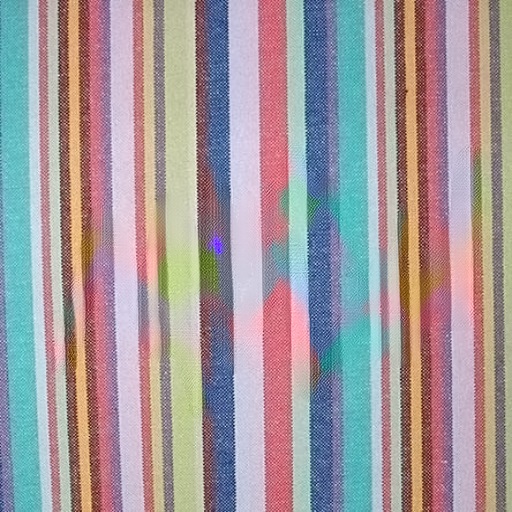}&
\includegraphics[width=0.195\linewidth]{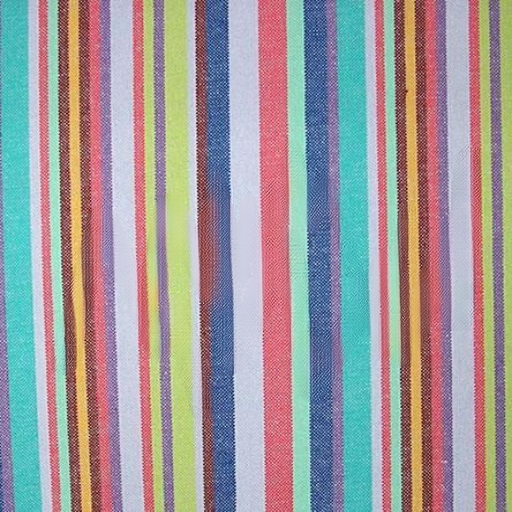}\\

\includegraphics[width=0.195\linewidth]{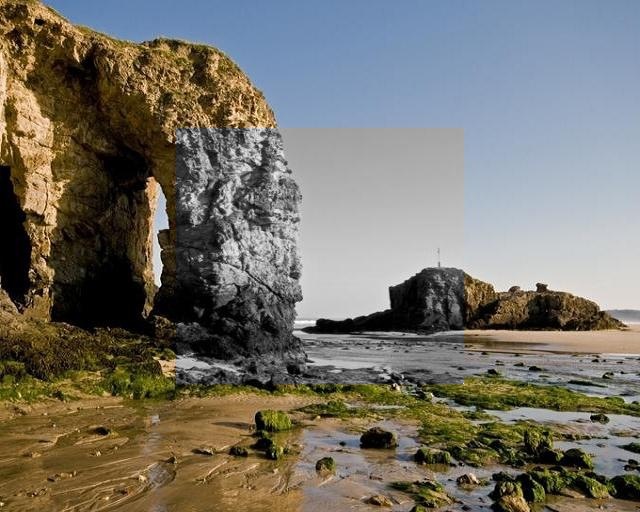}&
\includegraphics[width=0.195\linewidth]{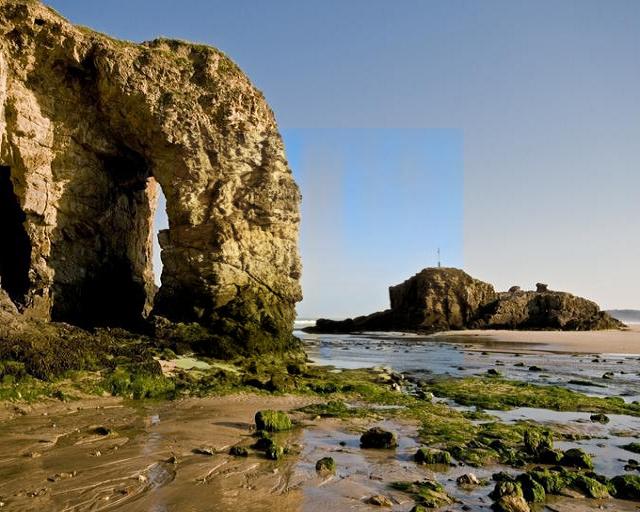}&
\includegraphics[width=0.195\linewidth]{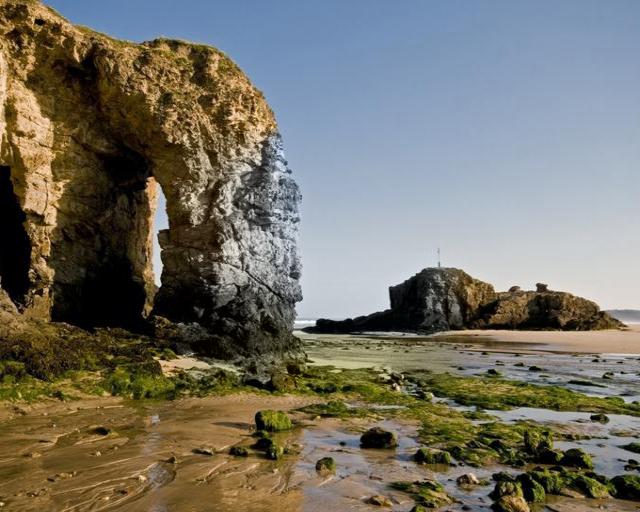}&
\includegraphics[width=0.195\linewidth]{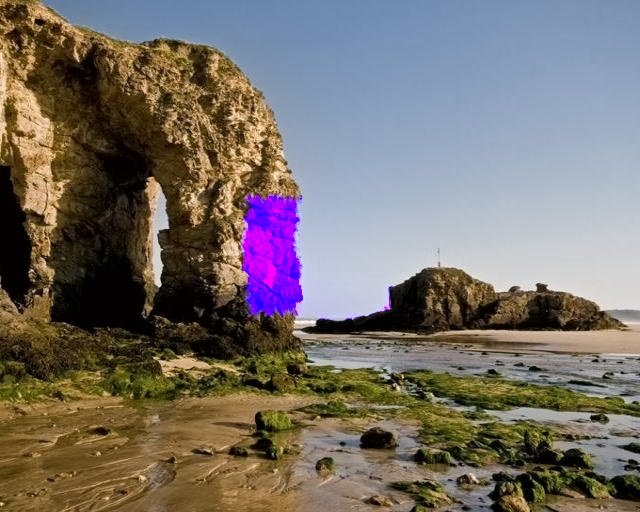}&
\includegraphics[width=0.195\linewidth]{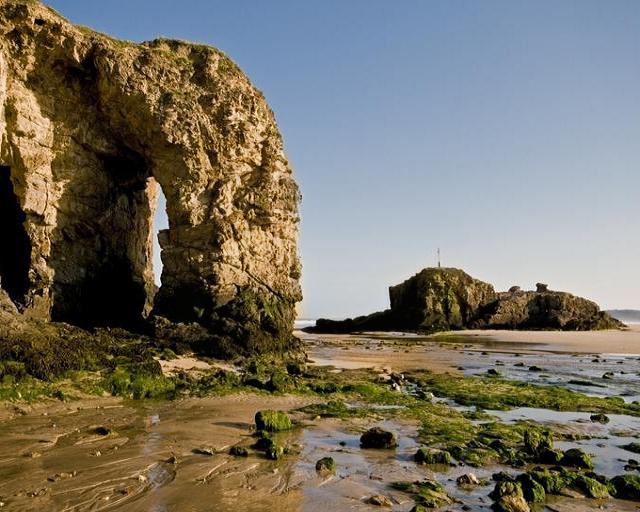}\\
   \small{Input}   & \small{Zhang et al.~\cite{zhang2017real}}  & \small{Levin et al.~\cite{levin2004colorization}}  & \small{Gastal et al.~\cite{gastal2011domain}}& \small{Ours} \\
\end{tabular}
\caption{Visual comparison with different guided colorization methods on the inpainted monochromic bottleneck (top) and natural (w/o inpainting) monochrome (bottom).  Zoom in for details.}
\label{fig:dtd-color}
\end{figure*}

\subsection{Motivation}
\subsubsection{Color Bleeding Removal}
\label{3.1}
Early image inpainting networks trained on large datasets usually suffer from the ``color bleeding'' artifacts. As shown in Fig.~\ref{fig:recolorize}, colors in the inpainted area of previous approaches~\cite{yu2019free,nazeri2019edgeconnect} show abrupt discrepancy from non-missing regions. For example, the green and pink color in the first image, and the purple color in the second image are very different from the color distribution of non-missing parts. This distribution gap indicates the possibility of improving inpainting quality by eliminating outliner colors in the missing region. Hence, we are motivated to further improve the color consistency by learning only from the internal color distribution of the non-missing parts. 

To show the visual quality gain brought by the internal colorization, we apply our  method to re-colorize the results of previous inpainting approaches. As in Fig.~\ref{fig:recolorize}, by strengthening the impact of internal color statistics in the single image, the abrupt colors can be eliminated.

\subsubsection{External-internal Learning}
However, learning only from internal statistics is inappropriate since external information is significant for content-aware image inpainting. A feasible solution is to set an intermediate bottleneck as a bridge between the external and internal learning. In traditional image reconstruction tasks, many researchers utilize monochromes to learn structures and then directly add color information back~\cite{sun2008image, shan2008fast}. Inspired from these works, we choose monochrome images as the intermediate output. This leads to another advantage that by reducing the output dimension from $\mathbb{R}^3$ to $\mathbb{R}^1$, the complexity of training is alleviated. We expect that models trained with monochromic bottlenecks can reconstruct higher-fidelity structures than original ones.

\subsection{External Monochromic  Reconstruction}
Our method can be easily applied to improve the reconstruction quality of  learning-based image inpainting models. Specifically, we concatenate the monochromic input to the original RGB input channel-wisely, and also modify the output from polychromic to monochromic images. We experiment with  representative inpainting baselines  as our reconstruction network:

\begin{itemize}[noitemsep,topsep=0pt]

  \item {\textbf{GMCNN}~\cite{wang2018image}}: a generative multi-column model, which synthesizes different image components in a parallel manner.
  \item {\textbf{HiFill}~\cite{yi2020contextual}}: a coarse-to-fine network for high-resolution images with light-weight gated convolution.  
  \item {\textbf{EdgeConnect}~\cite{nazeri2019edgeconnect}}: a two-stage adversarial method, which hallucinates missing edges first as guidance for image completion.
  \item {\textbf{GatedConv}~\cite{yu2019free}}: a coarse-to-fine network based on gated convolution, which achieves state-of-the-art inpainting performance with free-form masks. 
\end{itemize}

In our implementation, we convert an RGB image to a monochromic image by $0.30R + 0.59G + 0.11B$. For simplicity,  we denote our models with different  reconstruction networks as  \textbf{\textit{Ours (``backbone'')}}. 

\subsection{Internal Color Restoration}
\subsubsection{Guided Colorization}
In this stage, the input of the colorization network is the completed monochromic bottleneck from the first stage, while the goal is to restore colors consistent with the polychromic distribution of non-missing regions. We first tested with several guided colorization methods including: 

\begin{itemize}[noitemsep,topsep=0pt]

  \item \textbf{Zhang et al.}~\cite{zhang2017real}. A deep-learning based guided colorization method that learns semantic similarities from large datasets. 
  \item \textbf{Levin et al.}~\cite{levin2004colorization}. A quadratic optimization method that restores colors according to similar intensities.
  \item \textbf{Gastal et al.}~\cite{gastal2011domain}. A learning-free image processing method that is based on the edge-preserving filtering .
\end{itemize} 

However, as shown in Fig.~\ref{fig:dtd-color}, the external-learning method~\cite{zhang2017real} tends to magnify the inaccuracy in the monochrome and introduces color bleeding artifacts (e.g. ~red in the first example). On the contrary, utilizing color hints internally from the same image tends to avoid being confused by external color distributions. Previous learning-free methods~\cite{levin2004colorization,gastal2011domain} produce generally color-consistent results but fail in propagation when the mask region is large. We analyze the special features of our cases that are different from most of previous colorization settings as follows:  
\begin{itemize}
\renewcommand{\labelitemi}{\textbullet}
  \item  Unlike traditional sparse guidance such as color stroke and color palette, the guidance in our case is multiple accurate one-to-one mappings from monochrome to RGB. Since non-missing regions usually consist of an ample amount of pixels, the correspondence is extremely dense and covers most of patterns.
  \item Structures in the inpainted missing region  $I_{hole}$ and the non-missing region $I_{nhole}$ are often highly correlated.  
\end{itemize} 

Inspired by recent work~\cite{ulyanov2018deep}, rather than exploring similarity explicitly by feature matching, we propose to utilize a deep neural network $f$ to implicitly propagate color information. Specifically, we internally learn the color mapping function $f$ in the non-missing regions $I_{nhole}$  and directly apply it to the missing regions $I_{hole}$ for colorization.  However, similar monochromic inputs can map to different polychromic values even in a single image.   We,  therefore,  design a progressive colorization network to combine the local and global color context.

\subsubsection{Progressive Color Restoration}
Our model consists of a conditional generator pyramid $\{G_0,G_1,...,G_N\}$. We construct the corresponding grayscale image pyramid$\{I^g_0,I^g_1,...,I^g_N\}$,  color image pyramid $\{I^c_0,I^c_1,...,I^c_N\}$ and mask pyramid $\{M_0,M_1,...,M_N\}$ for internal learning. The colorization process begins at the coarsest scale and goes sequentially to the finest scale. In the coarsest scale, the model takes only the downsampled grayscale image:
\begin{align}
   \hat I_0 = G_0(I^g_0).
\end{align}
In the finer scale, the generator takes both the grayscale image and the upsampled color output from the lower level:
\begin{align}
   \hat I_n = G_n(I^g_n \oplus \hat I_{n-1} \uparrow), n=1,...,N
\end{align}
where $\oplus$ indicates concatenation in channel dimension and $\uparrow$ indicates bilinear upsampling. We adopt a ResNet-like architecture with box downsampling and bilinear upsampling for all generators. Since all the generators have the same receptive field, the model gradually captures global to local information as we process from coarse to fine. 

As the ground-truth pixels are only available in the non-missing region, we thus adopt a masked reconstruction loss for each generator, formulated as:

\begin{align}
   L_n = ||(\hat I_n - I^c_n) \odot (1 - M_n)||_1,
\end{align}
where $\odot$ indicates the Hadamard product. We use max-pooling for downsampling when building the mask pyramid to ensure that pixels from missing regions will not be included.

\section{Experiments} 
\subsection{Datasets}
 We  evaluate our method  on four public datasets: 

\textbf{Places2 Standard}~\cite{zhou2017places} contains more than 18 million natural images from 365 scene categories. We conduct experiment on all the categories, and use the original split for training. All images are resized to $512\times640$ when testing. 

\textbf{Paris StreetView}~\cite{pathak2016context} contains 15,000 outdoor building images. We use the original split for training. All images are resized to $256\times256$ when testing.

\textbf{CelebA-HQ}~\cite{karras2017progressive} contains 30,000 face images. We randomly select 3,000 images for testing, and others for training. All images are resized to $256\times256$ when testing.

\textbf{DTD}~\cite{cimpoi14describing} contains 5,640 texture images.We randomly select 840 images for testing, and others for training.  All images are resized to $512\times512$ when testing.

\textbf{Masks} We generate dense irregular masks by the algorithm proposed in~\cite{yu2019free}. In real-use cases, users usually behave like using an eraser
or brush to mask out undesired regions for inpainting. This algorithm simulates this behavior by randomly drawing lines and rotating angles, which are fair and suitable for our evaluation.

\subsection{Evaluation}

\begin{table*} [t]
\small
\centering 
\begin{tabular}{@{\hspace{1mm}}l @{\hspace{8mm}} c @{\hspace{3mm}} c @{\hspace{3mm}}  c @{\hspace{0.1mm}}c@{\hspace{6mm}} c @{\hspace{3mm}} c @{\hspace{3mm}} c @{\hspace{0.1mm}} c@{\hspace{6mm}}c @{\hspace{3mm}} c@{\hspace{3mm}} c @{\hspace{0.1mm}}c@{\hspace{6mm}} c @{\hspace{3mm}} c @{\hspace{3mm}} c @{\hspace{1mm}}} 
\toprule 
& \multicolumn{3}{c}{\textbf{Places2}}&& \multicolumn{3}{c}{\textbf{Paris Streetview}}&& \multicolumn{3}{c}{\textbf{CelebA-HQ}} && \multicolumn{3}{c}{\textbf{DTD}} \\ 
\cmidrule{2-4} \cmidrule{6-8} \cmidrule{10-12} \cmidrule{14-16}
Method&  PSNR & SSIM & LPIPS && PSNR & SSIM& LPIPS&& PSNR&SSIM&LPIPS&& PSNR&SSIM&LPIPS\\ 
\midrule 
GMCNN~\cite{wang2018image} & 22.18&  0.849 & 0.146&& 25.10 & 0.856 & 0.104&& 26.89 & 0.931 & 0.035&&27.58&0.932&0.071\\
Ours~(GMCNN) & 22.65&  0.858 & 0.133&& 25.67 & 0.859 & 0.097&& 27.03 & 0.933 &0.030&&28.30&0.945&0.057\\
EdgeConnect~\cite{nazeri2019edgeconnect}& 23.61&  0.874 &0.125 &&26.05& 0.863 & 0.088&& 27.24 & 0.944 & 0.027&&28.35&0.955&0.055\\
Ours~(EdgeConnect)& 23.90&  0.876 &0.117 &&26.36& 0.865 & 0.084&& 27.33 & \bf{0.947} & 0.026&&28.97&0.963&0.038\\
HiFill~\cite{yi2020contextual} & 24.35&  0.867 & 0.107&&  26.24 &  0.866 & 0.092 && 27.20  &  0.936 &  0.028&&29.14&0.950&0.046\\
Ours~(HiFill) & 24.52&  \bf{0.881} & 0.102&& 26.47  & 0.866  & 0.088 && 27.31  & 0.940 &0.026 &&\bf{29.38}&0.953&0.039\\
GatedConv~\cite{yu2019free} & 23.94&  0.871 & 0.112 &&26.32 & 0.861 & 0.090&& 27.36 & 0.938 & 0.028&&28.54&0.947&0.052\\ 
Ours~(GatedConv) & \bf{24.58}&  0.880 & \bf{0.098} &&\bf{26.75} & \bf{0.868} & \bf{0.082}&& \bf{27.51} & 0.945 & \bf{0.025}&& 29.31&\bf{0.961}&\bf{0.032}\\

\bottomrule
\end{tabular}
\vspace{1mm}
\caption{Quantitative comparisons on different datasets. The best results are \bf{boldfaced}. } 
\label{tab:metric} 
\end{table*}

\begin{figure*}[t]
    \centering
    \begin{tabular}{@{}c@{\hspace{0.3mm}}c@{\hspace{0.3mm}}c@{\hspace{0.3mm}}c@{\hspace{0.3mm}}c@{\hspace{0.3mm}}c@{\hspace{0.3mm}}c@{}}

    \includegraphics[width=0.14\linewidth]{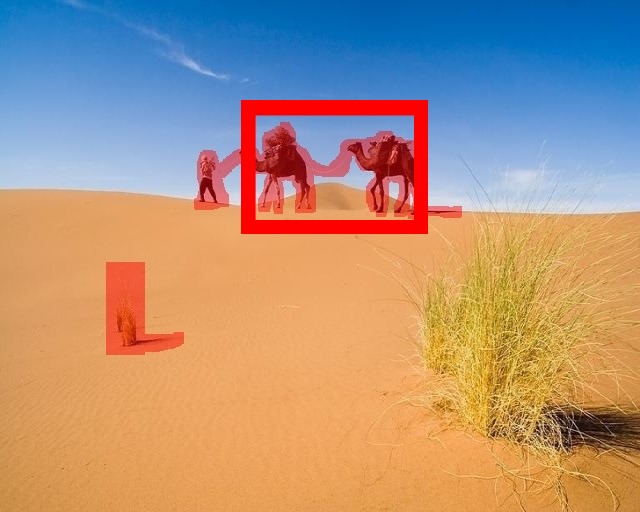}   &   
    \includegraphics[width=0.14\linewidth]{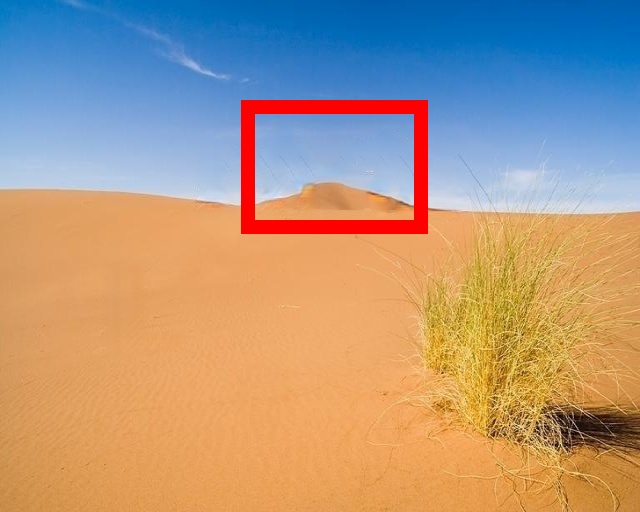}  & 
    \includegraphics[width=0.14\linewidth]{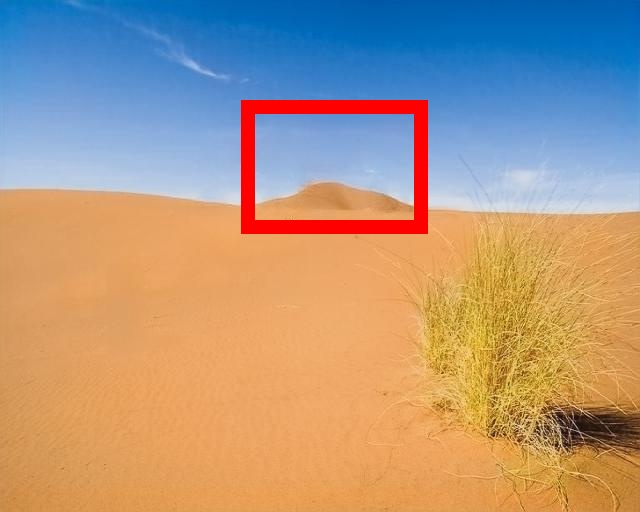}  & 
    \includegraphics[width=0.14\linewidth]{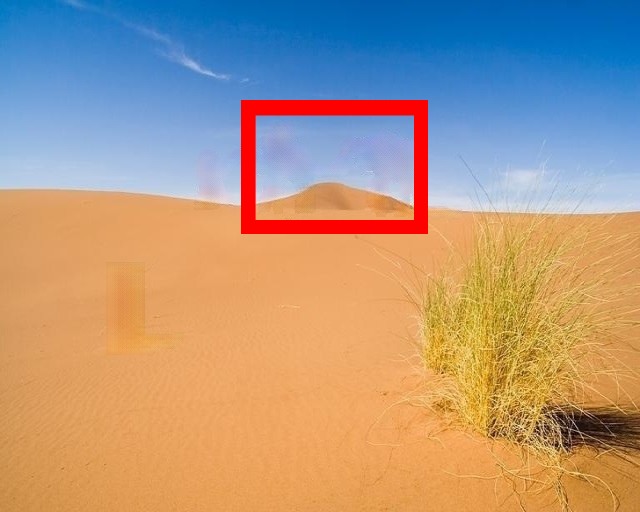} &   
    \includegraphics[width=0.14\linewidth]{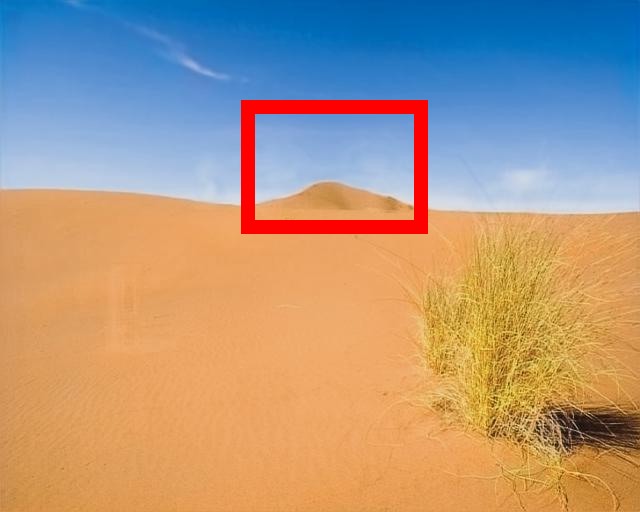} &
    \includegraphics[width=0.14\linewidth]{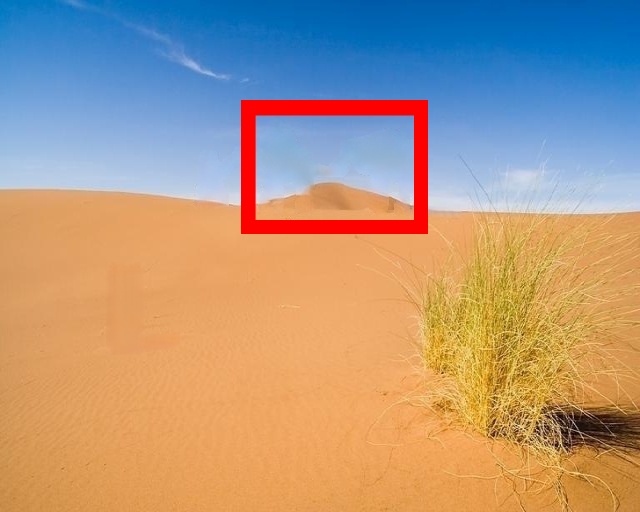}&
    \includegraphics[width=0.14\linewidth]{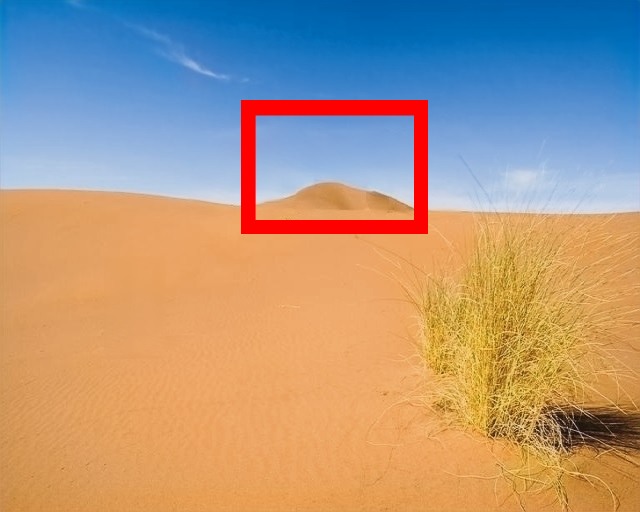}\\ 
    
    \includegraphics[width=0.14\linewidth]{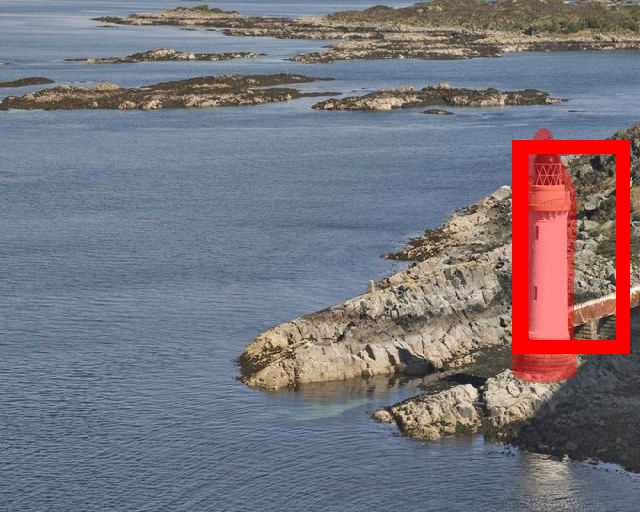}   &    \includegraphics[width=0.14\linewidth]{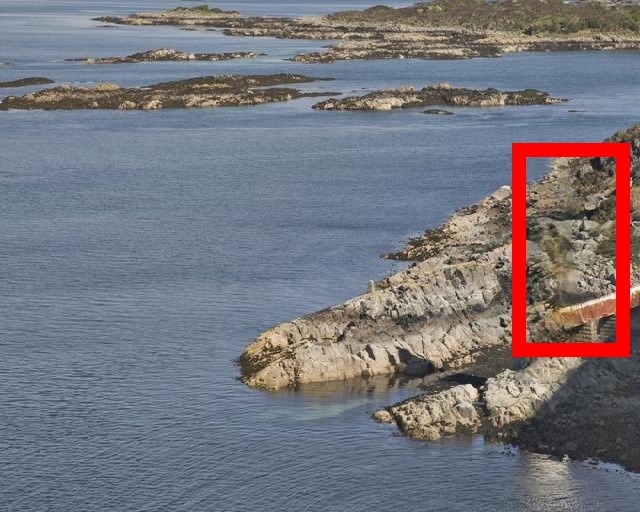}  & 
    \includegraphics[width=0.14\linewidth]{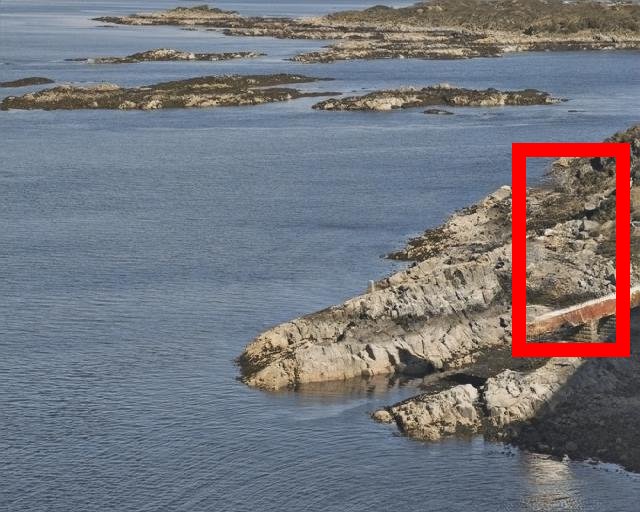}  & 
    \includegraphics[width=0.14\linewidth]{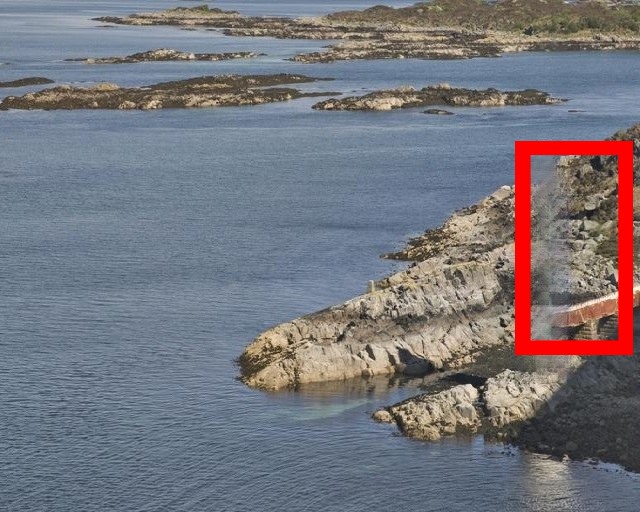} &    
    \includegraphics[width=0.14\linewidth]{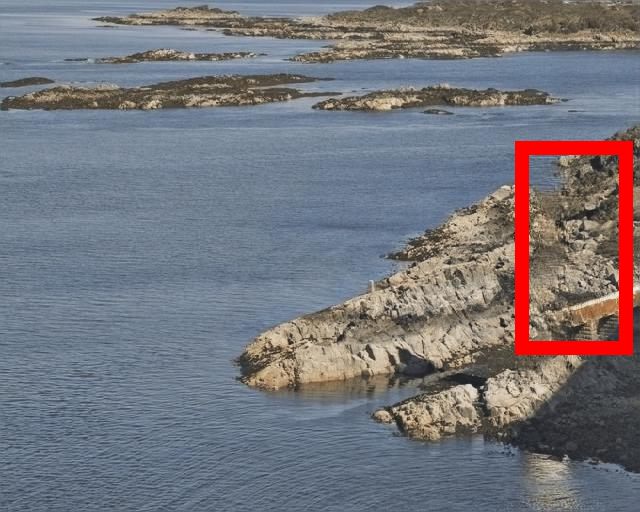} &
    \includegraphics[width=0.14\linewidth]{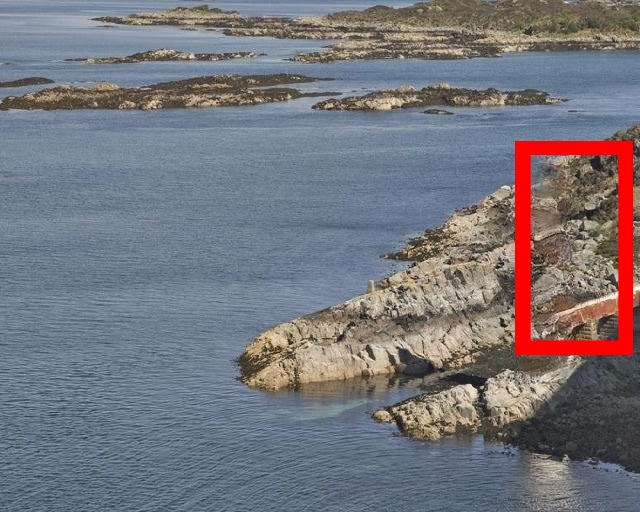}&
    \includegraphics[width=0.14\linewidth]{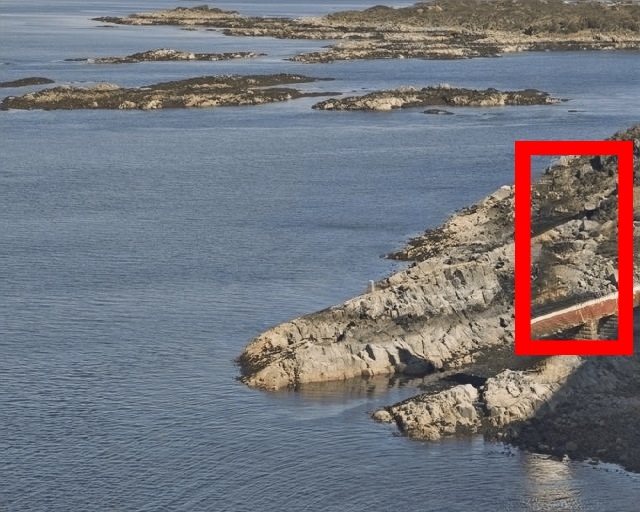}\\
    
     \includegraphics[width=0.14\linewidth]{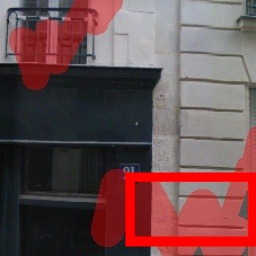} &    \includegraphics[width=0.14\linewidth]{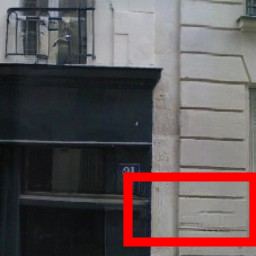}  &
     \includegraphics[width=0.14\linewidth]{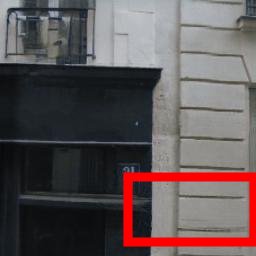}  &
     \includegraphics[width=0.14\linewidth]{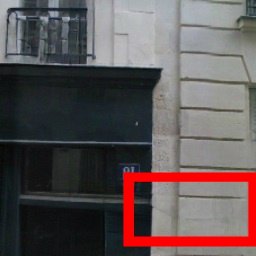} &   
     \includegraphics[width=0.14\linewidth]{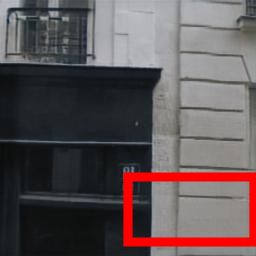} & 
     \includegraphics[width=0.14\linewidth]{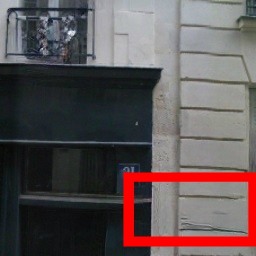}   &
     \includegraphics[width=0.14\linewidth]{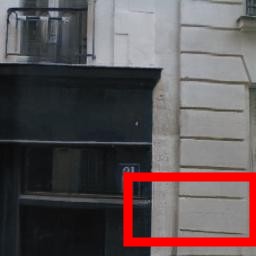}         \\     
     \includegraphics[width=0.14\linewidth]{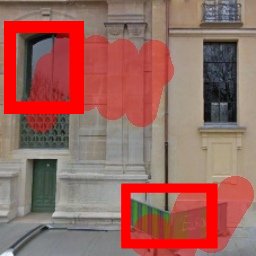}       &    \includegraphics[width=0.14\linewidth]{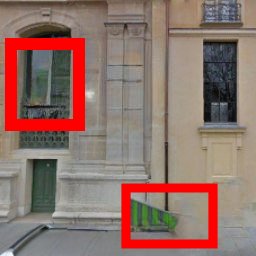}  &
     \includegraphics[width=0.14\linewidth]{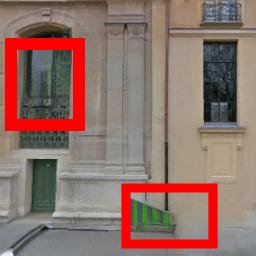}  &
     \includegraphics[width=0.14\linewidth]{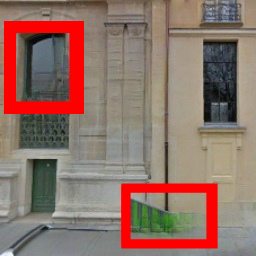} &   
     \includegraphics[width=0.14\linewidth]{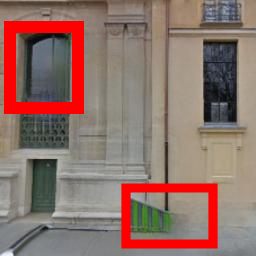} &    \includegraphics[width=0.14\linewidth]{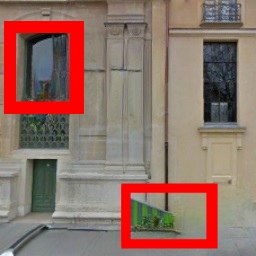}   &
     \includegraphics[width=0.14\linewidth]{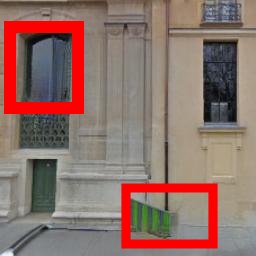}      \\       
 
    \scriptsize{Input}&\scriptsize{HiFill~\cite{yi2020contextual}}&\scriptsize{ Ours (HiFill)}&\scriptsize{EdgeConnect~\cite{nazeri2019edgeconnect}}&\scriptsize{ Ours (EdgeConnect)}&\scriptsize{GatedConv~\cite{yu2019free}}&\scriptsize{ Ours (GatedConv)}
    \end{tabular}
    \caption{Visual comparisons of different methods. Masked regions are visualized in red. Our method reconstructs coherent structures with fewer color artifacts. Zoom in for details.}
    \label{fig:places}
\end{figure*}

\paragraph{Quantitative Comparison}
As mentioned in previous work~\cite{yu2018generative}, there are no suitable objective metrics for inpainting tasks due to the ambiguity of ground truth. Nevertheless,  we still report evaluation results in terms of PSNR, SSIM~\cite{wang2004image}, and a learned perceptual metric LPIPS~\cite{zhang2018perceptual}. As shown in Table \ref{tab:metric}, for different backbone networks, the proposed external-internal scheme consistently improves the quantitative performance on diverse datasets.

\paragraph{Qualitative Comparison}
As shown in  Fig.~\ref{fig:places}, previous methods can produce semantically-reasonable content with small holes but still show blunt details and abrupt colors. As the hole becomes large, they tend to be unstable. In contrast, we observe that for each baseline network, our method produces compelling results with sharper structures and more consistent colors. This indicates that the proposed approach is not limited to one specific inpainting architecture but can be easily generalized to improve existing inpainting models.

\begin{figure}[t]
    \centering
    \includegraphics[width=\linewidth]{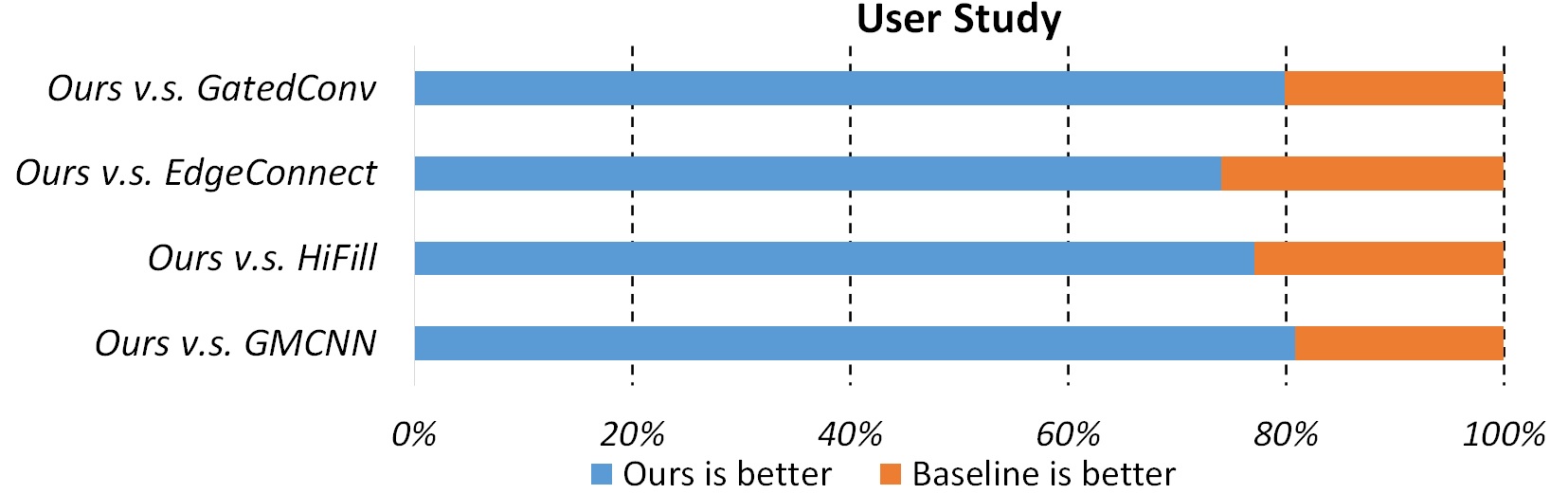}\\
    
    \caption{User study results. The reported value indicates the preference rate of Ours (`baseline') against the corresponding baseline.}
    \label{fig:user-study}
\end{figure}

\paragraph{User Study}
In addition to numerical metrics, we also perform a human perceptual study over the most challenging dataset Places2 on the Amazon Mechanical Turk. Participants are shown a random pair of images (ours and baseline) at once and are asked to select a more realistic image from the two in terms of both color consistency and structure preservation. All images are given at the same resolution in a shuffled order without time limitation. As shown in Fig.~\ref{fig:user-study}, models trained with the proposed scheme outperform the corresponding baselines perceptually by a large margin.

\subsection{Analysis on Monochromic Bottlenecks}
\label{4.3}
\subsubsection{Cross-dataset Analysis}
\begin{figure}[t]
\centering
\begin{tabular}{@{}c@{\hspace{0.5mm}}c@{\hspace{0.5mm}}c@{\hspace{1.5mm}}c@{\hspace{0.5mm}}c@{\hspace{0.5mm}}c@{}}

\includegraphics[width=0.24\linewidth]{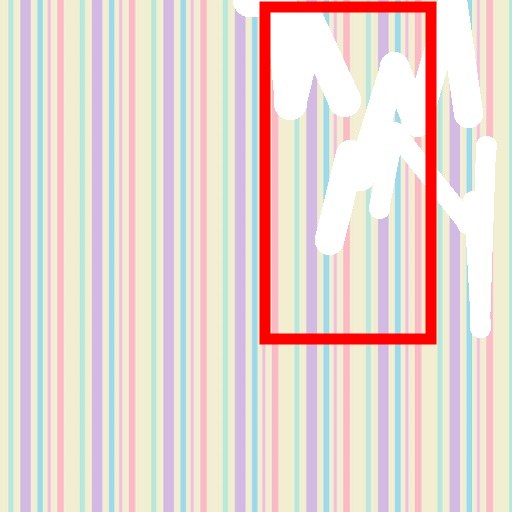}&
\includegraphics[width=0.12\linewidth]{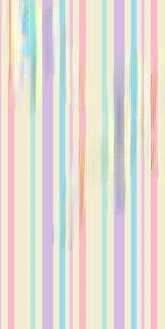}&
\includegraphics[width=0.12\linewidth]{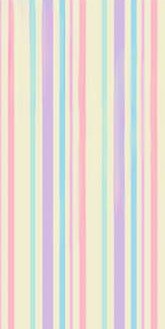}&
\includegraphics[width=0.24\linewidth]{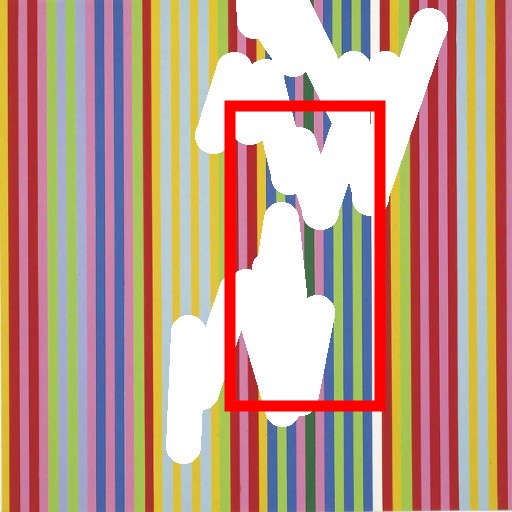}&
\includegraphics[width=0.12\linewidth]{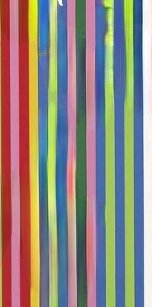}&
\includegraphics[width=0.12\linewidth]{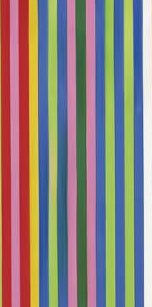}\\
\small {Input} & \cite{yu2019free} & \small {Ours} &\small {Input} & \cite{yu2019free} & \small {Ours}
\end{tabular}
\vspace{1mm}
\caption{Results of cross-dataset evaluation. We apply the models trained on Places2 to the unseen DTD images.}
\label{fig:cross-dataset}
\end{figure}

 Inpainting models trained on natural datasets usually show huge performance drop on images from other domains (e.g., textures) due to the distribution gaps. Although some frequent patterns (e.g., lines) in texture images are also ubiquitous in natural scenes, the distributions in polychromic space are still very different since some color patterns seldom appear in natural datasets. While in the monochromic space, this kind of gap is greatly narrowed. We conduct a cross-dataset evaluation to show the generalization ability gain brought by the monochromic bottleneck.   As shown in Fig.~\ref{fig:cross-dataset}, previous approaches show obvious structure distortion and color discrepancy in missing regions due to the distribution gap, while our model generates sharper lines and more consistent colors with seamless boundary. By learning to reconstruct structures in the monochromic space, the gap between different types of datasets is narrowed. We also find that if the domain gap between two datasets is huge (e.g., CelebA-HQ and others), our model fails to increase the cross-dataset test performance.  Otherwise, there is consistent improvement in Table~\ref{tab:cross}.
 
 \begin{table}[t]
\scriptsize
\centering 
\begin{tabular}{@{} c @{\hspace{2mm}} c@{\hspace{2mm}}  c@{\hspace{2mm}}  c@{\hspace{2mm}} c@{\hspace{2mm}} c@{\hspace{2mm}}  c@{\hspace{2mm}} c@{\hspace{2mm}}  c @{} } 
\toprule 
 Train &Test &Baseline &Ours&&Train &Test &Baseline &Ours\\ 

\cmidrule{1-4} \cmidrule{6-9}
Places2 &DTD & 21.85 & 23.16&& DTD&Places2 & 21.76 & 22.10\\
Places2 &Pairs & 26.24 & 26.57&& DTD&Paris & 24.62 & 24.85\\
Places2 &CelebA-HQ & 27.35 & 27.38&& DTD&CelebA-HQ & 26.83 & 26.86\\

\bottomrule \\
\end{tabular}
\caption{Quantitative results of cross-dataset evaluation.}
\label{tab:cross} 
\end{table}

\subsubsection{Effectiveness of Dimension Reduction}
 As observed in the above experiments, structures and shapes completed by our method are sharper than previous methods. Intuitively, it is easier to learn the reconstruction on monochrome than polychrome because the RGB optimization space $\mathbb{R}^3$ is much larger than the monochromic space of $\mathbb{R}$.  To better show the quality gain of reconstruction brought by this dimension reduction, we conduct further analysis on DTD. Since this dataset contains thousands of simple texture images such as line, circle, checkerboard with extremely diverse colors, it is a felicitous example to demonstrate the quality of structure reconstruction.
 
\begin{figure}[t]
    \centering 
    \begin{tabular}{@{}c@{\hspace{0.2mm}}c@{\hspace{0.2mm}}c@{\hspace{0.2mm}}c@{\hspace{0.2mm}}c@{}}
      \includegraphics[width=0.198\linewidth]{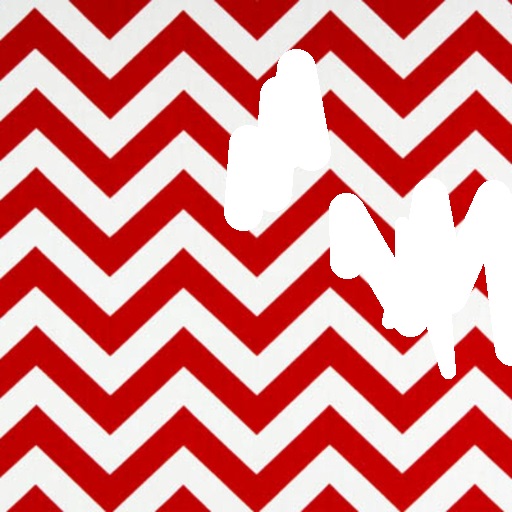} &  \includegraphics[width=0.198\linewidth]{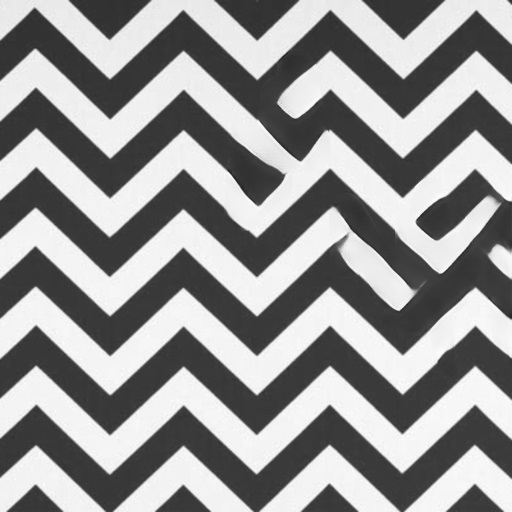} & \includegraphics[width=0.198\linewidth]{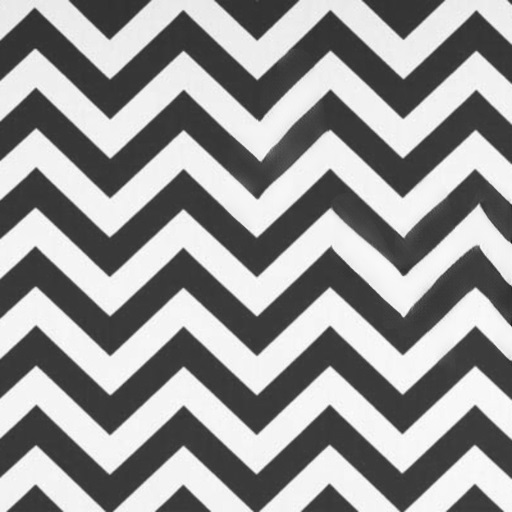}   &
      \includegraphics[width=0.198\linewidth]{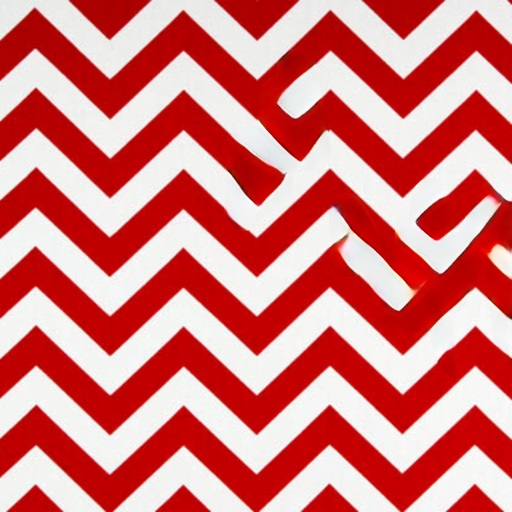}  & 
      \includegraphics[width=0.198\linewidth]{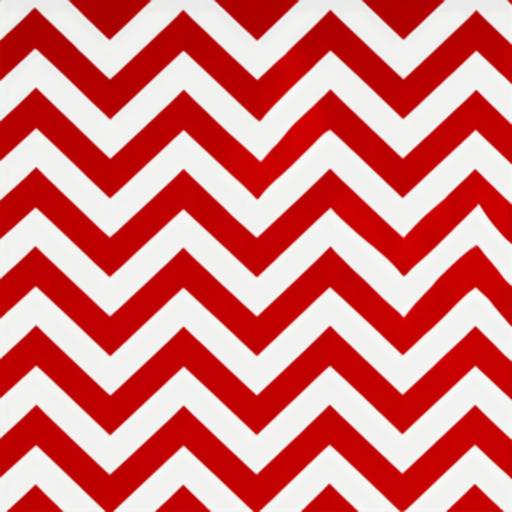}   \\   
      \includegraphics[width=0.198\linewidth]{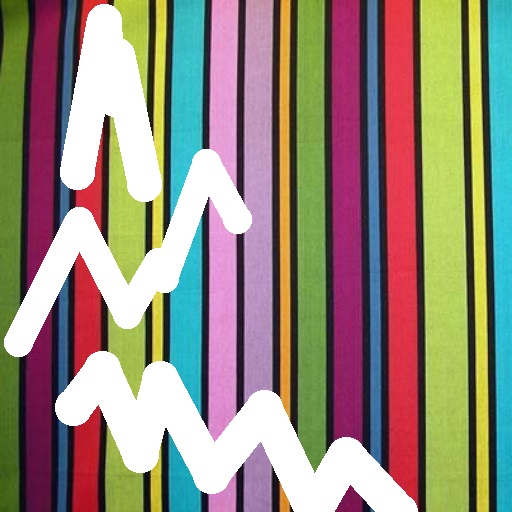} &  \includegraphics[width=0.198\linewidth]{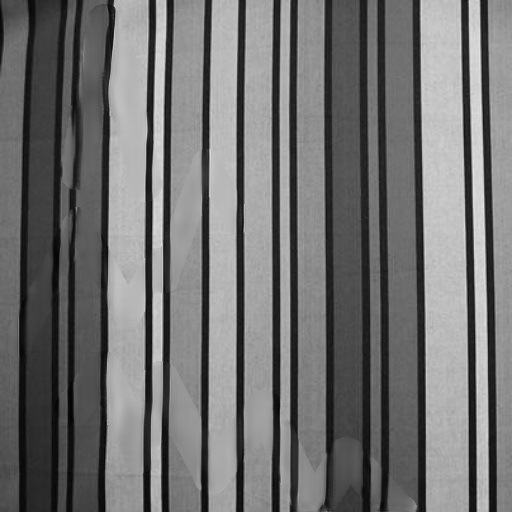} & \includegraphics[width=0.198\linewidth]{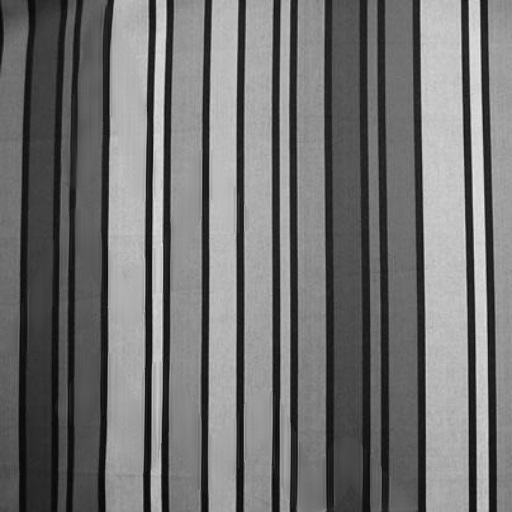}   & 
      \includegraphics[width=0.198\linewidth]{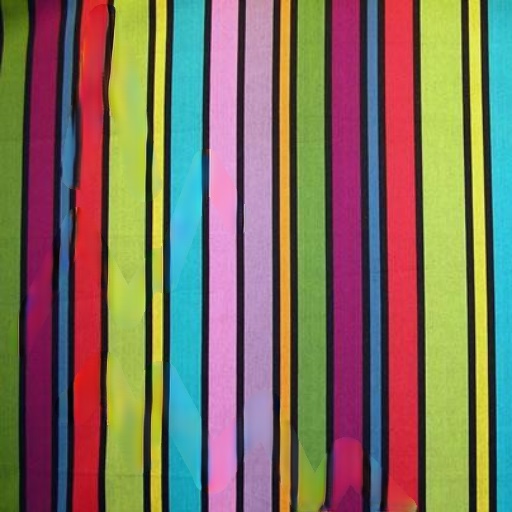} &
      \includegraphics[width=0.198\linewidth]{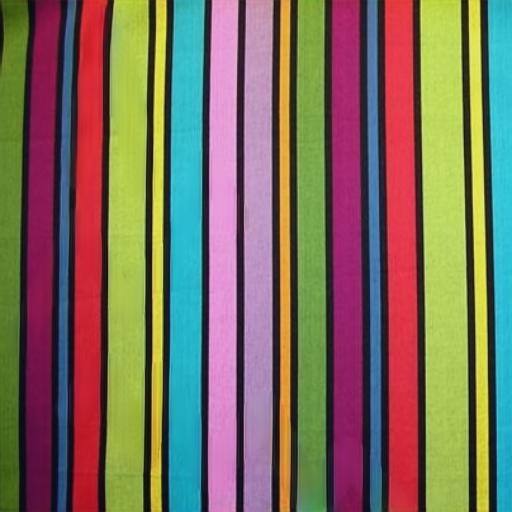}  \\        
      \includegraphics[width=0.198\linewidth]{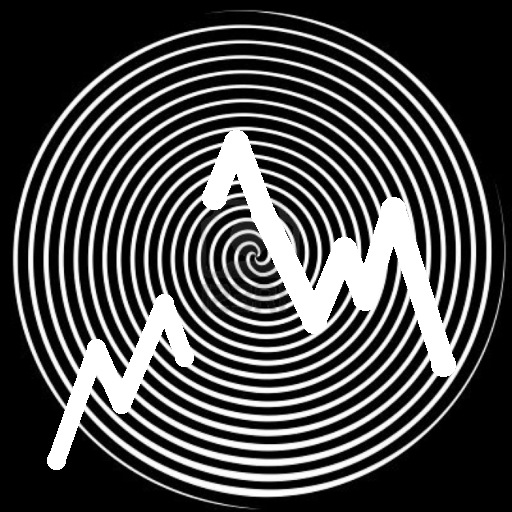} &
      \includegraphics[width=0.198\linewidth]{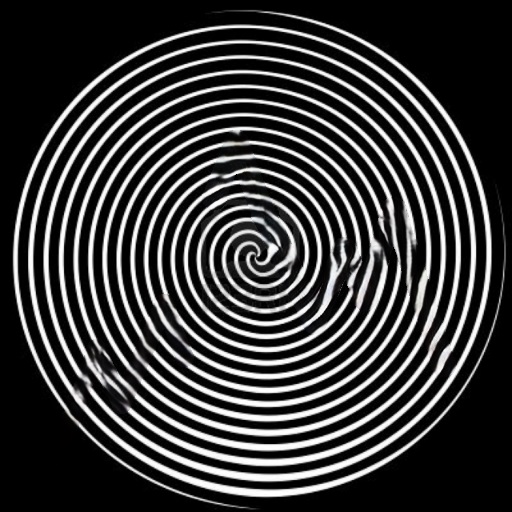} &
      \includegraphics[width=0.198\linewidth]{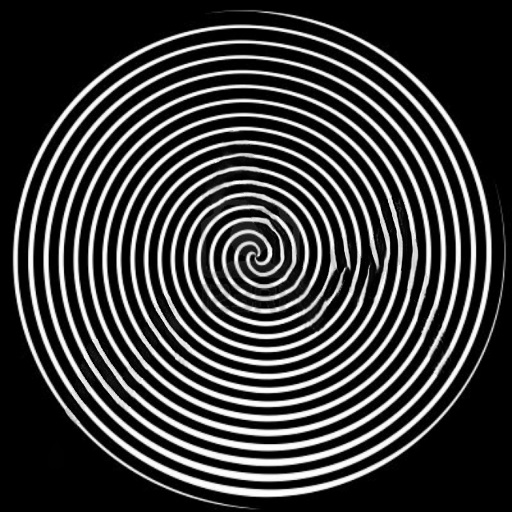}   & 
      \includegraphics[width=0.198\linewidth]{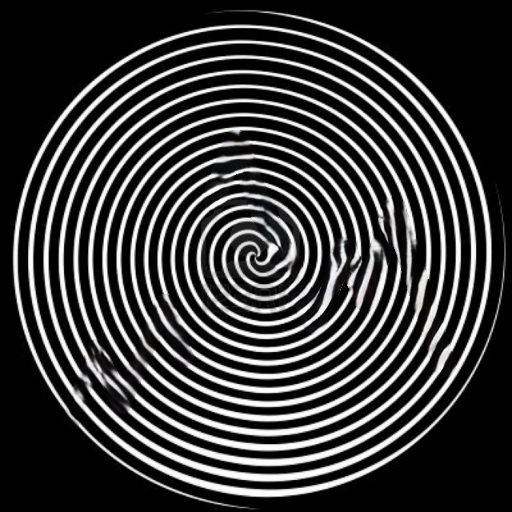}  & 
      \includegraphics[width=0.198\linewidth]{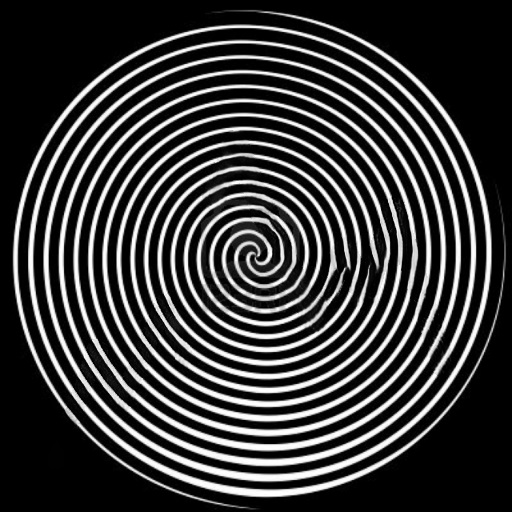}   \\        

    \scriptsize{Input}   & \scriptsize{GatedConv}  & \scriptsize{Ours} & \scriptsize{GatedConv} & \scriptsize{Ours} \\
    \end{tabular}
    \caption{Visual comparisons of reconstruction quality on DTD. We show grayscale images converted from the results of GatedConv in the second column for better comparison.  }
    \label{ fig:dtd-geometry}
\end{figure}

 As shown in Fig. \ref{ fig:dtd-geometry}, the original baseline model produces curved and blunt details in straight lines. The original model also fails to produce consistent color and seamless boundary when filling in regions with diverse colors. However, the model trained on monochromic space is able to capture the essence of structures and complete correct shapes. It indicates that ignoring color distraction can alleviate the learning complexity and facilitates structure reconstruction. 

\subsection{Analysis on Internal Color Restoration}
\subsubsection{Ablation Study on Mask Ratios}
\begin{figure}[t]
    \centering
    \small
    \begin{tabular}{@{}c@{\hspace{0.2mm}}c@{\hspace{0.2mm}}c@{\hspace{0.2mm}}c@{}}
 
     \includegraphics[width=0.247\linewidth]{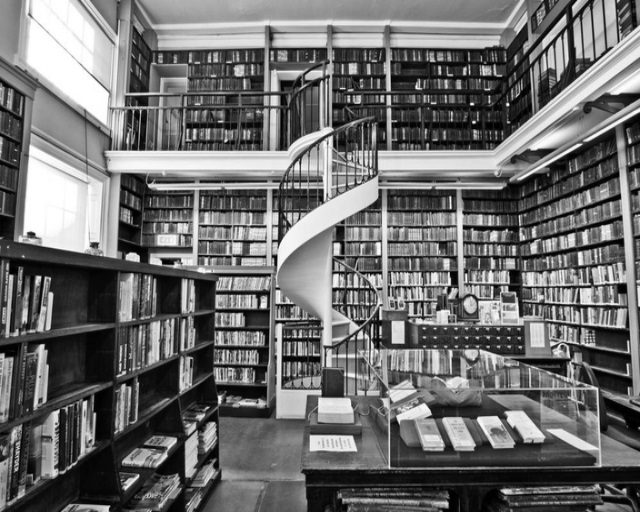}&
    \includegraphics[width=0.247\linewidth]{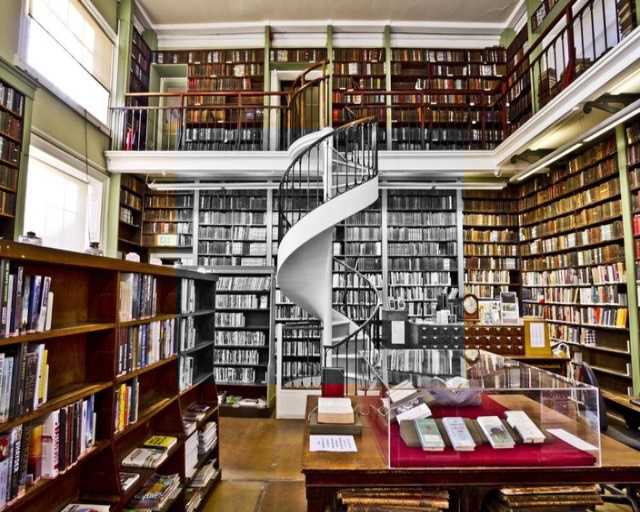}  & 
    \includegraphics[width=0.247\linewidth]{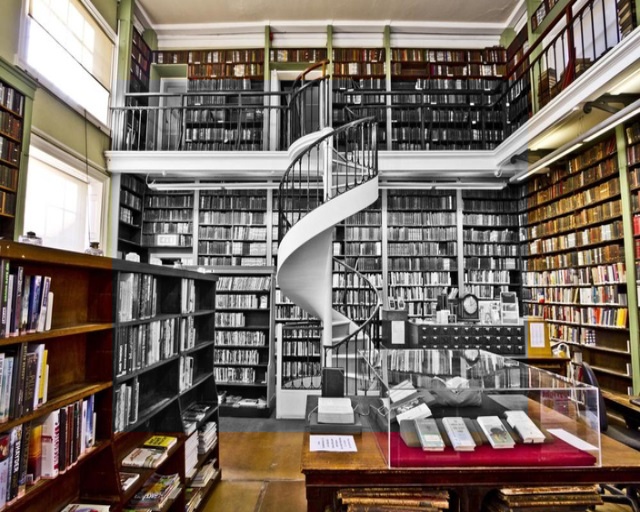} &    
    \includegraphics[width=0.247\linewidth]{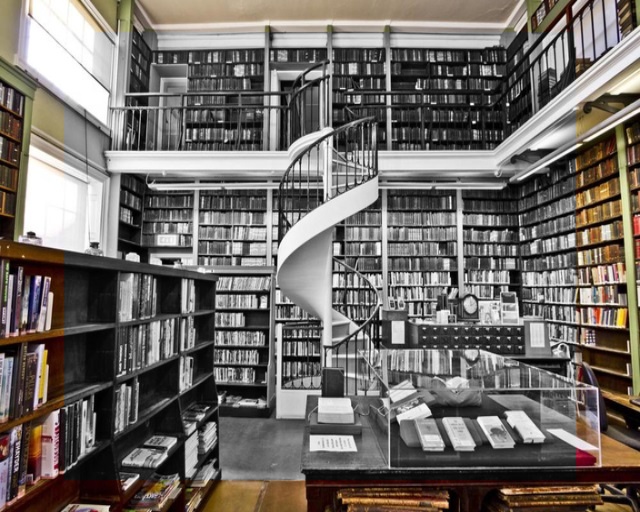}\\
      \includegraphics[width=0.247\linewidth]{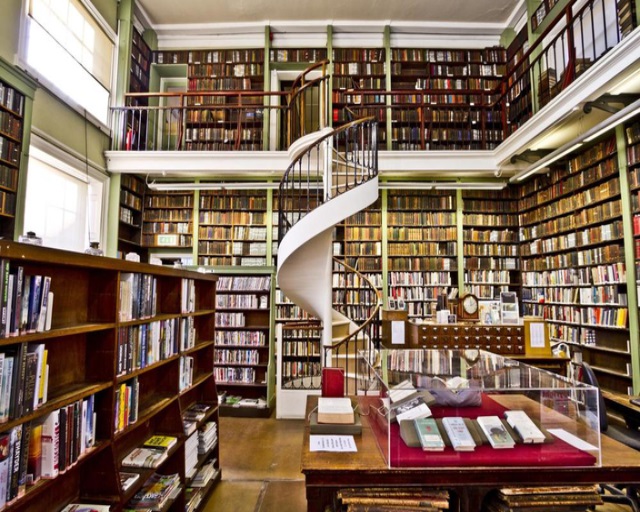}&
    \includegraphics[width=0.247\linewidth]{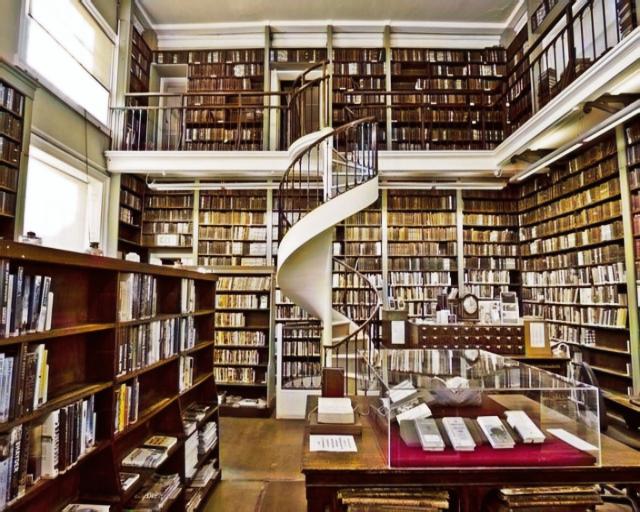}  & 
    \includegraphics[width=0.247\linewidth]{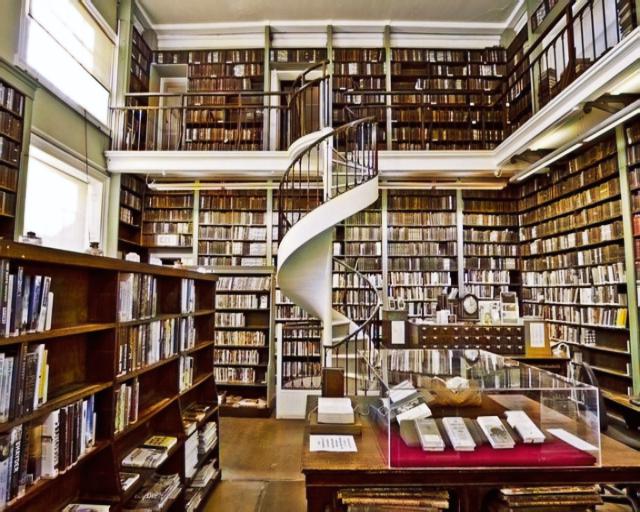} &    
    \includegraphics[width=0.247\linewidth]{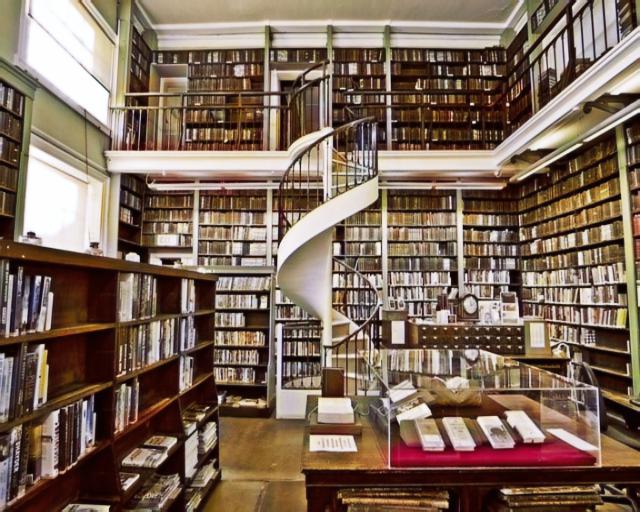}\\
  ground-truth& $22.5\%$  & $48.9\%$& $73.4\%$
    \end{tabular}
    \caption{Feasibility of our internal colorization method. We increase the mask ratio of $I_{hole}$ from $22.5\%$ to $73.4\%$ and colorize the natural monochrome with our method.}
    \label{fig:mask_ratio}
\end{figure}

One key factor that may affect the performance of our internal colorization method is the number of known pixel correspondences. In Fig.~\ref{fig:mask_ratio}, we increase the mask ratio of $I_{hole}$ from $22.5\%$ to $73.4\%$ and restore the color of a natural monochrome with our approach.  Even in the most challenging case where $73.4\%$ pixels are missing, the model still colorizes $I_{hole}$ in a harmonized style with $I_{nhole}$  without noticeable artifacts. Since in image inpainting, $I_{hole}$ usually accounts for less than  $70\%$ of the entire image, and the proposed internal scheme is feasible in most cases.

\subsubsection{Ablation Study on Progressive Restoration}
\begin{figure}[t]
\centering
\begin{tabular}{@{}c@{\hspace{0.2mm}}c@{\hspace{0.2mm}}c@{}}
\includegraphics[width=0.33\linewidth]{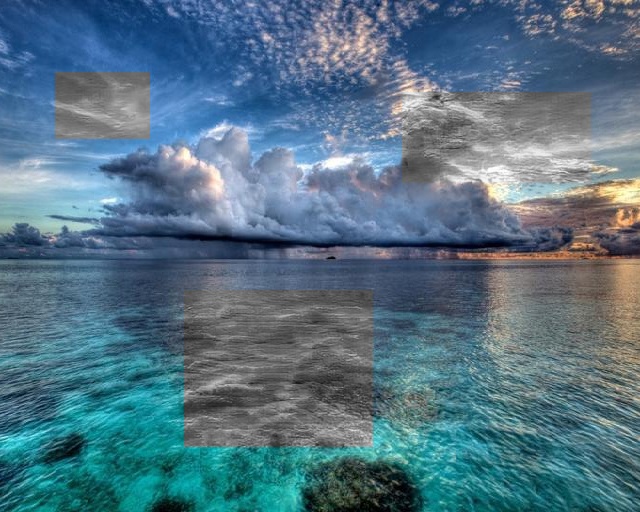}&
\includegraphics[width=0.33\linewidth]{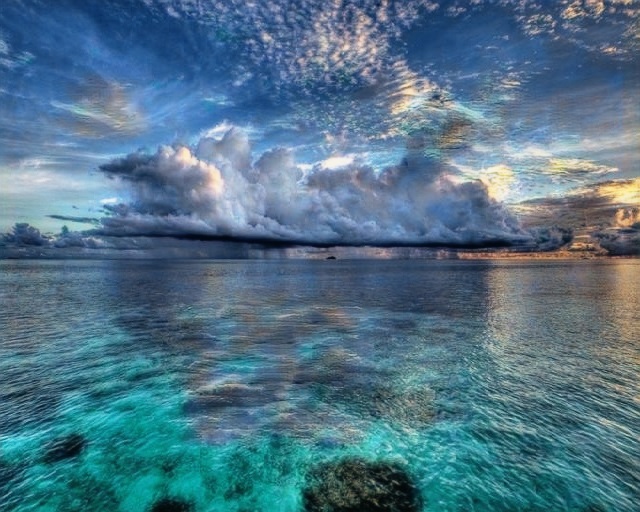}&
\includegraphics[width=0.33\linewidth]{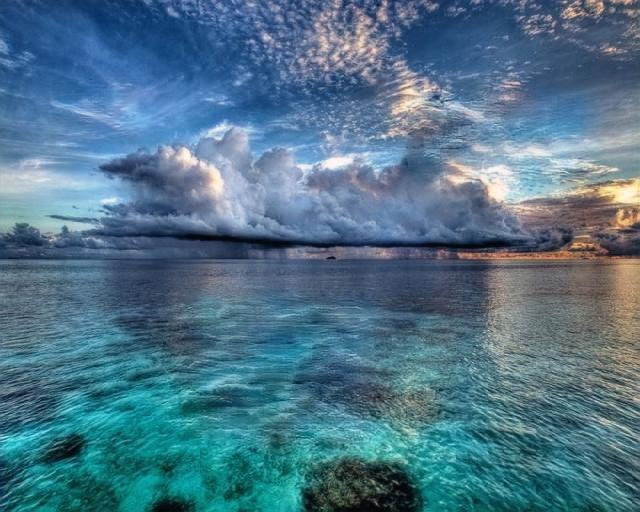}\\
\small{Input}  & \small{Ours~(base)} & \small{Ours~(full)} 
\end{tabular}
\caption{Ablation study of the progressive restoration strategy. We colorize the inpainted monochromic bottleneck with our method. Ours (base) is our model without the progressive design. }
\label{fig:ms}
\end{figure}

 We conduct ablation study to figure out how the progressive restoration strategy contributes to the internal colorization. Fig. \ref{fig:ms} shows that without the progressive scheme, our  model focuses only on local color mappings and generates obvious artifacts and hard boundaries. 

\subsubsection{Comparison with Other Colorization Methods}
As we discussed above, the inpainted monochromes are possibly different from the ground truth due to the ambiguity of the inpainting task. It is inappropriate to evaluate the colorization method by comparing colorized inpainted-monochrome with the imperfect ground truth. To avoid the influence of monochrome discrepancy and evaluate the colorization separately, we apply random rectangular and irregular masks to de-colorize ground-truth images from the most challenging dataset Places2. In this way, we simulate the behavior of inpainting masks while having the perfect ground-truth color images for metric calculation. The proposed method achieves a stable performance in Table~\ref{tab:color-tab}. 

\begin{table}[t]
\scriptsize
\centering 
\begin{tabular}{@{} l @{\hspace{3mm}} c@{\hspace{1.6mm}}  c@{\hspace{1.6mm}}  c@{\hspace{1.6mm}}   c@{\hspace{1.6mm}} c @{} } 
\toprule 
 Mask type&\textbf{Zhang et al.} &\textbf{Gastal et al.} &\textbf{Levin et al.}  &\textbf{Ours~(base)}&\textbf{Ours~(full)}\\ 

\midrule 
Rectangular &36.12 & 29.35& 36.68& 37.04 &\bf{38.45}\\
Irregular &38.77 & 39.26 & 39.24& 39.21 &\bf{39.50}\\
\bottomrule \\
\end{tabular}
\caption{Results (PSNR) of different guided colorization methods on natural monochromes with different types of masks. Ours (base) is our model without the progressive design.}
\label{tab:color-tab} 
\end{table}

\begin{figure}[!th]
\centering
\begin{tabular}{@{}c@{\hspace{0.2mm}}c@{\hspace{0.2mm}}c@{\hspace{0.2mm}}c@{}}

\includegraphics[width=0.245\linewidth]{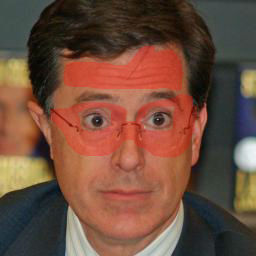}&
\includegraphics[width=0.245\linewidth]{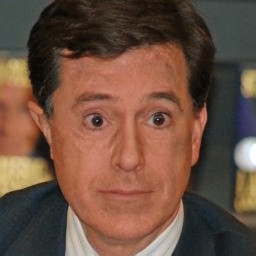}&
\includegraphics[width=0.245\linewidth]{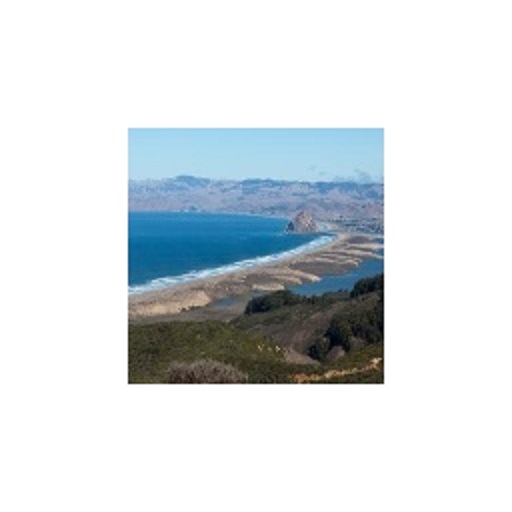}&
\includegraphics[width=0.245\linewidth]{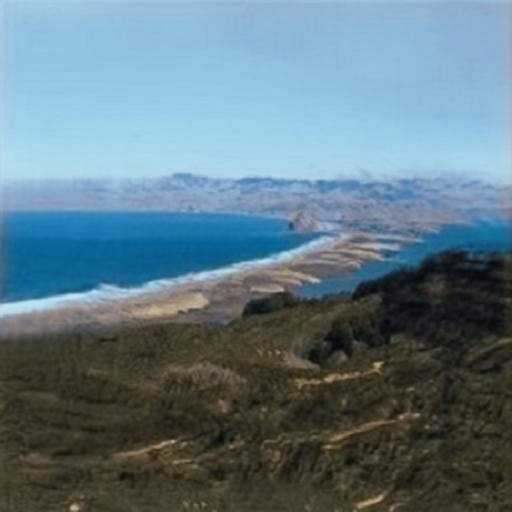}\\

 \small{Input} &  \small{Ours} & \small{Input} & \small{Ours}  \\
 \end{tabular}
 \begin{tabular}{@{}c@{\hspace{1mm}}c@{\hspace{0.2mm}}c@{\hspace{0.2mm}}c@{\hspace{0.2mm}}c@{\hspace{0.2mm}}c@{}}
\includegraphics[width=0.253\linewidth]{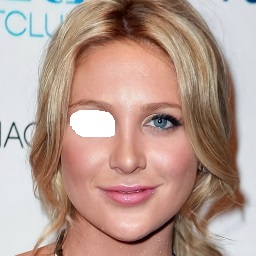}&
\includegraphics[width=0.18\linewidth]{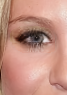}&
\includegraphics[width=0.18\linewidth]{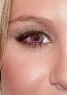}&
\includegraphics[width=0.18\linewidth]{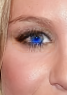}&
\includegraphics[width=0.18\linewidth]{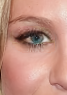}\\
\small{Input}&\multicolumn{4}{c}{ \small{Ours~( user-guided diverse outputs )  } }
\end{tabular} 
\caption{Examples of image editing, extrapolation, and user-guided inpainting. Users can control the style of inpainted content with our approach by giving extra color hints. }
\label{fig:extensions}
\end{figure}

\subsection{Extensions}
In Fig.~\ref{fig:extensions}, we show some applications of the proposed method. We demonstrate one user-guided inpainting example, where users can control the color of generated content by giving few color hints interactively. We utilize one extra color point as guidance to inpaint eyes with different colors. 

\subsection{Failure Cases}
\begin{figure}[t]
\centering
\begin{tabular}{@{}c@{\hspace{0.2mm}}c@{\hspace{0.2mm}}c@{\hspace{0.2mm}}c@{\hspace{0.8mm}}c@{}}
\rotatebox{90}{\small{\qquad Input}}&
\includegraphics[width=0.25\linewidth]{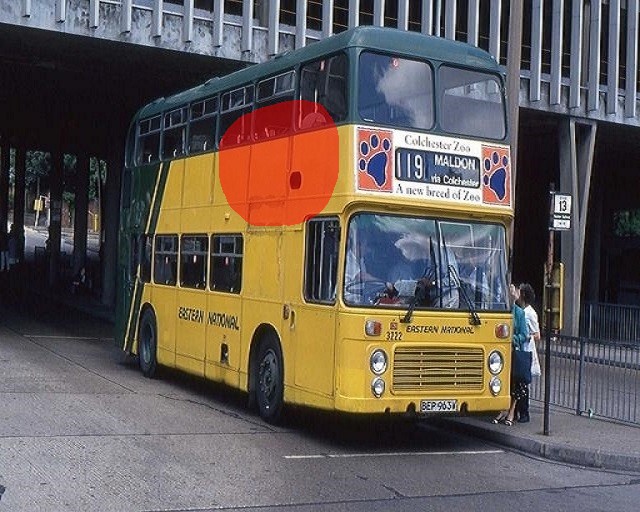}&
\includegraphics[width=0.25\linewidth]{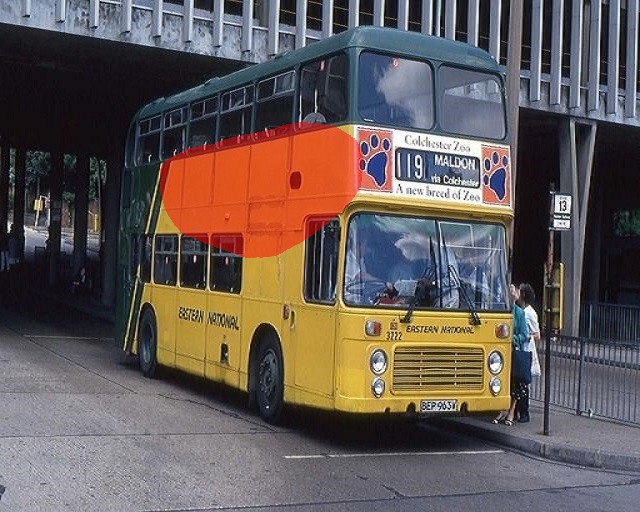}&
\includegraphics[width=0.25\linewidth]{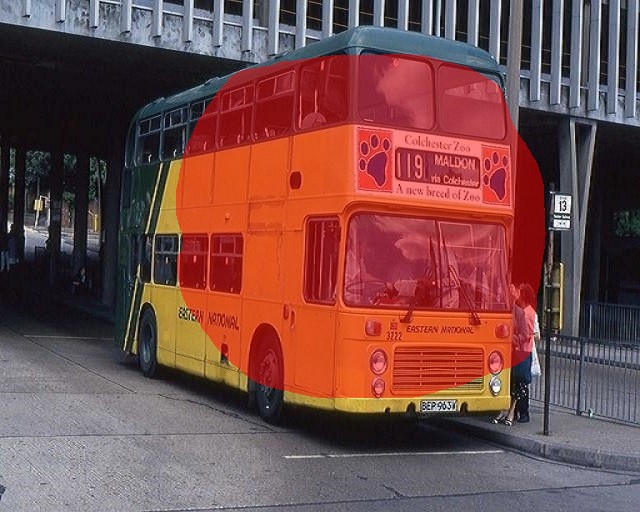}&
\includegraphics[width=0.2\linewidth]{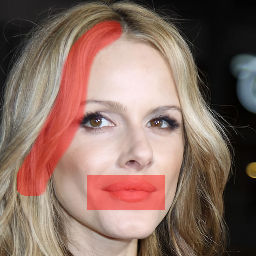}\\
\rotatebox{90}{\small{\qquad Ours}}&
\includegraphics[width=0.25\linewidth]{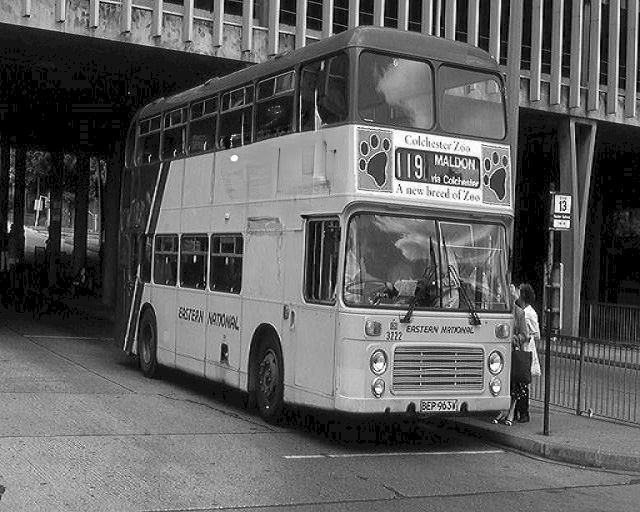}&
\includegraphics[width=0.25\linewidth]{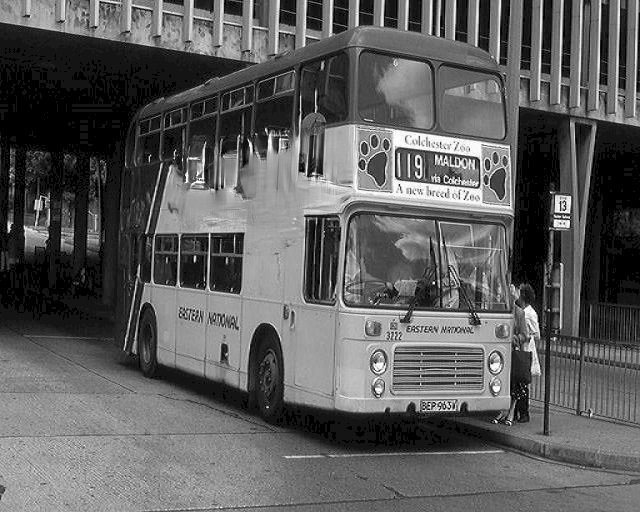}&
\includegraphics[width=0.25\linewidth]{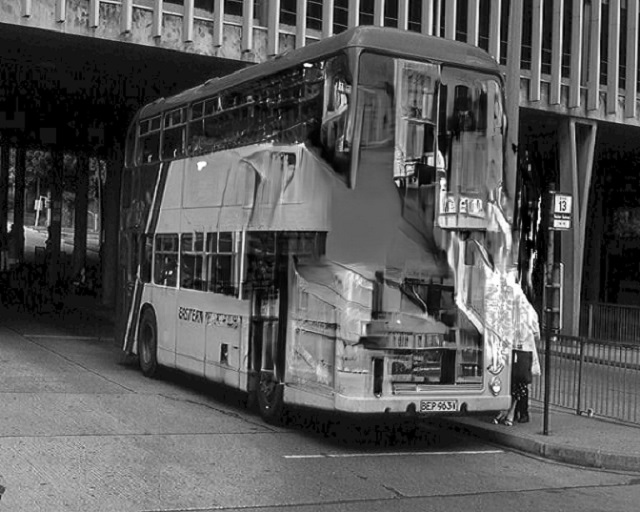}&
\includegraphics[width=0.2\linewidth]{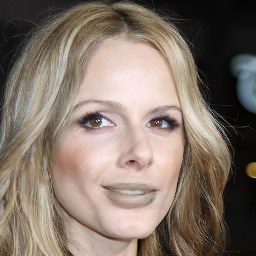}\\
\end{tabular}
\caption{Failure cases. We show failure cases of both reconstruction and colorization. }
\label{fig:failure}
\end{figure}

Fig.~\ref{fig:failure} shows  failure cases of the proposed inpainting method. In the first example, our method fails to reconstruct  a partially-masked bus when the mask is extremely large. Similar to previous inpainting approaches, our method  has difficulty in completing largely-masked foreground objects, since the structures of these categories are highly complex and diverse. In the second example, our model incorrectly colorizes the mouth  due to the lack of colorization hints. In this case, we can consider giving one extra color point as guidance to facilitate the color restoration. 

\section{Conclusion}
In this paper, we propose a general external-internal learning inpainting scheme with monochromic bottlenecks. It first reconstructs the monochrome utilizing semantic knowledge learned externally from large datasets, and then recovers colors internally from a single test image. Our method can produce more coherent structures and more visually harmonized colors compared with previous approaches. Extensive experiments show that our method can lead to stable improvement qualitatively and quantitatively on several backbone models. The major limitation of our method is the inference speed. Since an extra stage is needed for colorization, our method is slower than state-of-the-art approaches.  In the future, we plan to accelerate the colorization procedure further and extend the proposed scheme to other low-level vision tasks such as super-resolution. 
\clearpage

{\small
\bibliographystyle{ieee_fullname}
 \typeout{}
\bibliography{egbib}

\begin{thebibliography}{10}\itemsep=-1pt

\bibitem{ashikhmin2001synthesizing}
Michael Ashikhmin.
\newblock Synthesizing natural textures.
\newblock {\em SI3D}, 1:217--226, 2001.

\bibitem{ballester2001filling}
Coloma Ballester, Marcelo Bertalmio, Vicent Caselles, Guillermo Sapiro, and
  Joan Verdera.
\newblock Filling-in by joint interpolation of vector fields and gray levels.
\newblock {\em IEEE Transactions on image processing (TIP)}, 10(8):1200--1211,
  2001.

\bibitem{barnes2009patchmatch}
Connelly Barnes, Eli Shechtman, Adam Finkelstein, and Dan~B Goldman.
\newblock Patchmatch: A randomized correspondence algorithm for structural
  image editing.
\newblock In {\em ACM Transactions on Graphics (ToG)}, volume~28, page~24. ACM,
  2009.

\bibitem{chan2019everybody}
Caroline Chan, Shiry Ginosar, Tinghui Zhou, and Alexei~A Efros.
\newblock Everybody dance now.
\newblock In {\em Proceedings of the IEEE International Conference on Computer
  Vision (ICCV)}, 2019.

\bibitem{chang2015palette}
Huiwen Chang, Ohad Fried, Yiming Liu, Stephen DiVerdi, and Adam Finkelstein.
\newblock Palette-based photo recoloring.
\newblock {\em ACM Transactions on Graphics (TOG)}, 34(4):139--1, 2015.

\bibitem{cimpoi14describing}
M. Cimpoi, S. Maji, I. Kokkinos, S. Mohamed, , and A. Vedaldi.
\newblock Describing textures in the wild.
\newblock In {\em Proceedings of the IEEE Conference on Computer Vision and
  Pattern Recognition (CVPR)}, 2014.

\bibitem{deng2009imagenet}
Jia Deng, Wei Dong, Richard Socher, Li-Jia Li, Kai Li, and Li Fei-Fei.
\newblock Imagenet: A large-scale hierarchical image database.
\newblock In {\em 2009 IEEE Conference on Computer Vision and Pattern
  Recognition (CVPR)}, 2009.

\bibitem{drori2003fragment}
Iddo Drori, Daniel Cohen-Or, and Hezy Yeshurun.
\newblock Fragment-based image completion.
\newblock In {\em ACM Transactions on Graphics (TOG)}, volume~22, pages
  303--312. ACM, 2003.

\bibitem{esedoglu2002digital}
Selim Esedoglu and Jianhong Shen.
\newblock Digital inpainting based on the mumford--shah--euler image model.
\newblock {\em European Journal of Applied Mathematics}, 13(4):353--370, 2002.

\bibitem{gastal2011domain}
Eduardo~SL Gastal and Manuel~M Oliveira.
\newblock Domain transform for edge-aware image and video processing.
\newblock In {\em ACM SIGGRAPH 2011 papers}, pages 1--12. 2011.

\bibitem{he2017neural}
Mingming He, Jing Liao, Lu Yuan, and Pedro~V Sander.
\newblock Neural color transfer between images.
\newblock {\em arXiv preprint arXiv:1710.00756}, 2017.

\bibitem{hertzmann2001image}
Aaron Hertzmann, Charles~E Jacobs, Nuria Oliver, Brian Curless, and David~H
  Salesin.
\newblock Image analogies.
\newblock In {\em Conference on Computer graphics and interactive techniques},
  2001.

\bibitem{huang2005adaptive}
Yi-Chin Huang, Yi-Shin Tung, Jun-Cheng Chen, Sung-Wen Wang, and Ja-Ling Wu.
\newblock An adaptive edge detection based colorization algorithm and its
  applications.
\newblock In {\em Proceedings of ACM international conference on Multimedia
  (MM)}, 2005.

\bibitem{iizuka2017globally}
Satoshi Iizuka, Edgar Simo-Serra, and Hiroshi Ishikawa.
\newblock Globally and locally consistent image completion.
\newblock {\em ACM Transactions on Graphics (ToG)}, 36(4):107, 2017.

\bibitem{karras2017progressive}
Tero Karras, Timo Aila, Samuli Laine, and Jaakko Lehtinen.
\newblock Progressive growing of {GAN}s for improved quality, stability, and
  variation.
\newblock In {\em International Conference on Learning Representations (ICLR)},
  2018.

\bibitem{levin2004colorization}
Anat Levin, Dani Lischinski, and Yair Weiss.
\newblock Colorization using optimization.
\newblock In {\em ACM SIGGRAPH}. 2004.

\bibitem{li2019progressive}
Jingyuan Li, Fengxiang He, Lefei Zhang, Bo Du, and Dacheng Tao.
\newblock Progressive reconstruction of visual structure for image inpainting.
\newblock In {\em Proceedings of the IEEE International Conference on Computer
  Vision (ICCV)}, 2019.

\bibitem{li2015image}
Xujie Li, Hanli Zhao, Guizhi Nie, and Hui Huang.
\newblock Image recoloring using geodesic distance based color harmonization.
\newblock {\em Computational Visual Media}, 1(2):143--155, 2015.

\bibitem{liu2018image}
Guilin Liu, Fitsum~A Reda, Kevin~J Shih, Ting-Chun Wang, Andrew Tao, and Bryan
  Catanzaro.
\newblock Image inpainting for irregular holes using partial convolutions.
\newblock In {\em Proceedings of the European Conference on Computer Vision
  (ECCV)}, 2018.

\bibitem{nazeri2019edgeconnect}
Kamyar Nazeri, Eric Ng, Tony Joseph, Faisal Qureshi, and Mehran Ebrahimi.
\newblock Edgeconnect: Generative image inpainting with adversarial edge
  learning.
\newblock In {\em Workshop on the IEEE International Conference on Computer
  Vision (ICCVW)}, 2019.

\bibitem{pathak2016context}
Deepak Pathak, Philipp Krahenbuhl, Jeff Donahue, Trevor Darrell, and Alexei~A
  Efros.
\newblock Context encoders: Feature learning by inpainting.
\newblock In {\em Proceedings of the IEEE conference on Computer Vision and
  Pattern Recognition (CVPR)}, 2016.

\bibitem{qu2006manga}
Yingge Qu, Tien-Tsin Wong, and Pheng-Ann Heng.
\newblock Manga colorization.
\newblock {\em ACM Transactions on Graphics (TOG)}, 25(3):1214--1220, 2006.

\bibitem{reinhard2001color}
Erik Reinhard, Michael Adhikhmin, Bruce Gooch, and Peter Shirley.
\newblock Color transfer between images.
\newblock {\em IEEE Computer graphics and applications}, 21(5):34--41, 2001.

\bibitem{ren2019structureflow}
Yurui Ren, Xiaoming Yu, Ruonan Zhang, Thomas~H Li, Shan Liu, and Ge Li.
\newblock Structureflow: Image inpainting via structure-aware appearance flow.
\newblock In {\em Proceedings of the IEEE International Conference on Computer
  Vision (ICCV)}, 2019.

\bibitem{shaham2019singan}
Tamar~Rott Shaham, Tali Dekel, and Tomer Michaeli.
\newblock Singan: Learning a generative model from a single natural image.
\newblock In {\em Proceedings of the IEEE International Conference on Computer
  Vision (ICCV)}, 2019.

\bibitem{shan2008fast}
Qi Shan, Zhaorong Li, Jiaya Jia, and Chi-Keung Tang.
\newblock Fast image/video upsampling.
\newblock {\em ACM Transactions on Graphics (TOG)}, 27(5):1--7, 2008.

\bibitem{shocher2019ingan}
Assaf Shocher, Shai Bagon, Phillip Isola, and Michal Irani.
\newblock Ingan: Capturing and remapping the “dna” of a natural image.
\newblock In {\em Proceedings of the IEEE International Conference on Computer
  Vision (ICCV)}, 2019.

\bibitem{shocher2018zero}
Assaf Shocher, Nadav Cohen, and Michal Irani.
\newblock “zero-shot” super-resolution using deep internal learning.
\newblock In {\em Proceedings of the IEEE Conference on Computer Vision and
  Pattern Recognition (CVPR)}, 2018.

\bibitem{song2018spg}
Yuhang Song, Chao Yang, Yeji Shen, Peng Wang, Qin Huang, and C-C~Jay Kuo.
\newblock Spg-net: Segmentation prediction and guidance network for image
  inpainting.
\newblock In {\em British Machine Vision Conference (BMVC)}, 2018.

\bibitem{sun2008image}
Jian Sun, Zongben Xu, and Heung-Yeung Shum.
\newblock Image super-resolution using gradient profile prior.
\newblock In {\em IEEE Conference on Computer Vision and Pattern Recognition
  (CVPR)}, pages 1--8. IEEE, 2008.

\bibitem{ulyanov2018deep}
Dmitry Ulyanov, Andrea Vedaldi, and Victor Lempitsky.
\newblock Deep image prior.
\newblock In {\em Proceedings of the IEEE Conference on Computer Vision and
  Pattern Recognition (CVPR)}, 2018.

\bibitem{wang2018image}
Yi Wang, Xin Tao, Xiaojuan Qi, Xiaoyong Shen, and Jiaya Jia.
\newblock Image inpainting via generative multi-column convolutional neural
  networks.
\newblock In {\em Advances in Neural Information Processing Systems (NeurIPS)},
  2018.

\bibitem{wang2004image}
Zhou Wang, Alan~C Bovik, Hamid~R Sheikh, and Eero~P Simoncelli.
\newblock Image quality assessment: from error visibility to structural
  similarity.
\newblock {\em IEEE Transactions on image processing (TIP)}, 13(4):600--612,
  2004.

\bibitem{xiong2019foreground}
Wei Xiong, Jiahui Yu, Zhe Lin, Jimei Yang, Xin Lu, Connelly Barnes, and Jiebo
  Luo.
\newblock Foreground-aware image inpainting.
\newblock In {\em Proceedings of the IEEE conference on Computer Vision and
  Pattern Recognition (CVPR)}, 2019.

\bibitem{yatziv2006fast}
Liron Yatziv and Guillermo Sapiro.
\newblock Fast image and video colorization using chrominance blending.
\newblock {\em IEEE Transactions on image processing (TIP)}, 15(5):1120--1129,
  2006.

\bibitem{yi2020contextual}
Zili Yi, Qiang Tang, Shekoofeh Azizi, Daesik Jang, and Zhan Xu.
\newblock Contextual residual aggregation for ultra high-resolution image
  inpainting.
\newblock In {\em IEEE Conference on Computer Vision and Pattern Recognition
  (CVPR)}, pages 7508--7517, 2020.

\bibitem{yu2018generative}
Jiahui Yu, Zhe Lin, Jimei Yang, Xiaohui Shen, Xin Lu, and Thomas~S Huang.
\newblock Generative image inpainting with contextual attention.
\newblock In {\em Proceedings of the IEEE Conference on Computer Vision and
  Pattern Recognition (CVPR)}, 2018.

\bibitem{yu2019free}
Jiahui Yu, Zhe Lin, Jimei Yang, Xiaohui Shen, Xin Lu, and Thomas~S Huang.
\newblock Free-form image inpainting with gated convolution.
\newblock In {\em Proceedings of the IEEE International Conference on Computer
  Vision (ICCV)}, 2019.

\bibitem{zeng2020high}
Yu Zeng, Zhe Lin, Jimei Yang, Jianming Zhang, Eli Shechtman, and Huchuan Lu.
\newblock High-resolution image inpainting with iterative confidence feedback
  and guided upsampling.
\newblock In {\em European Conference on Computer Vision (ECCV)}, pages 1--17.
  Springer, 2020.

\bibitem{zhang2018perceptual}
Richard Zhang, Phillip Isola, Alexei~A Efros, Eli Shechtman, and Oliver Wang.
\newblock The unreasonable effectiveness of deep features as a perceptual
  metric.
\newblock In {\em Proceedings of the IEEE Conference on Computer Vision and
  Pattern Recognition (CVPR)}, 2018.

\bibitem{zhang2017real}
Richard Zhang, Jun-Yan Zhu, Phillip Isola, Xinyang Geng, Angela~S Lin, Tianhe
  Yu, and Alexei~A Efros.
\newblock Real-time user-guided image colorization with learned deep priors.
\newblock {\em ACM Transactions on Graphics (TOG)}, 9(4), 2017.

\bibitem{zhou2017places}
Bolei Zhou, Agata Lapedriza, Aditya Khosla, Aude Oliva, and Antonio Torralba.
\newblock Places: A 10 million image database for scene recognition.
\newblock {\em IEEE Transactions on Pattern Analysis and Machine Intelligence
  (PAMI)}, 2017.

\bibitem{zhou2018non}
Yang Zhou, Zhen Zhu, Xiang Bai, Dani Lischinski, Daniel Cohen-Or, and Hui
  Huang.
\newblock Non-stationary texture synthesis by adversarial expansion.
\newblock {\em ACM Transactions on Graphics (ToG)}, 37(4):49:1--49:13, 2018.

\end{thebibliography}
}
\clearpage

\appendix 
\section*{\LARGE Appendix}
\section{Implementation Details}
After experimenting with different network architectures (U-net, ResNet, SkipNet) for our progressive internal colorization network, we adopted a ResNet-like architecture  since it is the fastest among all the models that generate high-quality results. The number of feature channels is set to 32, and the filter size is 3. The network is composed of 9 two-layer residual blocks, 1 input Conv layer, and 1 output Conv layer. The input channel number of the first generator is 1 (gray only), and for other generators is 4 (gray+RGB). The output channel for each generator is 3 (RGB). Note that all our experiments are conducted in the RGB color-space instead of Lab color-space, so the luminance of the input image can possibly change slightly.  BatchNorm, reflection padding, and LeakyReLU are adopted. \\

We choose the pyramid height for our progressive color propagation based on the image size and image content. The default pyramid height  is set to 3 for Places2 images. The corresponding iteration number and learning rate at each level is [500,1000,1000] and [0.01, 0.005, 0.003]. We empirically found that with a larger learning rate at a lower level, the output contains more diverse colors, and with a lower learning rate at a higher level, the output contains more fine-grained details. In case the default configuration dose not yield a compelling colorization results, we can tune the pyramid height and learning rates for better performance.\\

\section{Feasibility of Internal Colorization}

\begin{figure*}[t]
    \centering
    \begin{tabular}{@{}c@{\hspace{0.3mm}}c@{\hspace{0.3mm}}c@{\hspace{0.3mm}}c@{\hspace{1.5mm}}c@{\hspace{0.3mm}}c@{\hspace{0.3mm}}c@{}}
  \rotatebox{90}{\small{ ~Input}}&
  \includegraphics[width=0.16\linewidth]{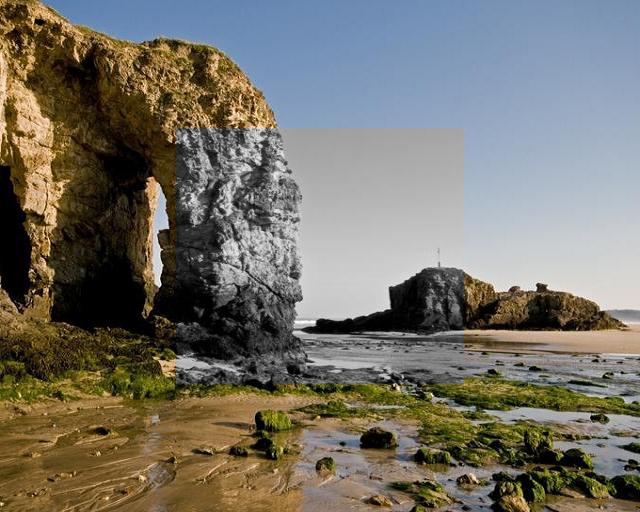}  & 
    \includegraphics[width=0.16\linewidth]{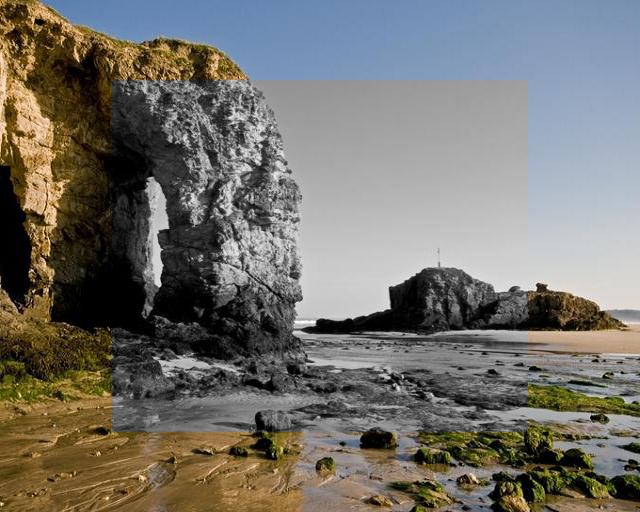} &   
    \includegraphics[width=0.16\linewidth]{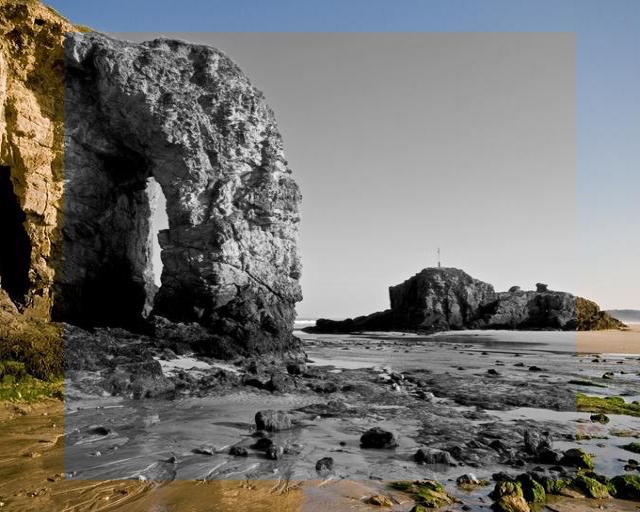}&  \includegraphics[width=0.16\linewidth]{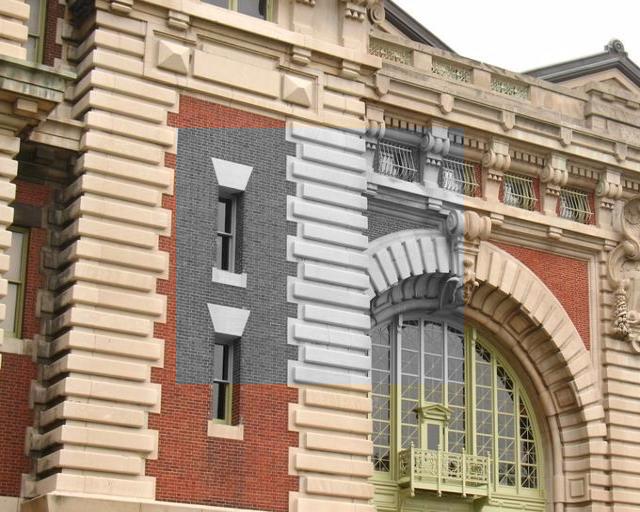}  & 
    \includegraphics[width=0.16\linewidth]{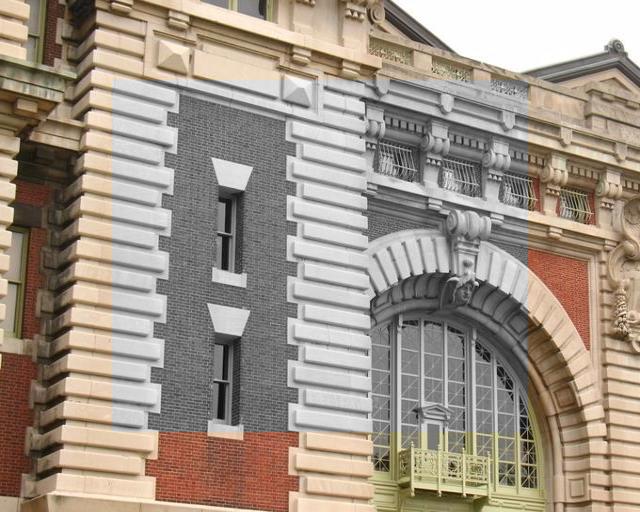} &   
    \includegraphics[width=0.16\linewidth]{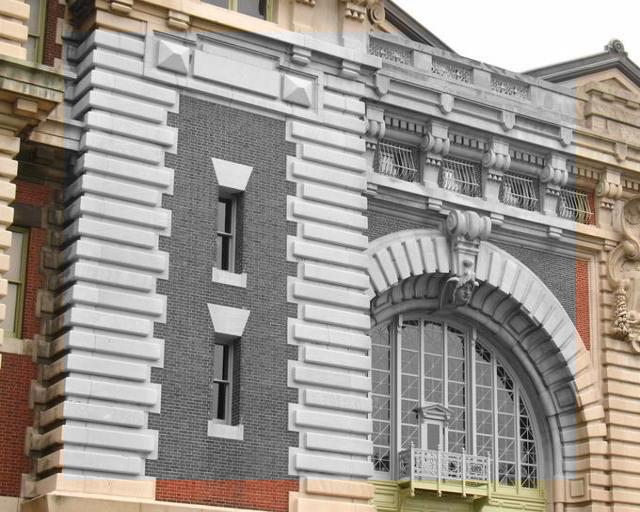}\\
    
    \rotatebox{90}{\small{ Gastal's}}&
 \includegraphics[width=0.16\linewidth]{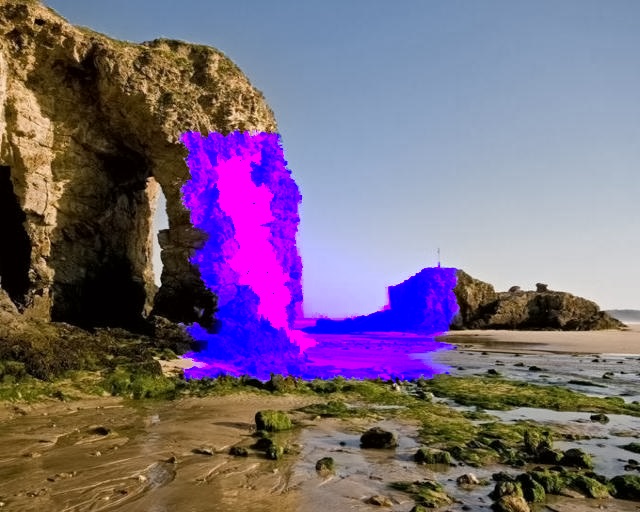}  & 
    \includegraphics[width=0.16\linewidth]{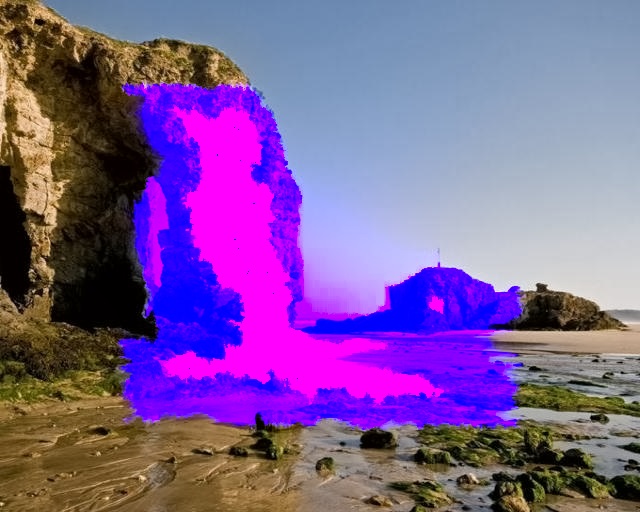} &   
    \includegraphics[width=0.16\linewidth]{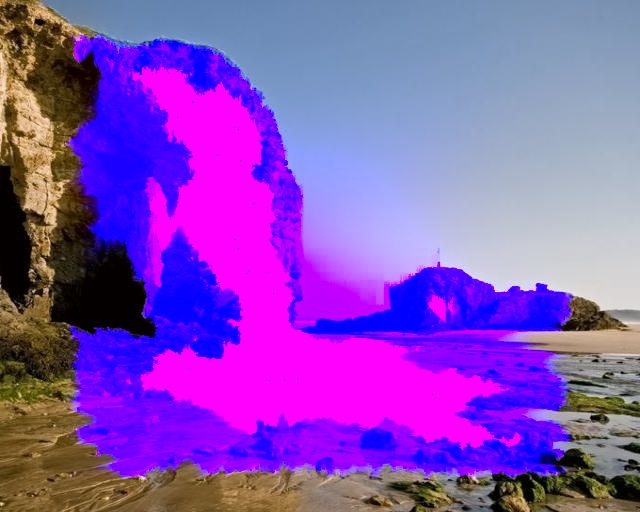}&
     \includegraphics[width=0.16\linewidth]{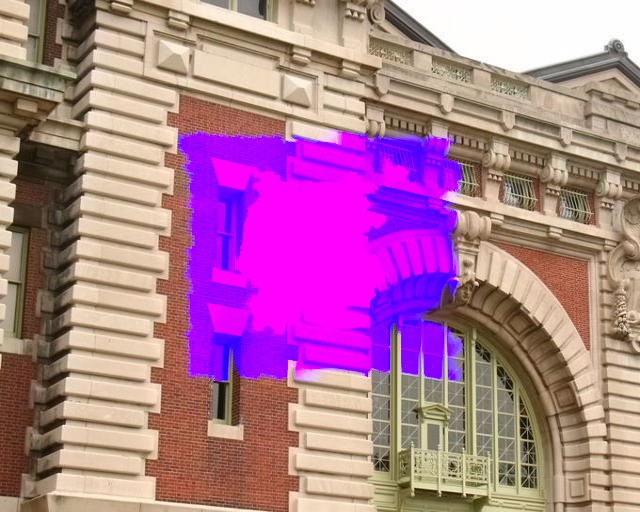}  & 
    \includegraphics[width=0.16\linewidth]{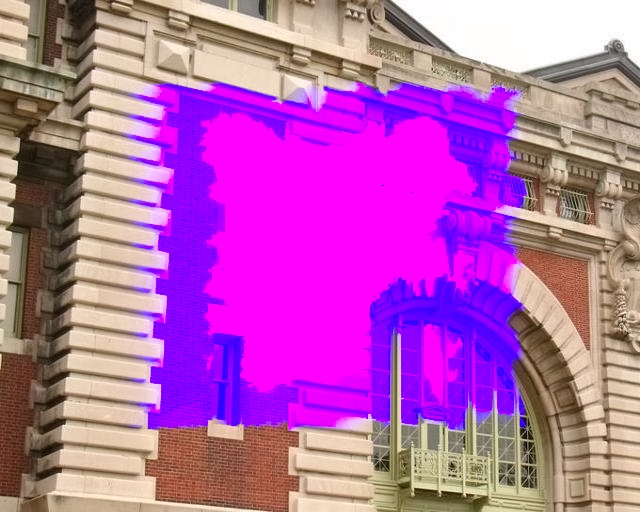} &   
    \includegraphics[width=0.16\linewidth]{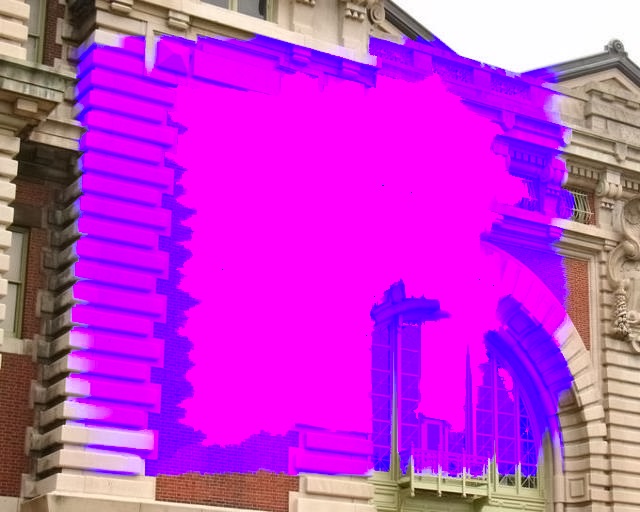}\\  

  \rotatebox{90}{ \small{ zhang's}}&
  \includegraphics[width=0.16\linewidth]{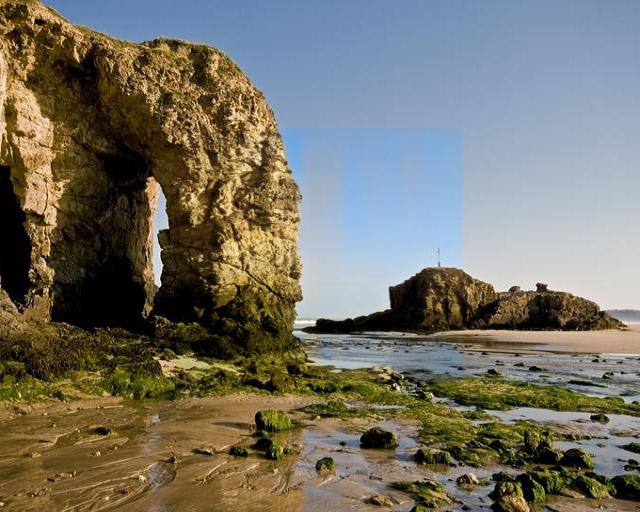}  & 
    \includegraphics[width=0.16\linewidth]{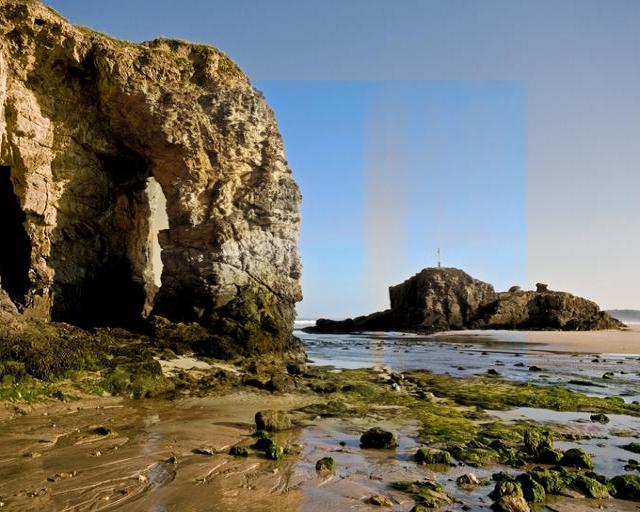} &   
    \includegraphics[width=0.16\linewidth]{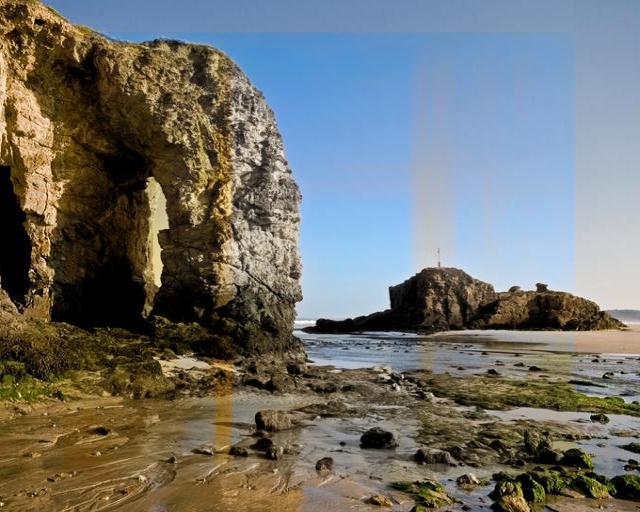}&
      \includegraphics[width=0.16\linewidth]{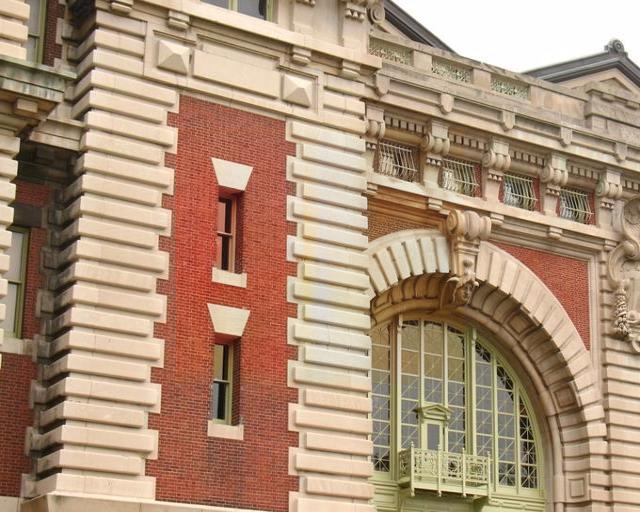}  &
    \includegraphics[width=0.16\linewidth]{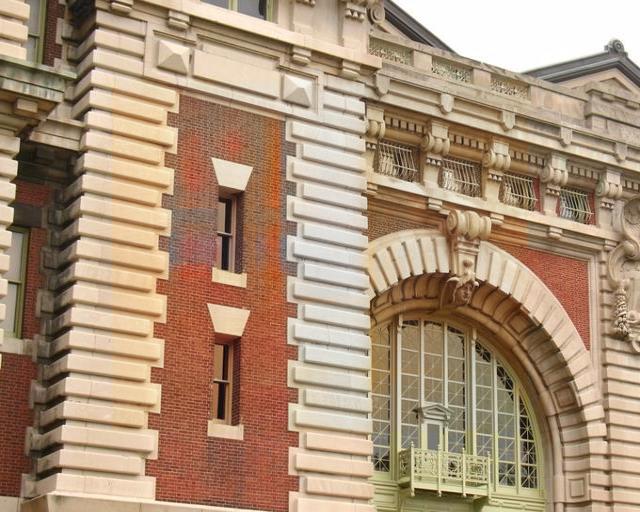} &   
    \includegraphics[width=0.16\linewidth]{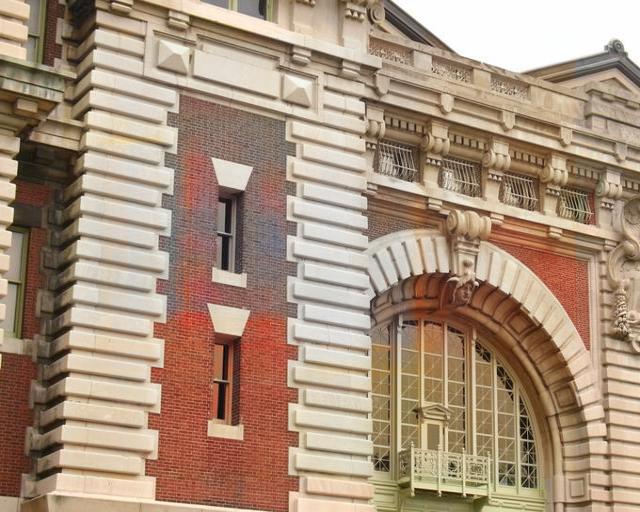}\\  

    \rotatebox{90}{\small{ levin's}}&
 \includegraphics[width=0.16\linewidth]{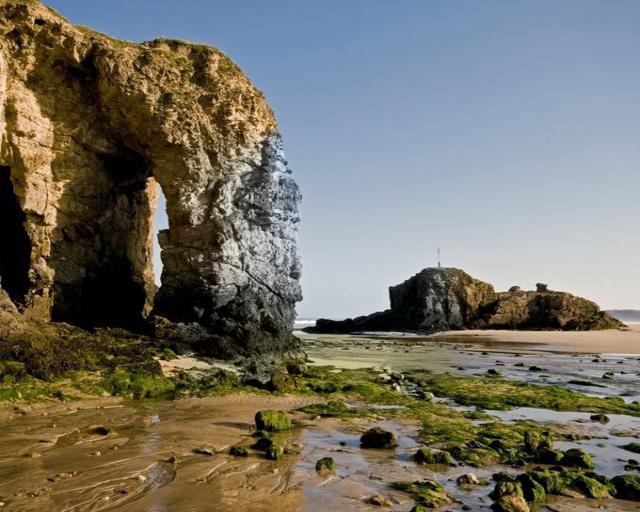}  & 
    \includegraphics[width=0.16\linewidth]{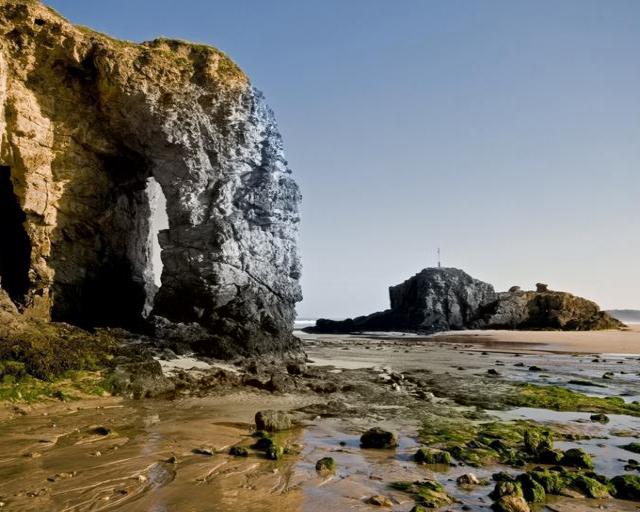} &   
    \includegraphics[width=0.16\linewidth]{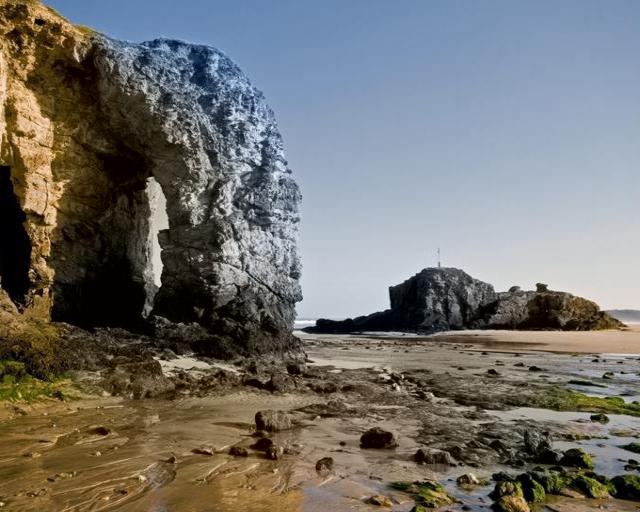}&
     \includegraphics[width=0.16\linewidth]{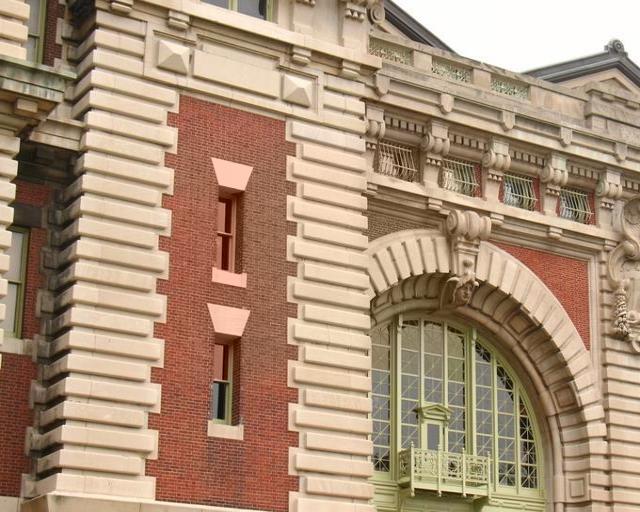}  & 
    \includegraphics[width=0.16\linewidth]{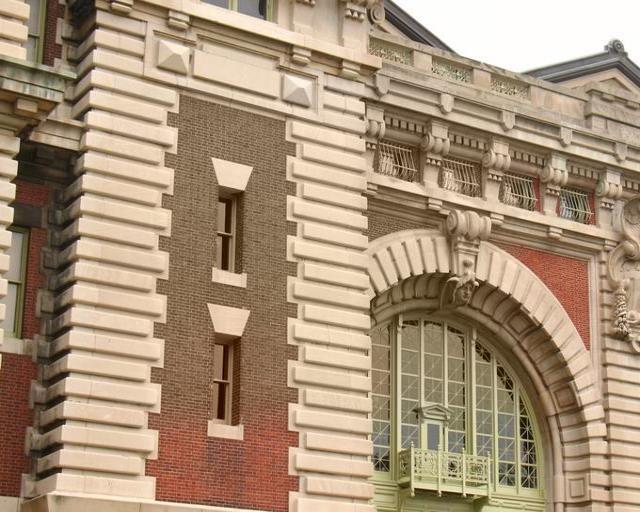} &   
    \includegraphics[width=0.16\linewidth]{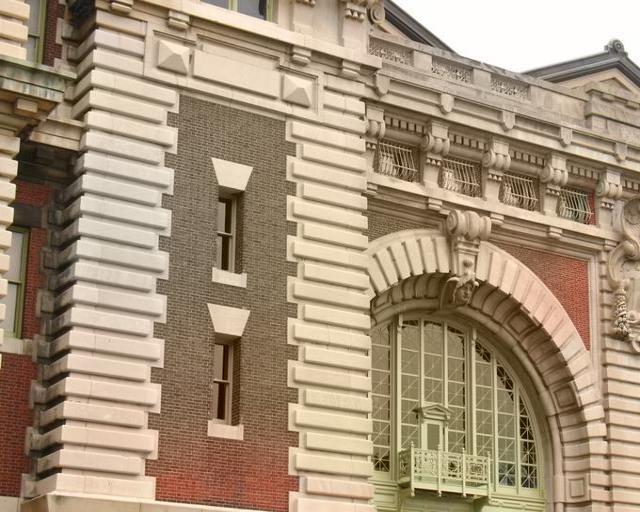}\\    
    
          \rotatebox{90}{\small{ ~Ours}}&
  \includegraphics[width=0.16\linewidth]{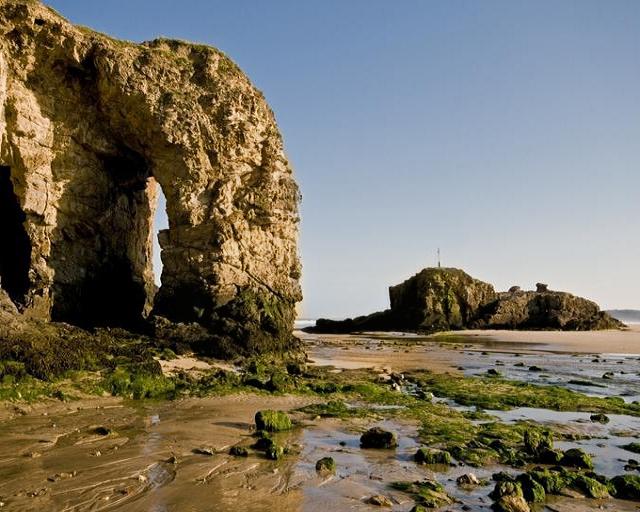}  & 
    \includegraphics[width=0.16\linewidth]{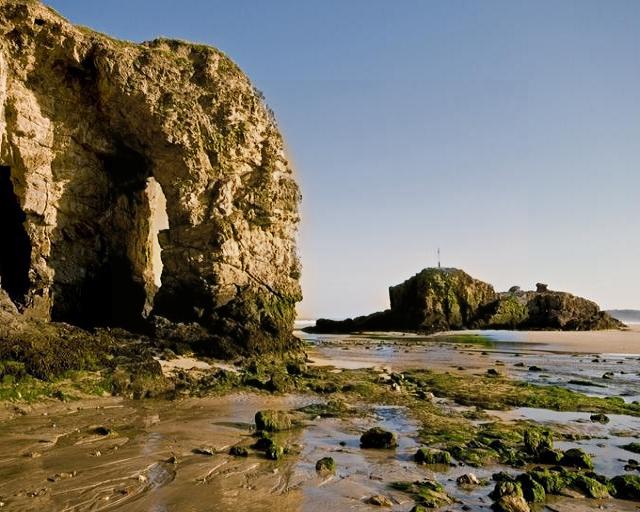} &   
    \includegraphics[width=0.16\linewidth]{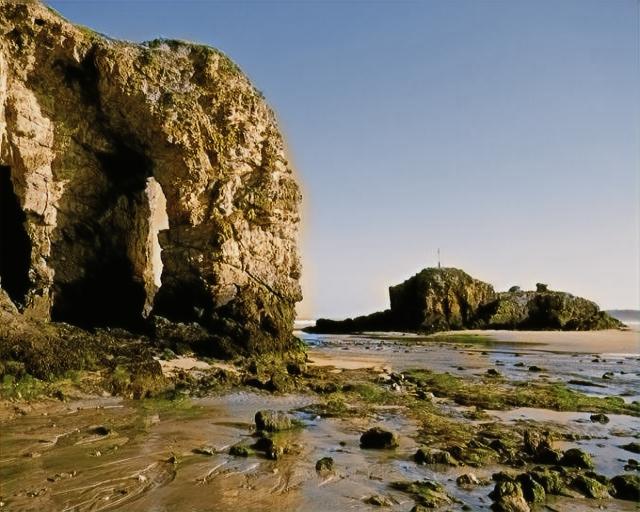}&
      \includegraphics[width=0.16\linewidth]{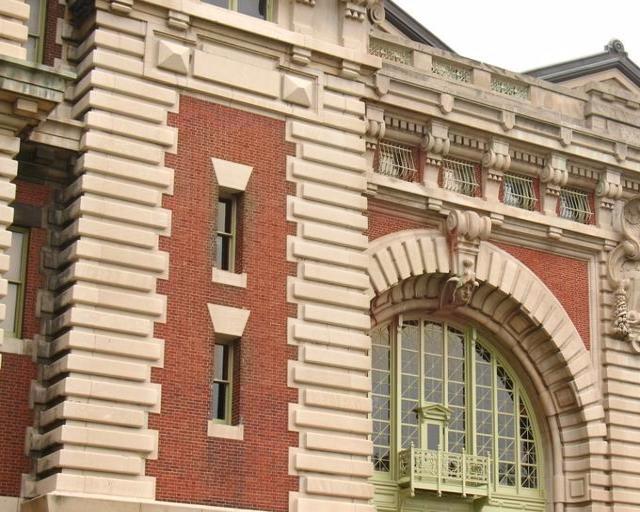}  &
    \includegraphics[width=0.16\linewidth]{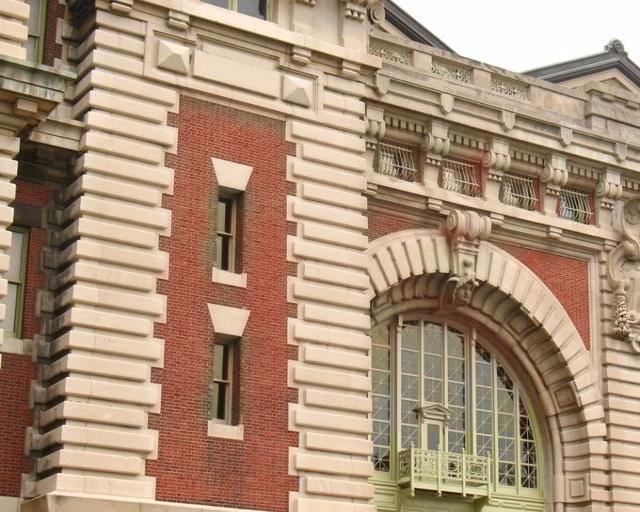} &   
    \includegraphics[width=0.16\linewidth]{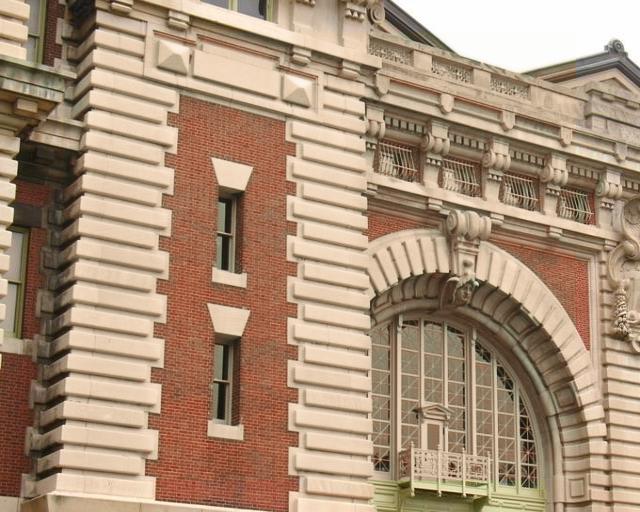}\\

\rotatebox{90}{\small{~Original}}& \includegraphics[width=0.16\linewidth]{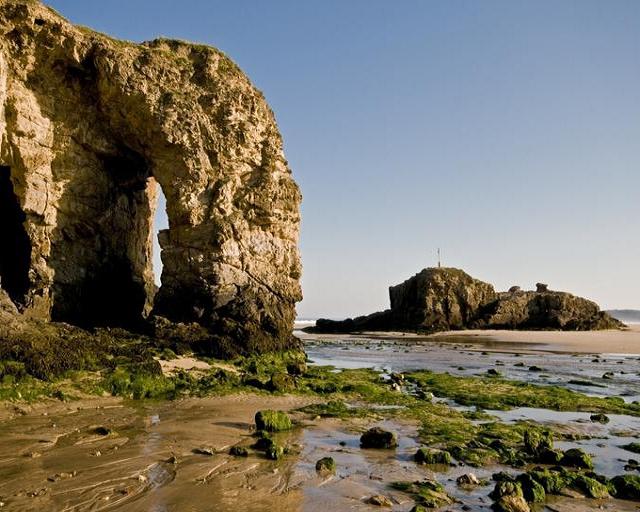}&     \includegraphics[width=0.16\linewidth]{supp/ratio/2/gt.jpg}& 
\includegraphics[width=0.16\linewidth]{supp/ratio/2/gt.jpg}&  
\includegraphics[width=0.16\linewidth]{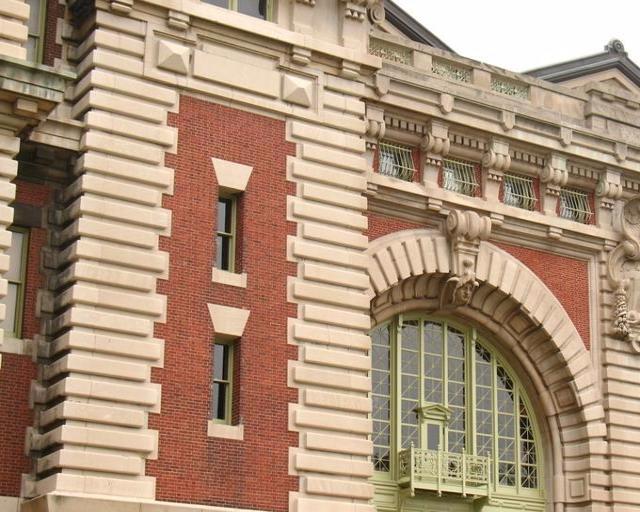}&     \includegraphics[width=0.16\linewidth]{supp/ratio/1/gt.jpg}& 
\includegraphics[width=0.16\linewidth]{supp/ratio/1/gt.jpg}\\    

 &\small{$22.5\%$}  & \small{$48.9\%$} & \small{$73.4\%$} &\small{$22.5\%$}  & \small{$48.9\%$} & \small{$73.4\%$}
    \end{tabular}
    \caption{Feasibility of our internal colorization method. We increase the mask ratio of $I_{hole}$ from $22.5\%$ to $73.4\%$ and colorize the ground-truth grayscale image with our internal colorization method. We also give colorization results by Zhang et al.~\cite{zhang2017real}, Gastal et al. ~\cite{gastal2011domain} and Levin et al.~\cite{levin2004colorization} for comparison.}
    \label{fig:mask_ratio3}
\end{figure*}

\begin{figure*}[t]
    \centering
    \begin{tabular}{c@{\hspace{0.3mm}}c@{\hspace{0.3mm}}c@{\hspace{0.3mm}}c@{\hspace{1.5mm}}c@{\hspace{0.3mm}}c@{\hspace{0.3mm}}c}

  \rotatebox{90}{\small{ ~Input}}&
  \includegraphics[width=0.16\linewidth]{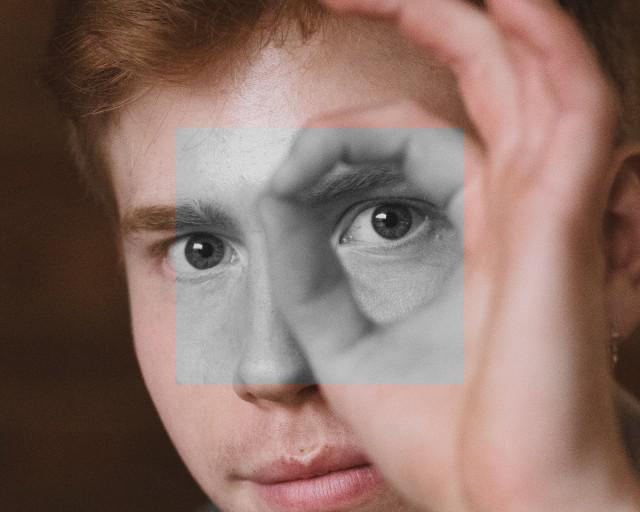}  &
    \includegraphics[width=0.16\linewidth]{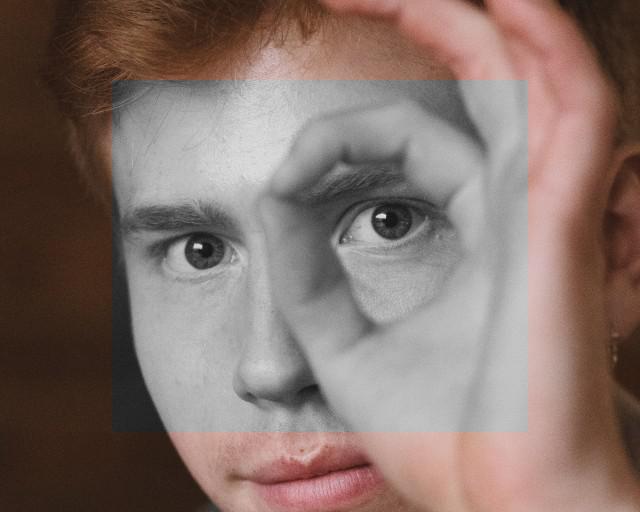} &
    \includegraphics[width=0.16\linewidth]{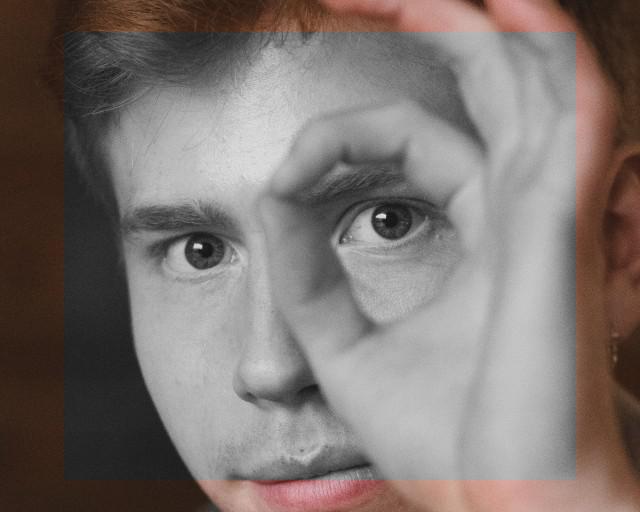}&  \includegraphics[width=0.16\linewidth]{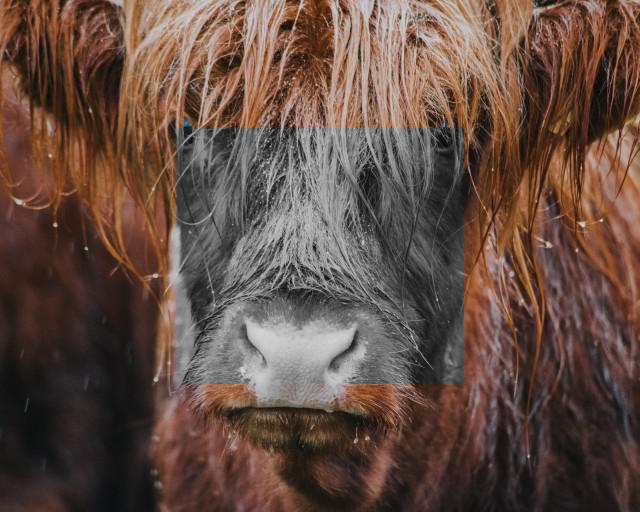}  & 
    \includegraphics[width=0.16\linewidth]{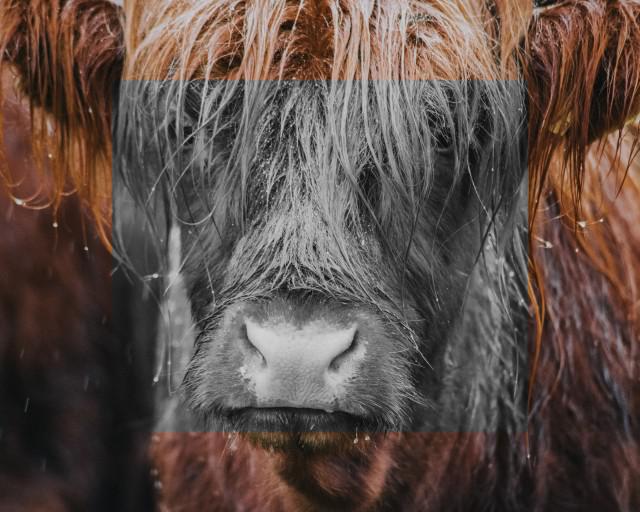} &
    \includegraphics[width=0.16\linewidth]{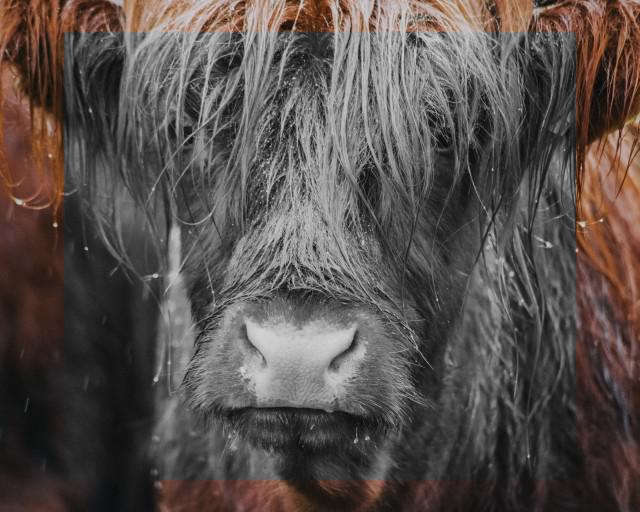}\\
    
    \rotatebox{90}{\small{ Gastal's}}&
 \includegraphics[width=0.16\linewidth]{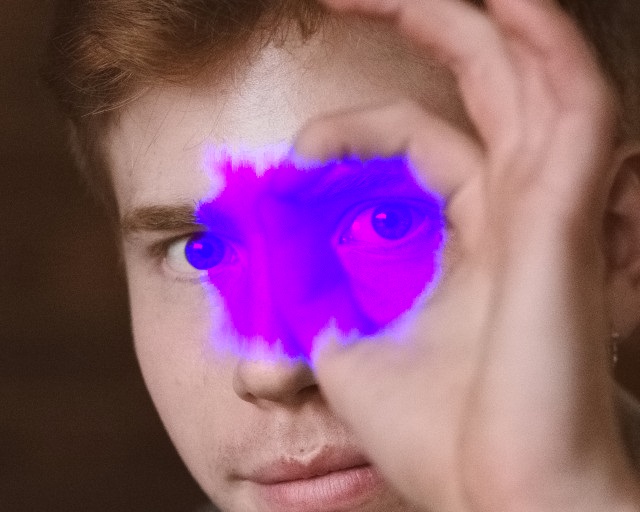}  & 
    \includegraphics[width=0.16\linewidth]{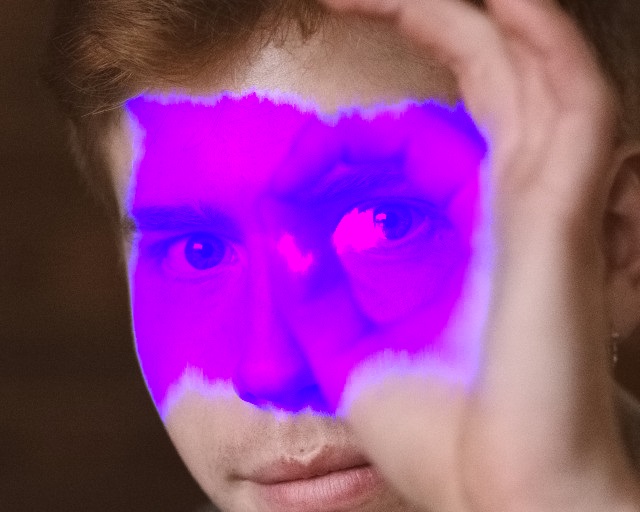} &    
    \includegraphics[width=0.16\linewidth]{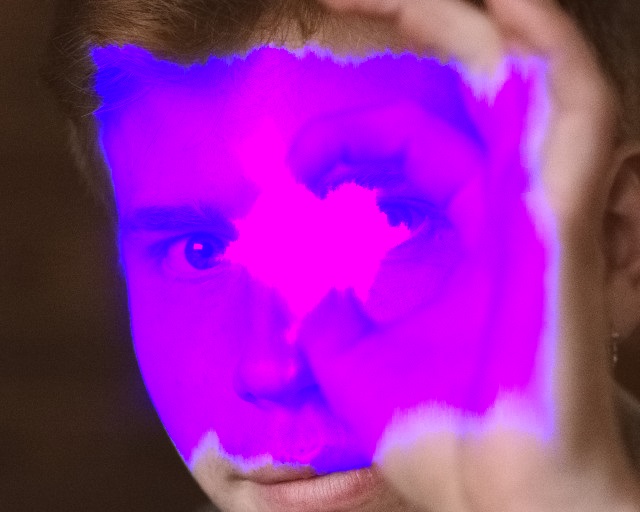}&
     \includegraphics[width=0.16\linewidth]{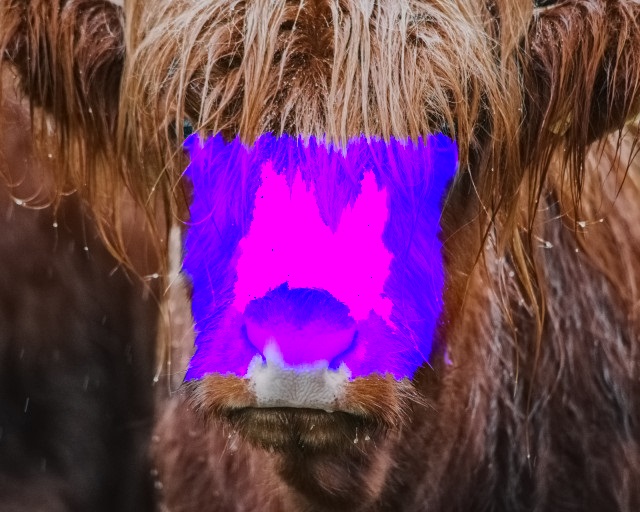}  & 
    \includegraphics[width=0.16\linewidth]{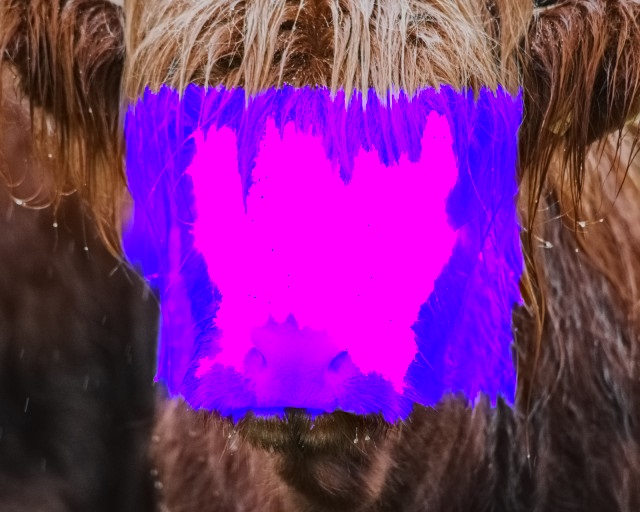} &    
    \includegraphics[width=0.16\linewidth]{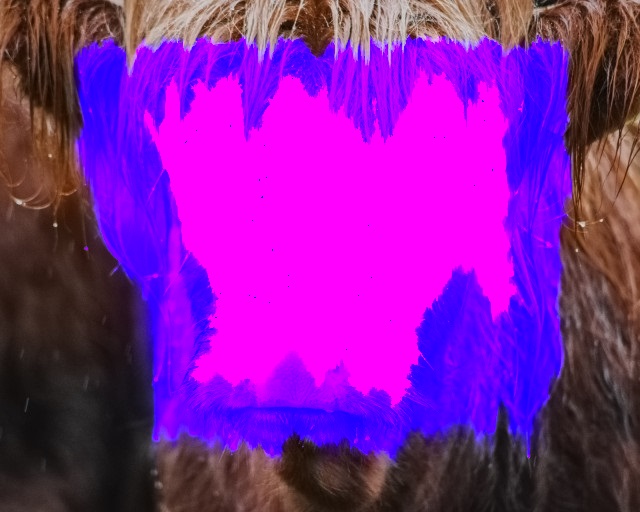}\\  

  \rotatebox{90}{ \small{ zhang's}}&
  \includegraphics[width=0.16\linewidth]{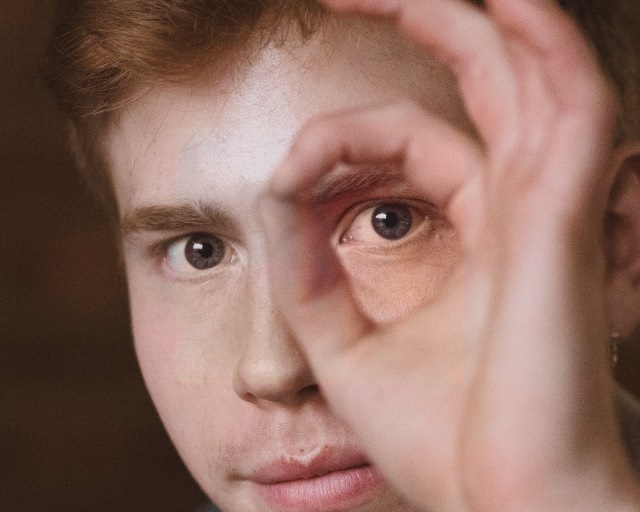}  & 
    \includegraphics[width=0.16\linewidth]{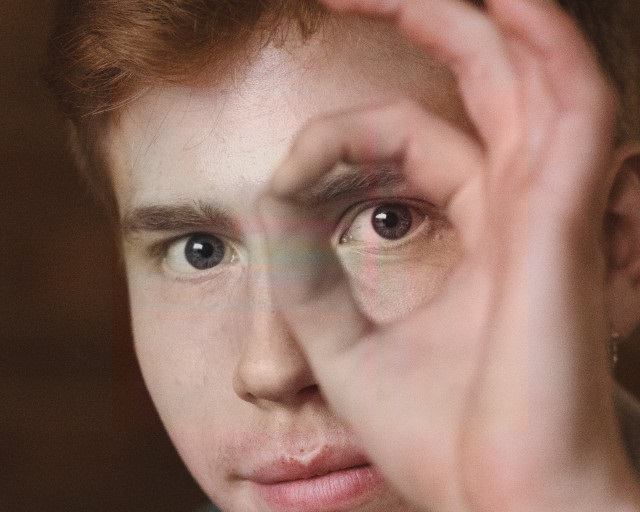} &    
    \includegraphics[width=0.16\linewidth]{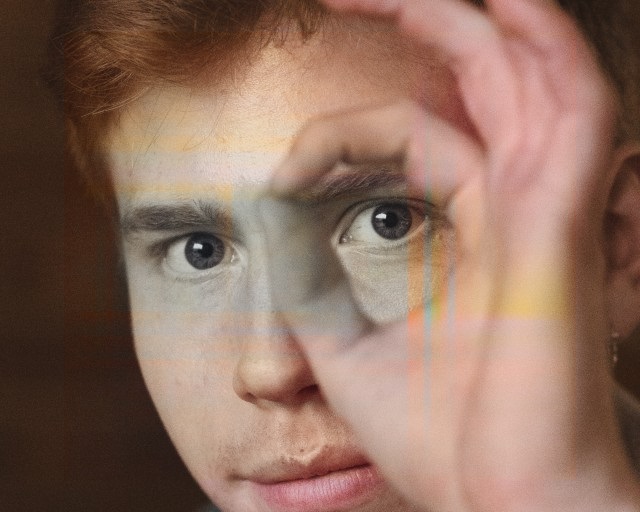}&
      \includegraphics[width=0.16\linewidth]{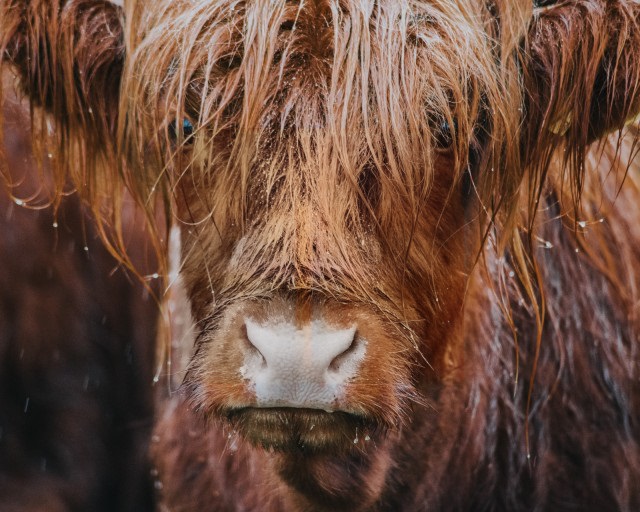}  & 
    \includegraphics[width=0.16\linewidth]{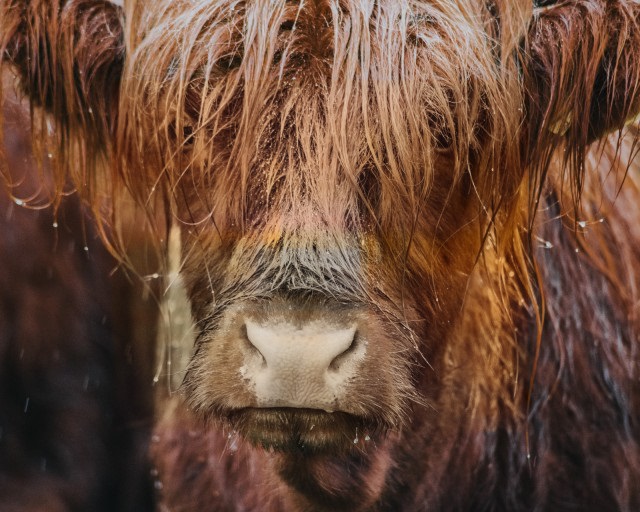} &    
    \includegraphics[width=0.16\linewidth]{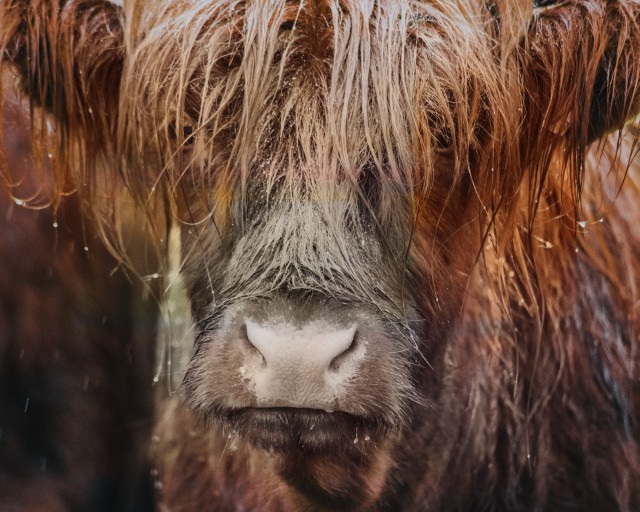}\\  

    \rotatebox{90}{\small{ levin's}}&
 \includegraphics[width=0.16\linewidth]{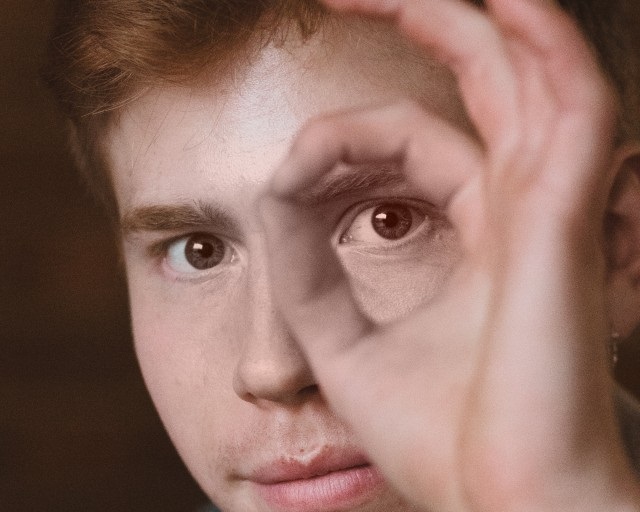}  & 
    \includegraphics[width=0.16\linewidth]{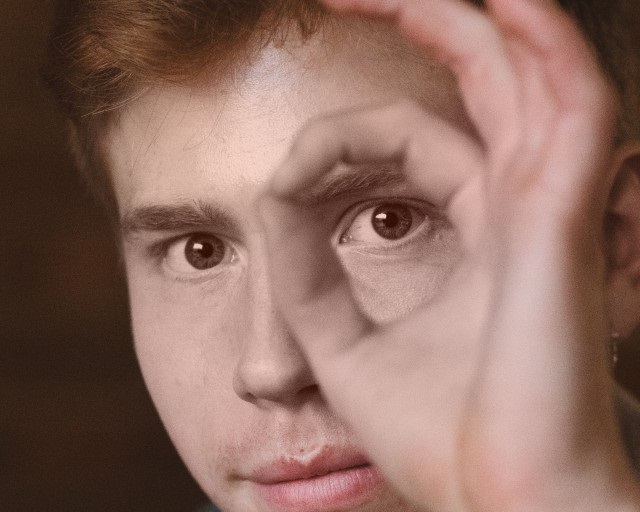} &    
    \includegraphics[width=0.16\linewidth]{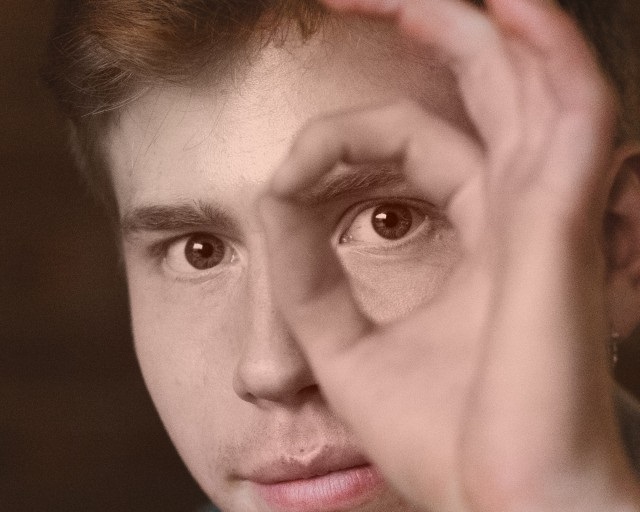}&
     \includegraphics[width=0.16\linewidth]{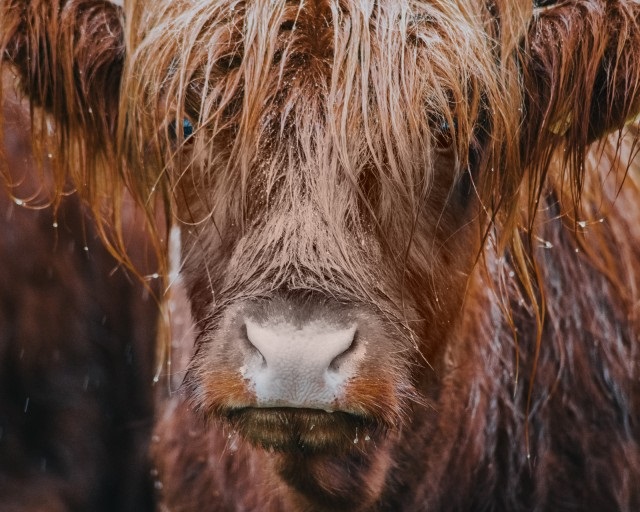}  & 
    \includegraphics[width=0.16\linewidth]{supp/ratio/4/c1.jpg} &    
    \includegraphics[width=0.16\linewidth]{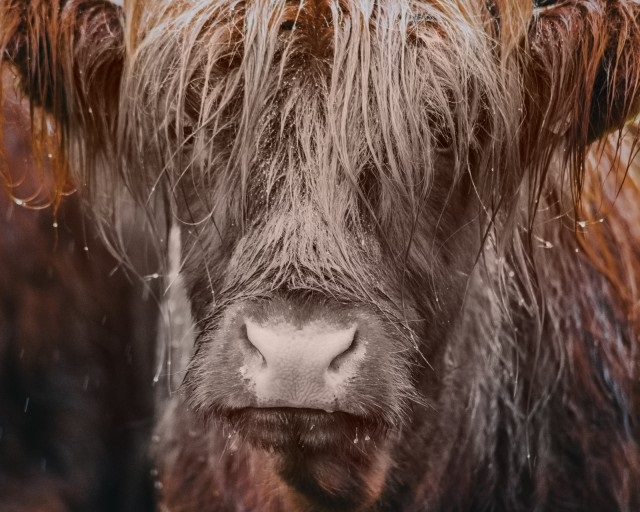}\\    
    
          \rotatebox{90}{\small{ ~Ours}}&
  \includegraphics[width=0.16\linewidth]{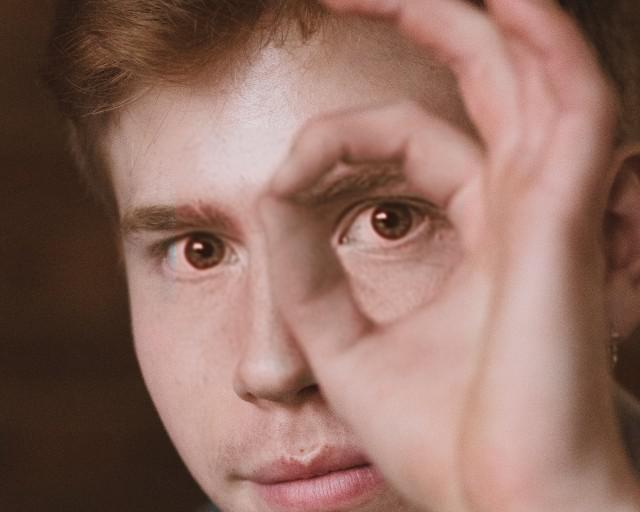}  &
    \includegraphics[width=0.16\linewidth]{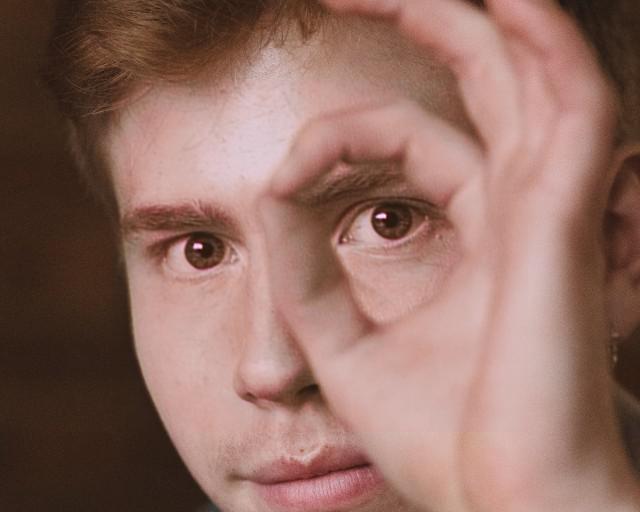} &  
    \includegraphics[width=0.16\linewidth]{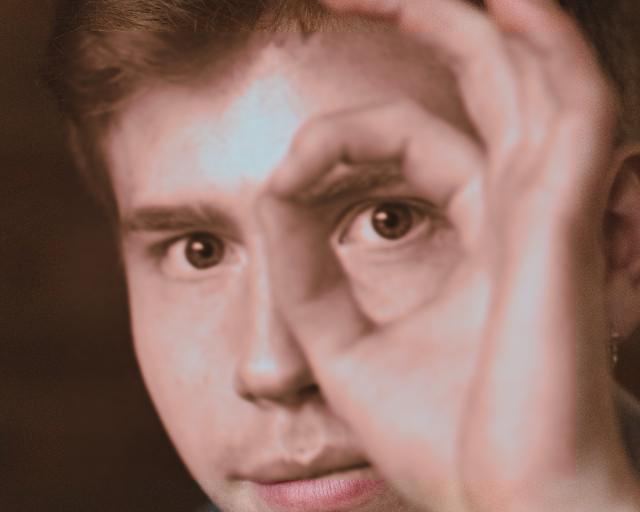}&
     \includegraphics[width=0.16\linewidth]{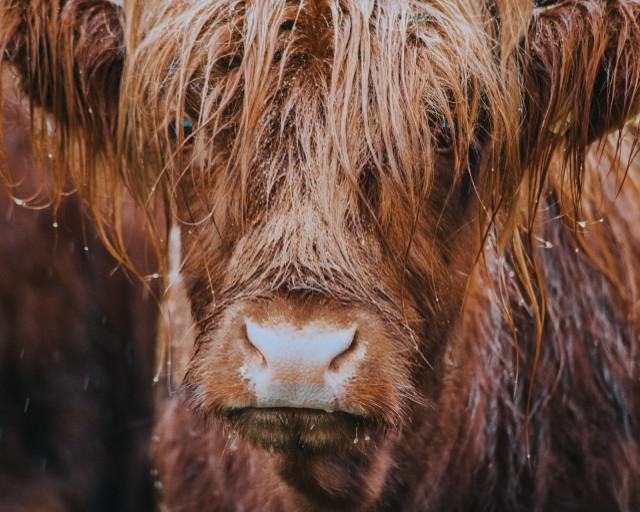}  &
    \includegraphics[width=0.16\linewidth]{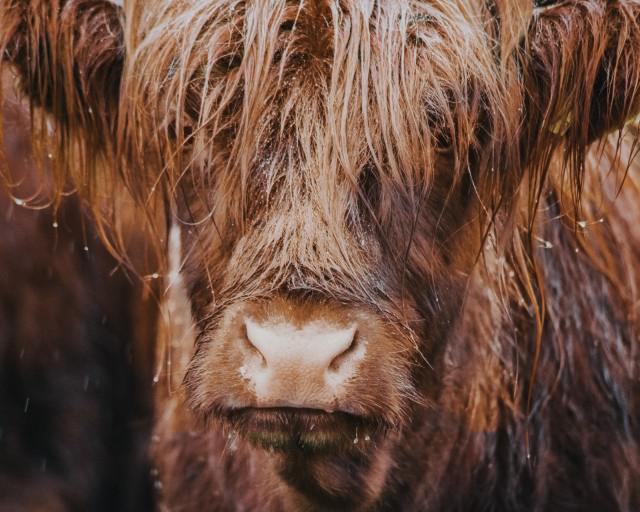} &  
    \includegraphics[width=0.16\linewidth]{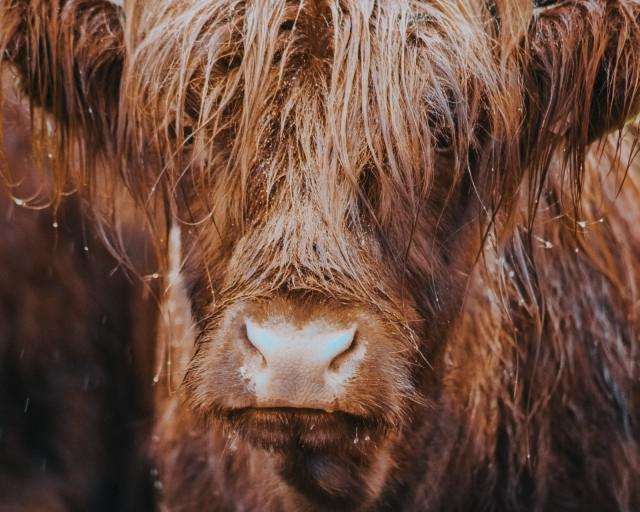}\\

\rotatebox{90}{\small{~Original}}& \includegraphics[width=0.16\linewidth]{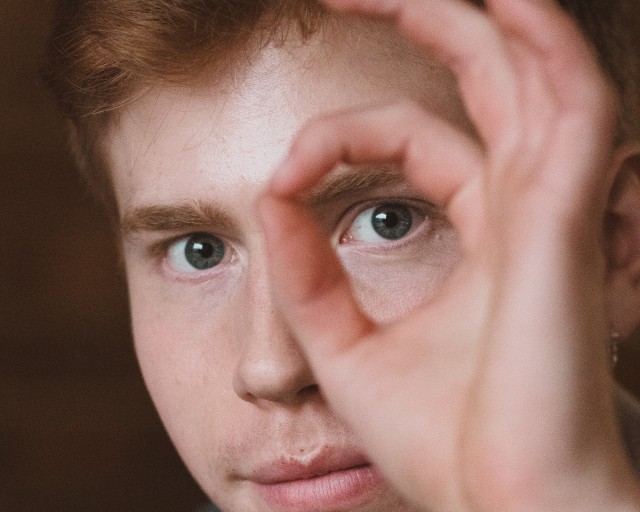}&     \includegraphics[width=0.16\linewidth]{supp/ratio/3/gt.jpg}& 
\includegraphics[width=0.16\linewidth]{supp/ratio/3/gt.jpg}&  
\includegraphics[width=0.16\linewidth]{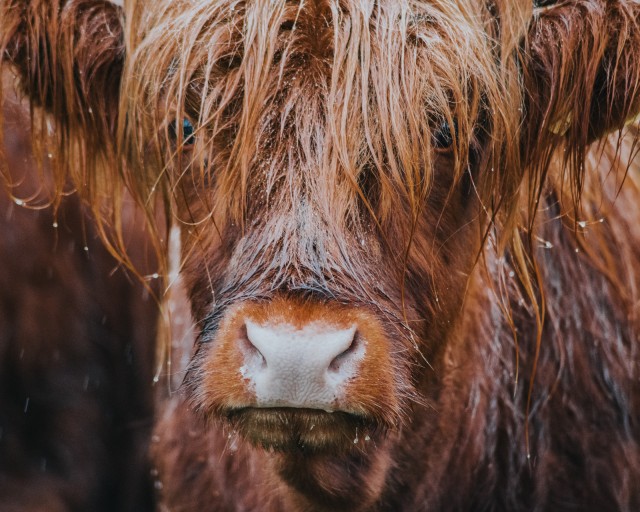}&     \includegraphics[width=0.16\linewidth]{supp/ratio/4/gt.jpg}& 
\includegraphics[width=0.16\linewidth]{supp/ratio/4/gt.jpg}\\    

 &\small{$22.5\%$}  & \small{$48.9\%$} & \small{$73.4\%$} &\small{$22.5\%$}  & \small{$48.9\%$} & \small{$73.4\%$}
    \end{tabular}
    \caption{Feasibility of our internal colorization method. We increase the mask ratio of $I_{hole}$ from $22.5\%$ to $73.4\%$ and colorize the ground-truth grayscale image with our internal colorization method. We also give colorization results by Zhang et al.~\cite{zhang2017real}, Gastal et al. ~\cite{gastal2011domain} and Levin et al.~\cite{levin2004colorization} for comparison.}
    \label{fig:mask_ratio2}
\end{figure*}

We also provide more examples to validate the feasibility of the proposed internal propagation. These examples are from different categories, including natural scenery, buildings, human faces, and animals. As shown in Fig.~\ref{fig:mask_ratio3} and Fig.~\ref{fig:mask_ratio2}, our scheme achieves stable colorization results on various images with different mask ratios.

\section{User-guided Inpainting}
\begin{figure*}[t]
\centering

\begin{tabular}{c@{\hspace{0.3mm}}c@{\hspace{0.3mm}}c@{\hspace{0.3mm}}c@{\hspace{3mm}}c@{\hspace{0.3mm}}c@{\hspace{0.3mm}}c@{\hspace{0.3mm}}c}
\\
\includegraphics[width=0.12\linewidth]{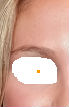}&
\includegraphics[width=0.12\linewidth]{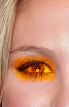}&
\includegraphics[width=0.12\linewidth]{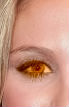}&
\includegraphics[width=0.12\linewidth]{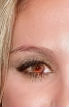}&
\includegraphics[width=0.12\linewidth]{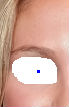}&
\includegraphics[width=0.12\linewidth]{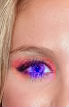}&
\includegraphics[width=0.12\linewidth]{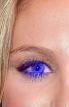}&
\includegraphics[width=0.12\linewidth]{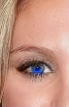}\\

\includegraphics[width=0.12\linewidth]{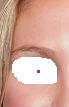}&
\includegraphics[width=0.12\linewidth]{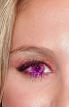}&
\includegraphics[width=0.12\linewidth]{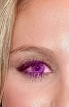}&
\includegraphics[width=0.12\linewidth]{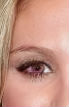}&
\includegraphics[width=0.12\linewidth]{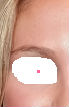}&
\includegraphics[width=0.12\linewidth]{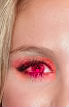}&
\includegraphics[width=0.12\linewidth]{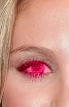}&
\includegraphics[width=0.12\linewidth]{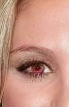}\\

\small{Hint}&\small{zhang's}&\small{levin's}&\small{Ours}&\small{Hint}&\small{zhang's}&\small{levin's}&\small{Ours}
\end{tabular} 

\caption{Examples of user-guided inpainting by our method. Users can control the color of inpainted content.}
\label{fig:color}
\end{figure*}

Users can control the color of inpainted content with our  inpainting method. We provide more details of user-guided inpainting and comparison results in Fig.~\ref{fig:color}. In this example, our model utilizes only one extra user-guided color point as a hint to generate realistic eyes with different colors. In contrast, other guided colorization methods fail to produce visually pleasing results and show obvious color bleeding artifacts in the eyes. 
 
\section{More Results}
\begin{figure*}[t]
    \centering
    \begin{tabular}{c@{\hspace{0.35mm}}c@{\hspace{0.35mm}}c@{\hspace{2mm}}c@{\hspace{0.35mm}}c@{\hspace{0.35mm}}c}
    \includegraphics[width=0.16\linewidth]{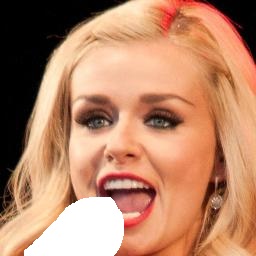}  &    \includegraphics[width=0.16\linewidth]{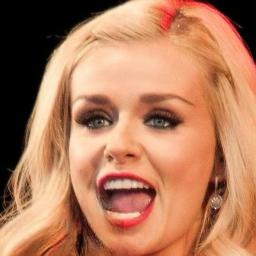}  & 
    \includegraphics[width=0.16\linewidth]{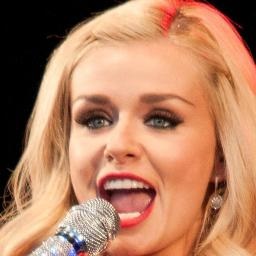} &    
    \includegraphics[width=0.16\linewidth]{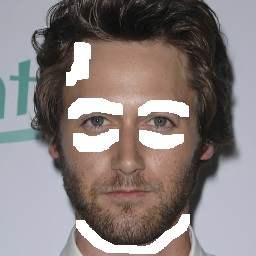}&
    \includegraphics[width=0.16\linewidth]{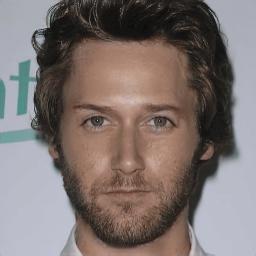}& 
    \includegraphics[width=0.16\linewidth]{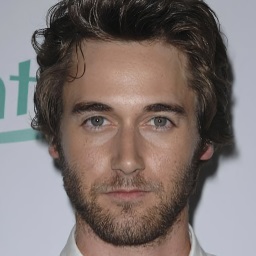} \\
    \includegraphics[width=0.16\linewidth]{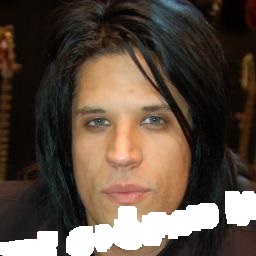}  &   
    \includegraphics[width=0.16\linewidth]{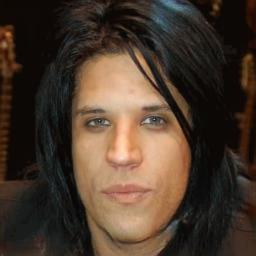}  &
    \includegraphics[width=0.16\linewidth]{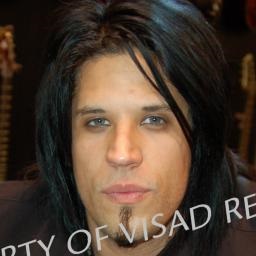} &   
    \includegraphics[width=0.16\linewidth]{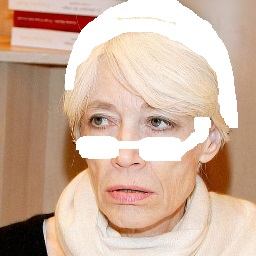}&
    \includegraphics[width=0.16\linewidth]{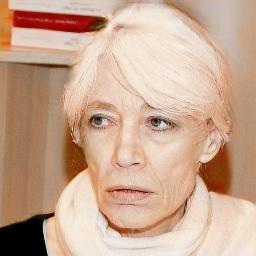}& 
    \includegraphics[width=0.16\linewidth]{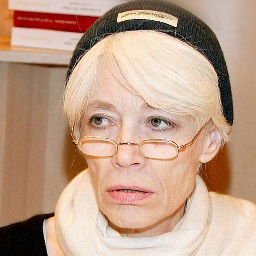} \\
    \includegraphics[width=0.16\linewidth]{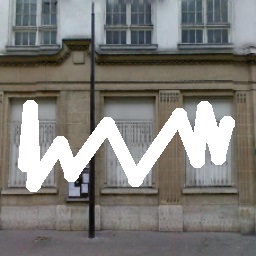}&
    \includegraphics[width=0.16\linewidth]{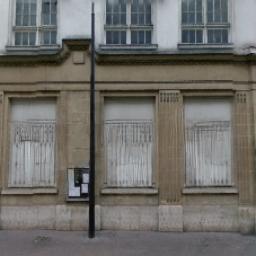}& 
    \includegraphics[width=0.16\linewidth]{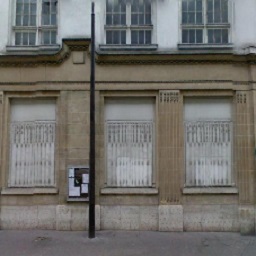} &   
    \includegraphics[width=0.16\linewidth]{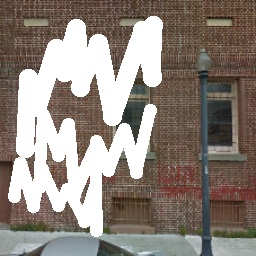}&
    \includegraphics[width=0.16\linewidth]{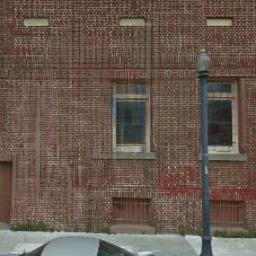}& 
    \includegraphics[width=0.16\linewidth]{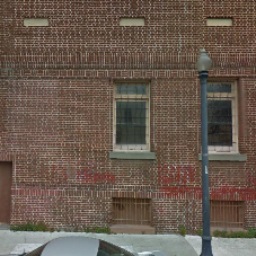} \\
    \small{Input}&\small{Ours}&\small{Original}&\small{Input}&\small{Ours}&\small{Original}
    \end{tabular}
    \caption{More results. }
    \label{fig:more}
\end{figure*}

\begin{figure*}[t]
\centering
\begin{tabular}{@{}c@{\hspace{0.4mm}}c@{\hspace{0.4mm}}c@{\hspace{0.4mm}}c@{\hspace{0.4mm}}c@{\hspace{0.4mm}}c@{\hspace{0.4mm}}c@{}}

  \rotatebox{90}{\small{ ~Input}}&
\includegraphics[width=0.13\linewidth]{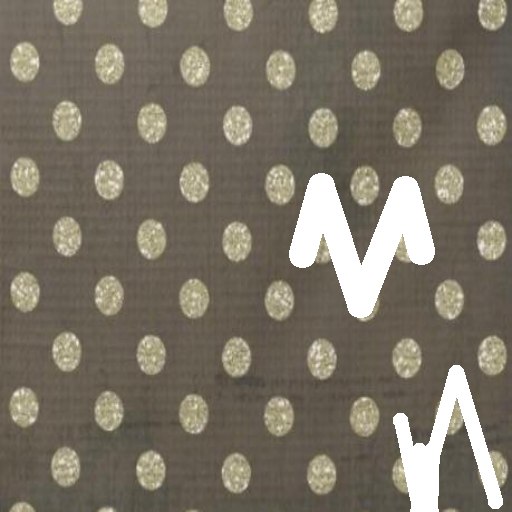}  &
\includegraphics[width=0.13\linewidth]{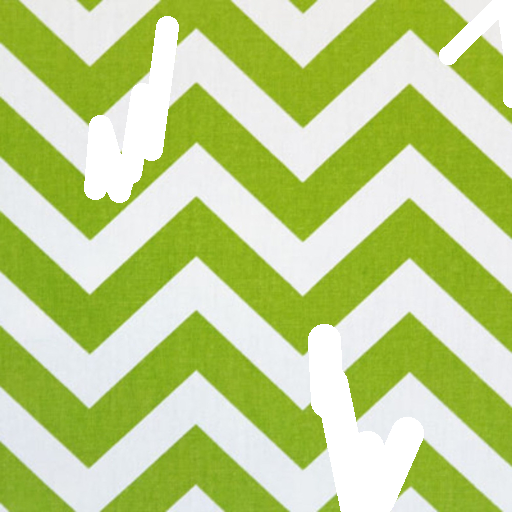}  & \includegraphics[width=0.13\linewidth]{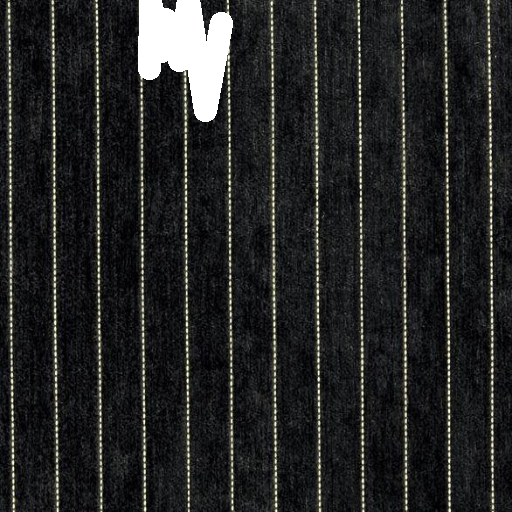}  &
\includegraphics[width=0.13\linewidth]{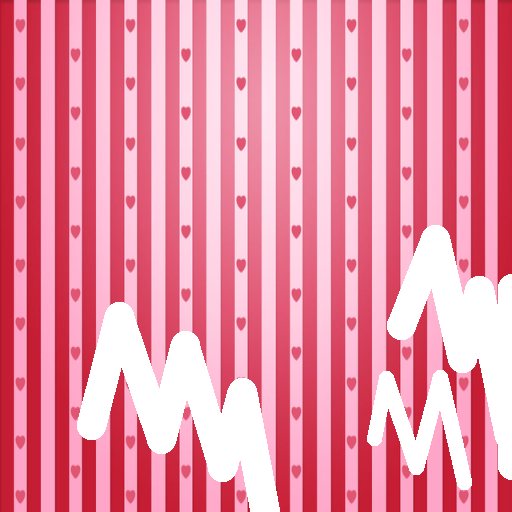}  &
\includegraphics[width=0.13\linewidth]{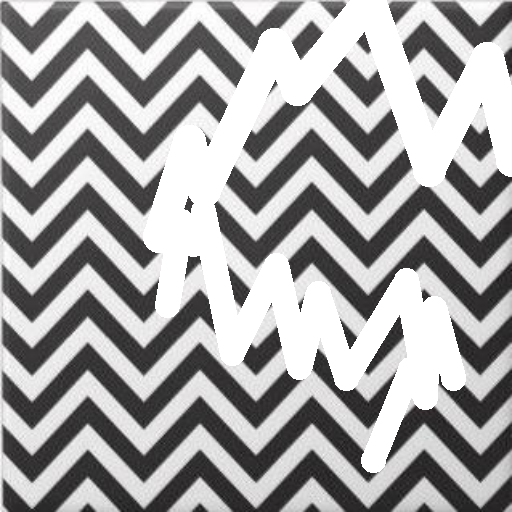}&
\includegraphics[width=0.13\linewidth]{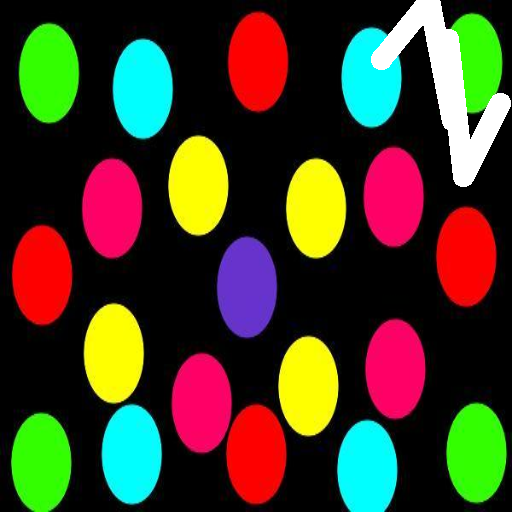}\\
    
\rotatebox{90}{\small{HiFill~\cite{yi2020contextual}}}&
\includegraphics[width=0.13\linewidth]{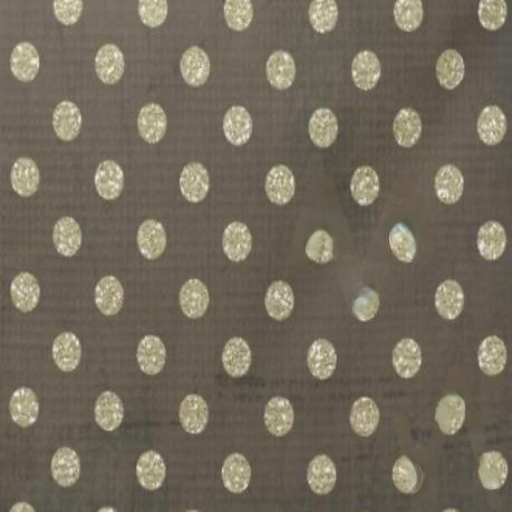} & 
\includegraphics[width=0.13\linewidth]{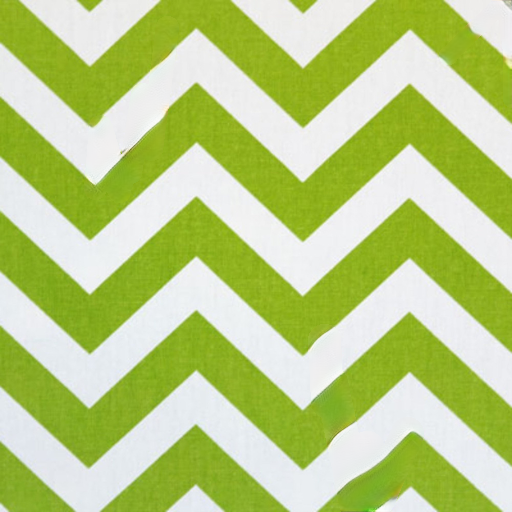} & 
\includegraphics[width=0.13\linewidth]{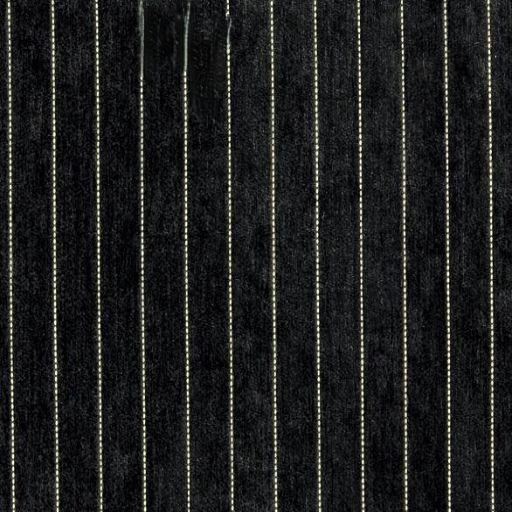} & 
\includegraphics[width=0.13\linewidth]{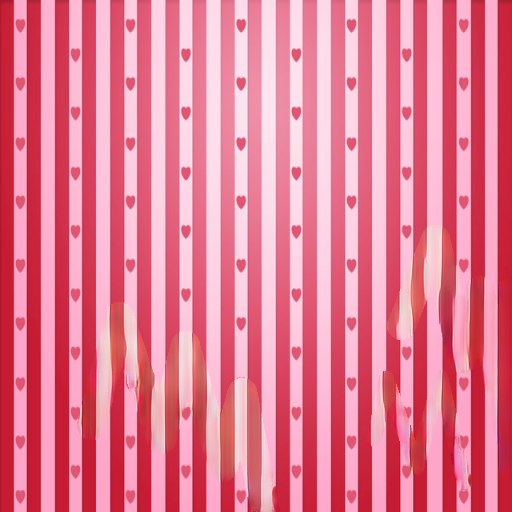} & 
\includegraphics[width=0.13\linewidth]{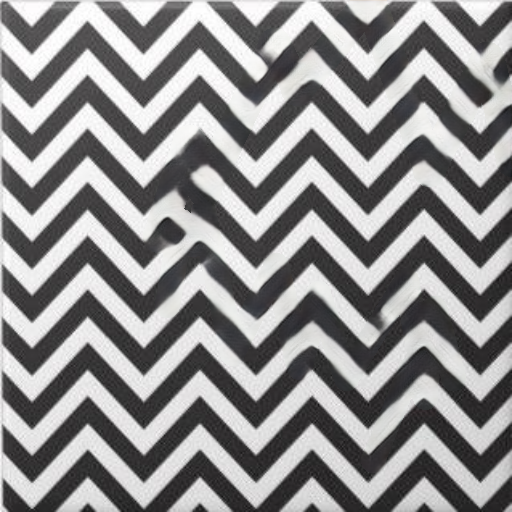} &
\includegraphics[width=0.13\linewidth]{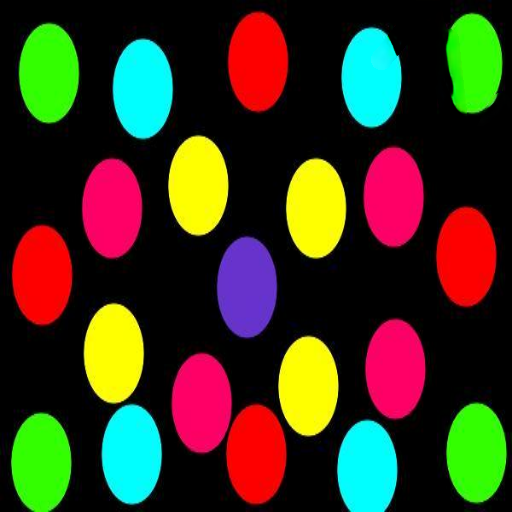}\\  

  \rotatebox{90}{ \small{ Ours }}&
\includegraphics[width=0.13\linewidth]{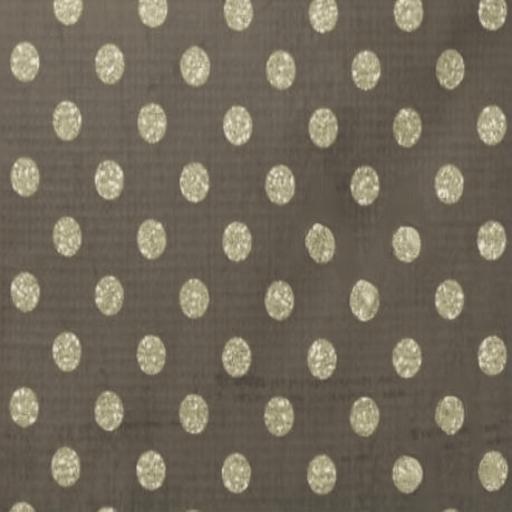} &
\includegraphics[width=0.13\linewidth]{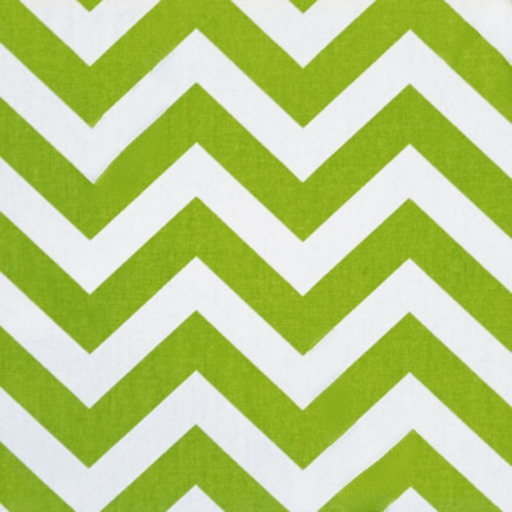} & 
\includegraphics[width=0.13\linewidth]{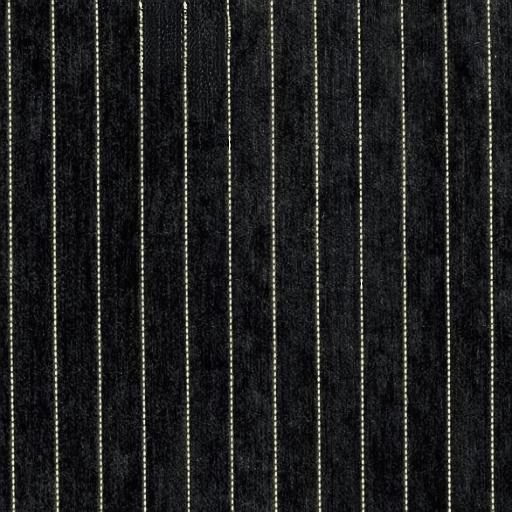} & 
\includegraphics[width=0.13\linewidth]{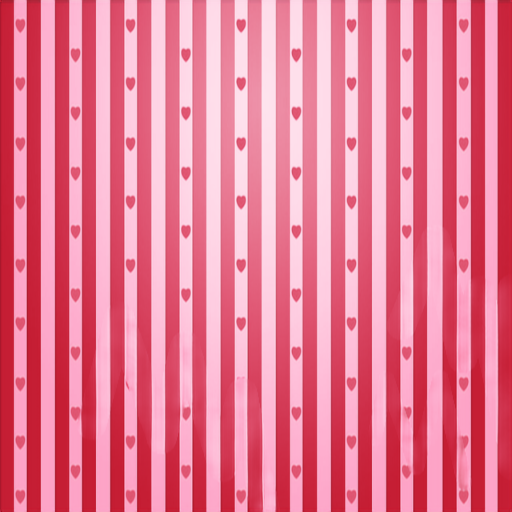}& 
\includegraphics[width=0.13\linewidth]{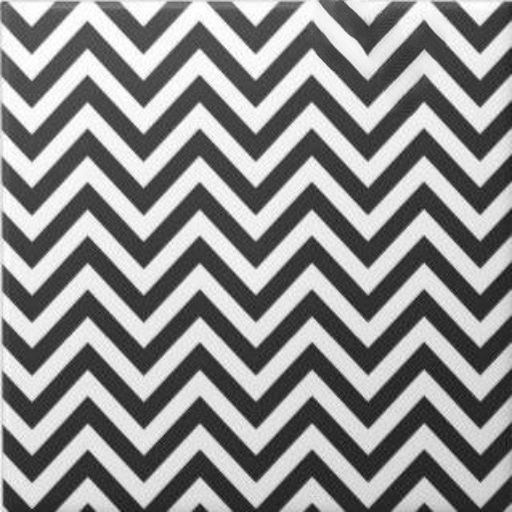}&
\includegraphics[width=0.13\linewidth]{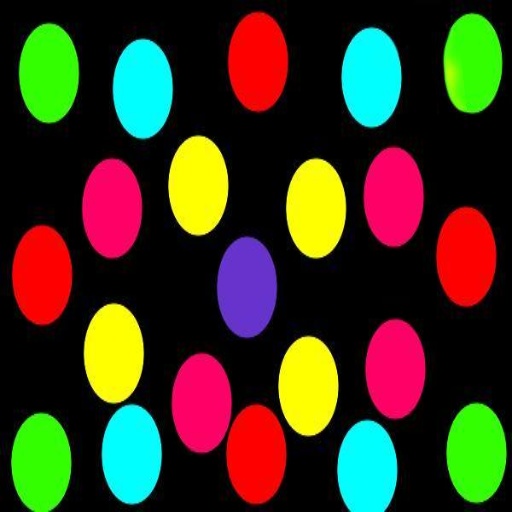}
\\  

  \rotatebox{90}{\small{ ~Input}}&
\includegraphics[width=0.13\linewidth]{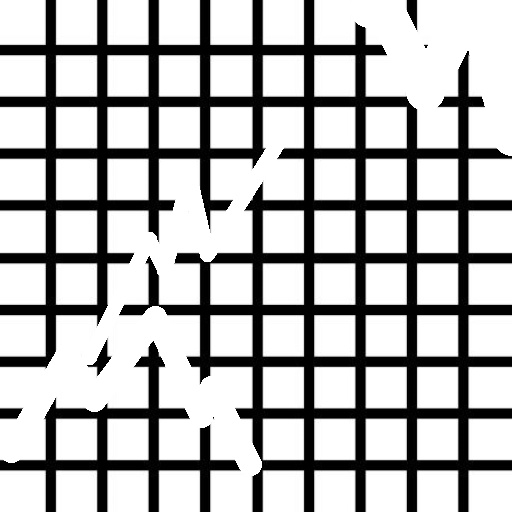}  &
\includegraphics[width=0.13\linewidth]{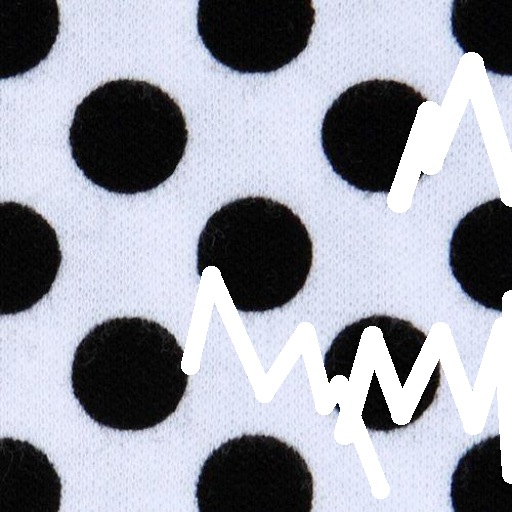}  & \includegraphics[width=0.13\linewidth]{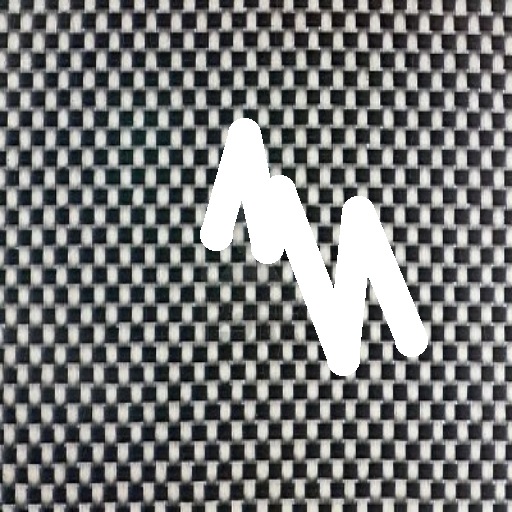}  &
\includegraphics[width=0.13\linewidth]{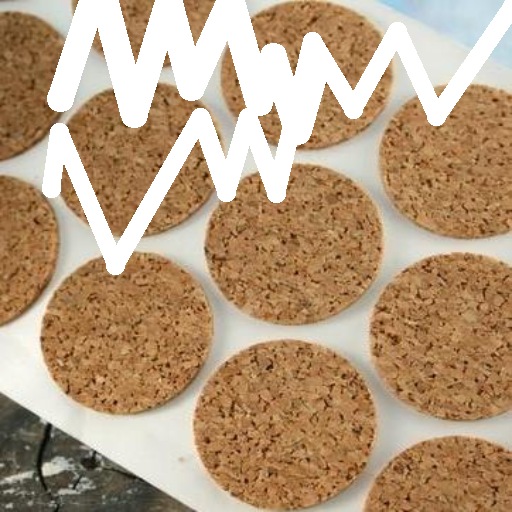}  &
\includegraphics[width=0.13\linewidth]{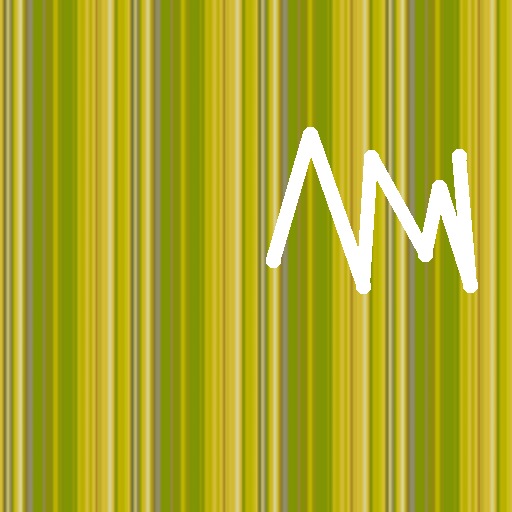}&
\includegraphics[width=0.13\linewidth]{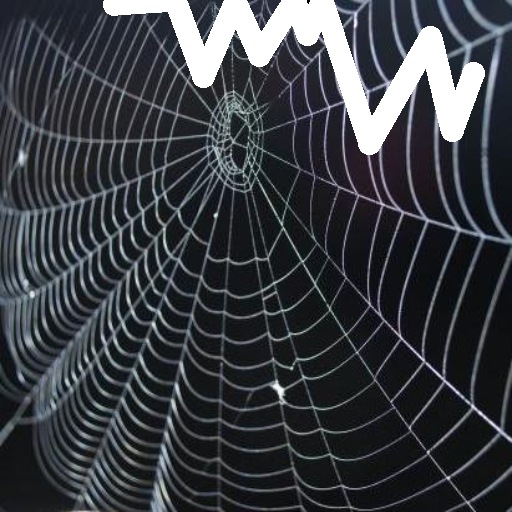}\\
    
    \rotatebox{90}{\small{HiFill~\cite{yi2020contextual}}}&
\includegraphics[width=0.13\linewidth]{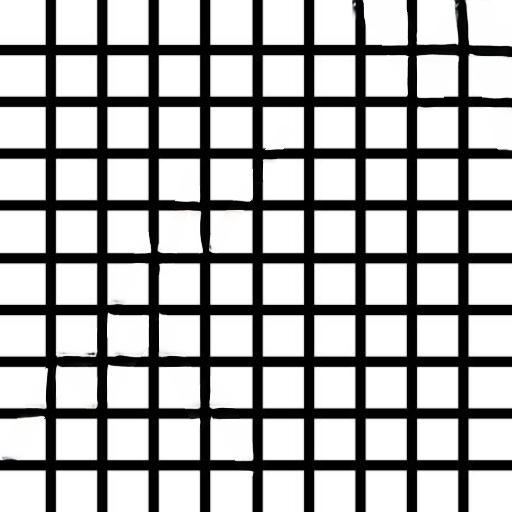} & 
\includegraphics[width=0.13\linewidth]{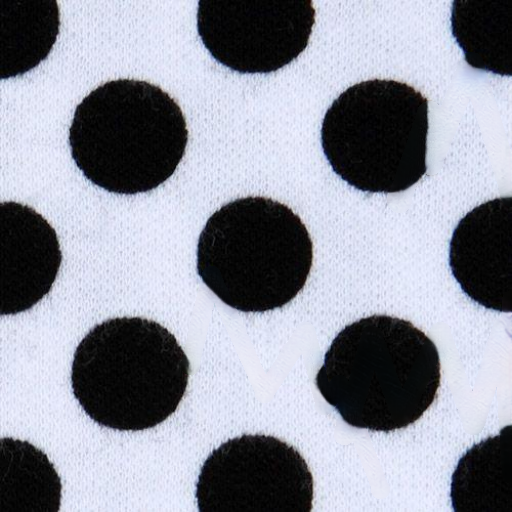} & 
\includegraphics[width=0.13\linewidth]{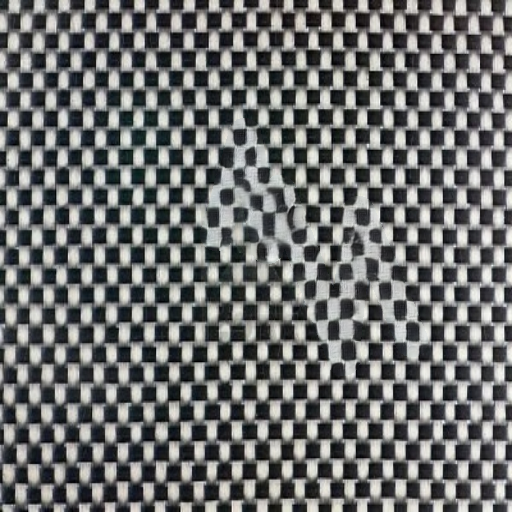} & 
\includegraphics[width=0.13\linewidth]{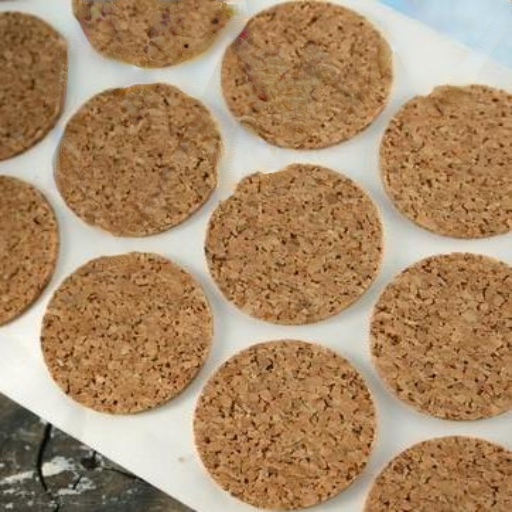} & 
\includegraphics[width=0.13\linewidth]{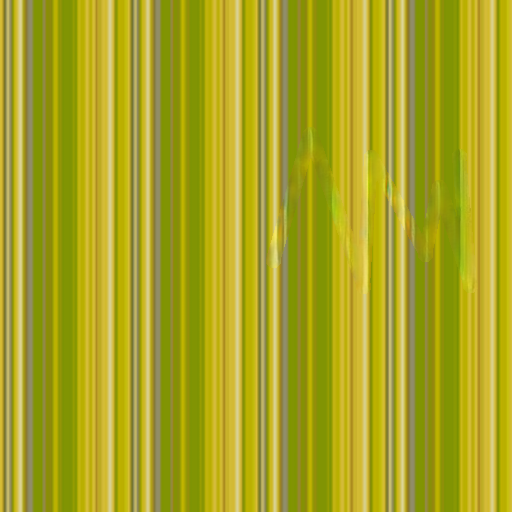} &
\includegraphics[width=0.13\linewidth]{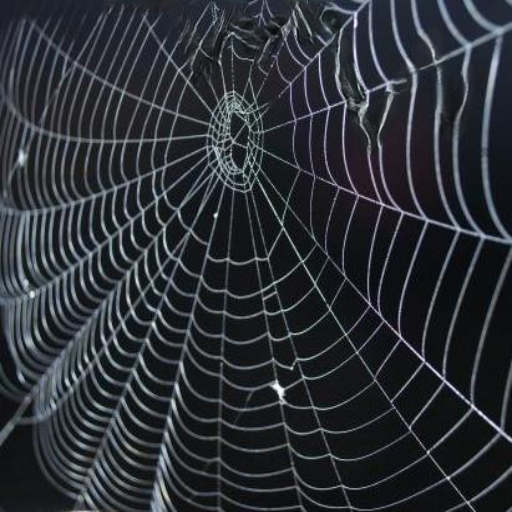}\\  

  \rotatebox{90}{ \small{ Ours }}&
\includegraphics[width=0.13\linewidth]{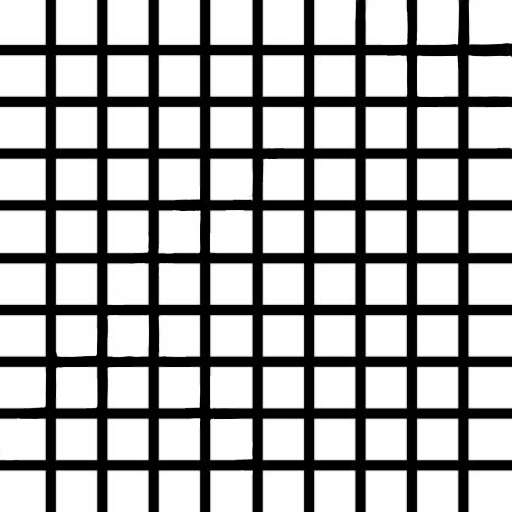} &
\includegraphics[width=0.13\linewidth]{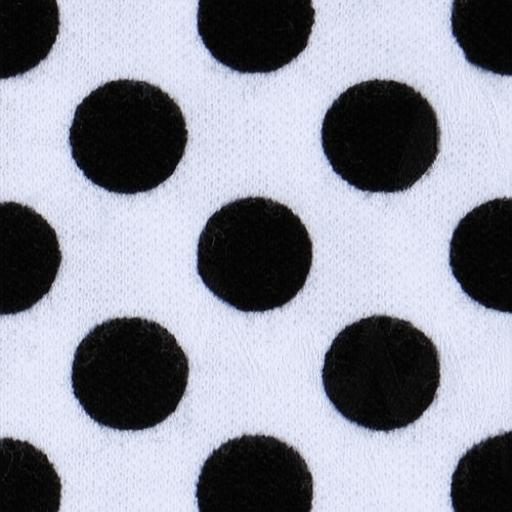} & 
\includegraphics[width=0.13\linewidth]{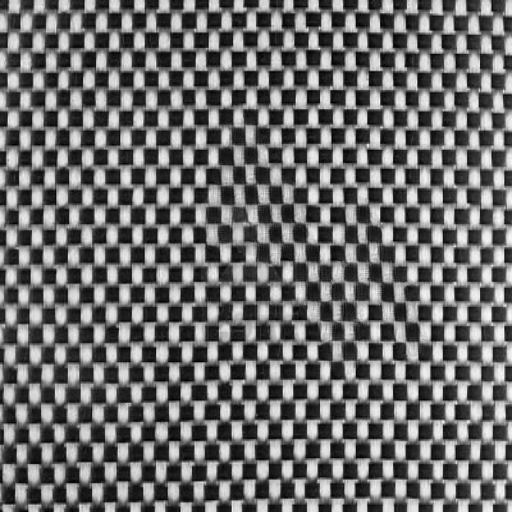} & 
\includegraphics[width=0.13\linewidth]{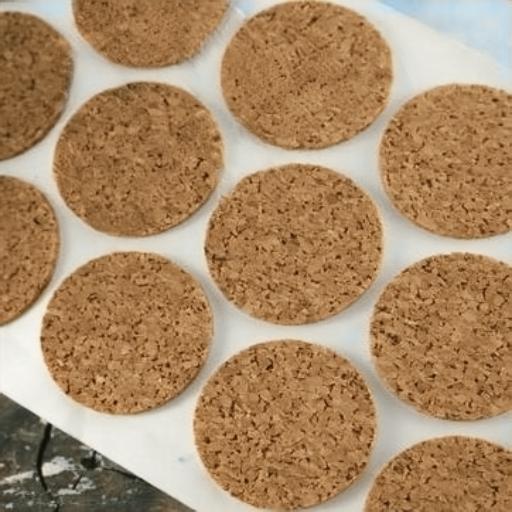}& 
\includegraphics[width=0.13\linewidth]{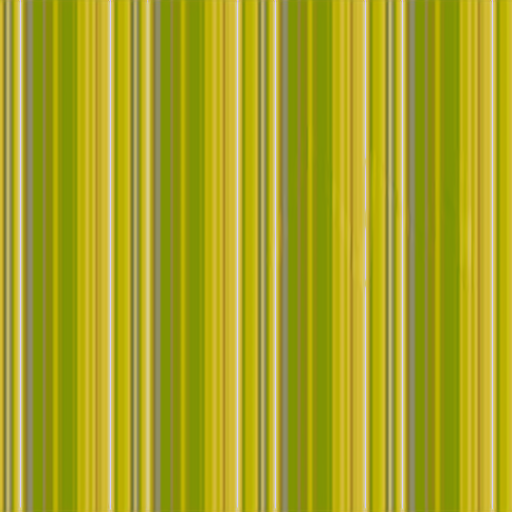}&
\includegraphics[width=0.13\linewidth]{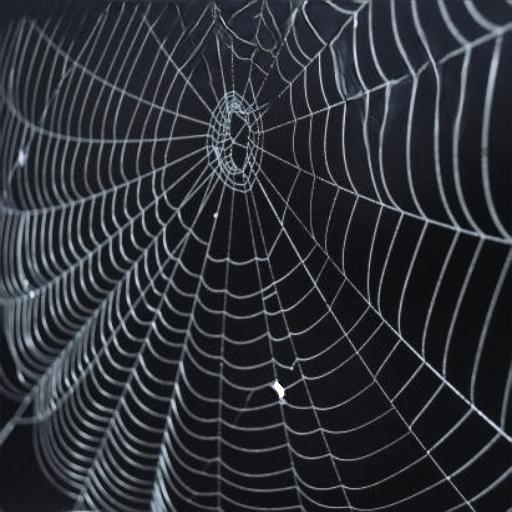}
\\  

  \rotatebox{90}{\small{ ~Input}}&
\includegraphics[width=0.13\linewidth]{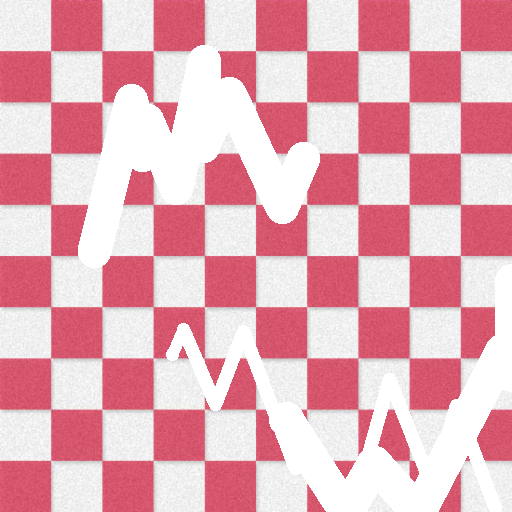}  &
\includegraphics[width=0.13\linewidth]{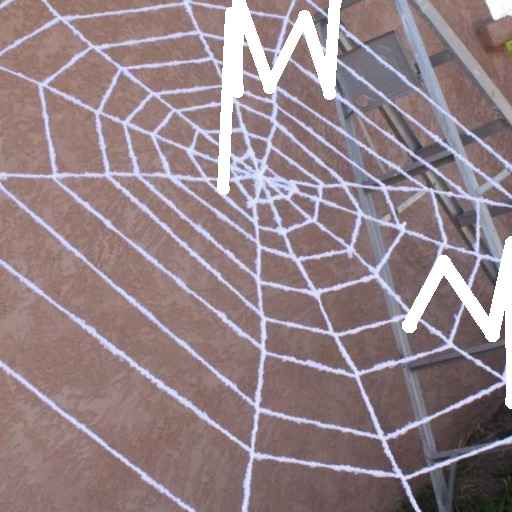}  & \includegraphics[width=0.13\linewidth]{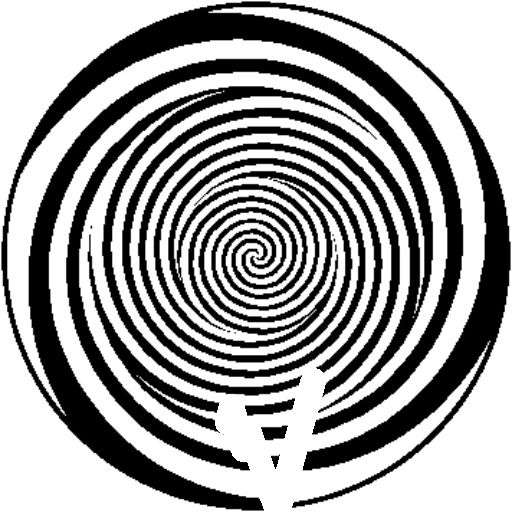}  &
\includegraphics[width=0.13\linewidth]{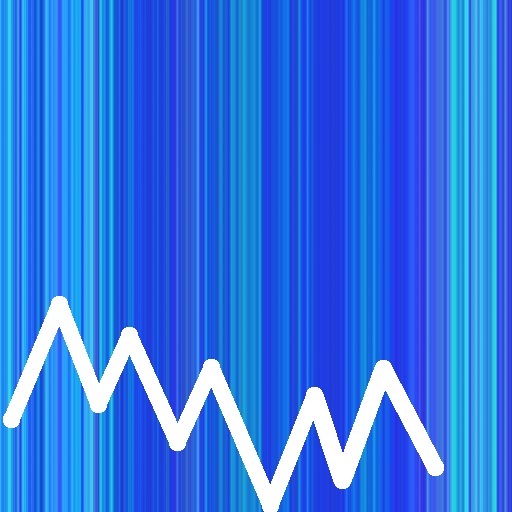}  &
\includegraphics[width=0.13\linewidth]{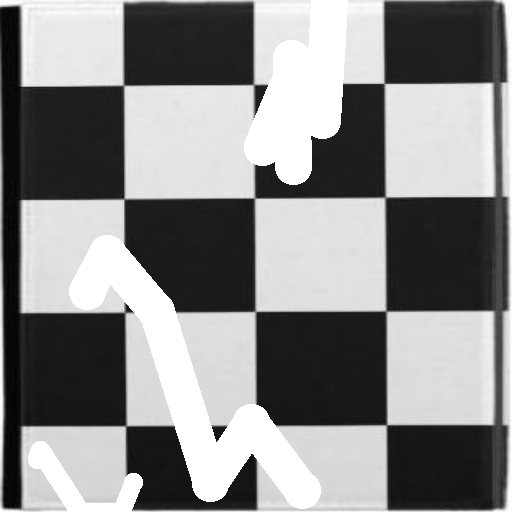}&
\includegraphics[width=0.13\linewidth]{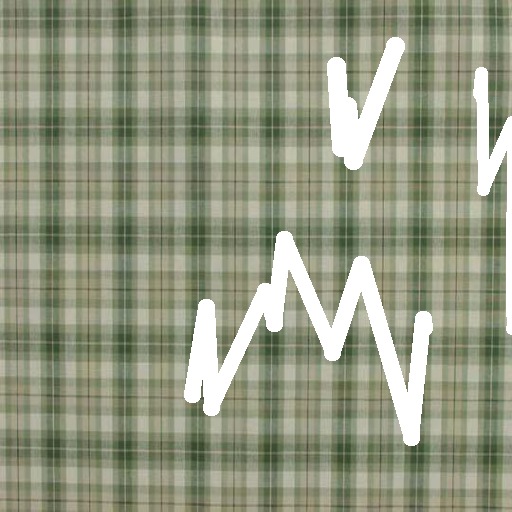}\\
    
    \rotatebox{90}{\small{HiFill~\cite{yi2020contextual}}}&
\includegraphics[width=0.13\linewidth]{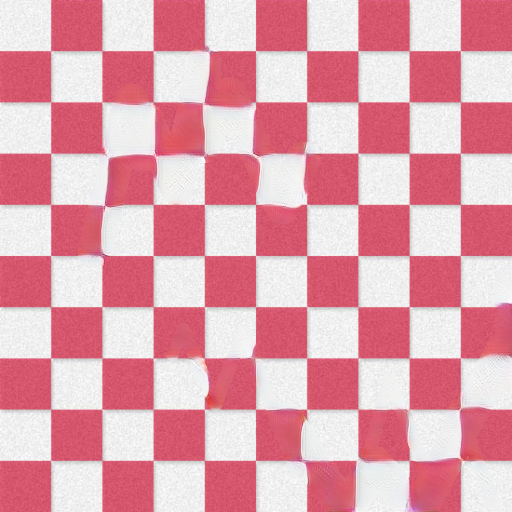} & 
\includegraphics[width=0.13\linewidth]{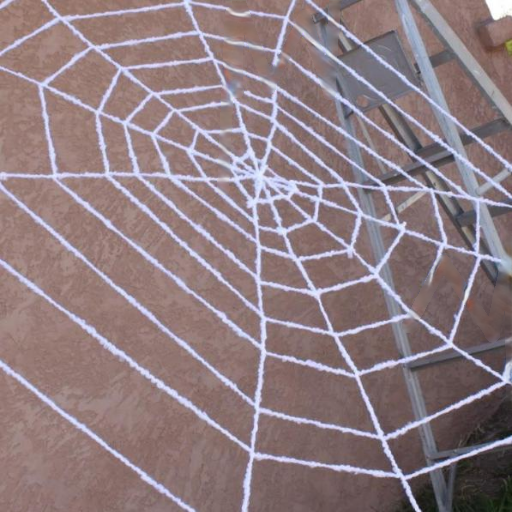} & 
\includegraphics[width=0.13\linewidth]{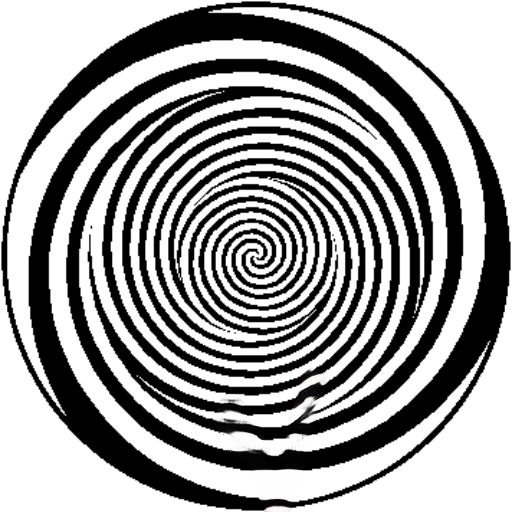} & 
\includegraphics[width=0.13\linewidth]{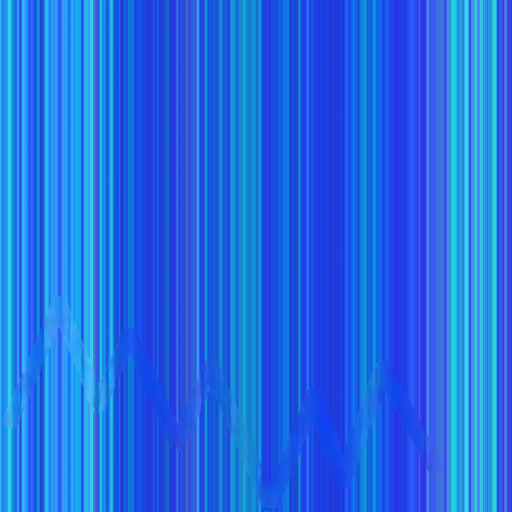} & 
\includegraphics[width=0.13\linewidth]{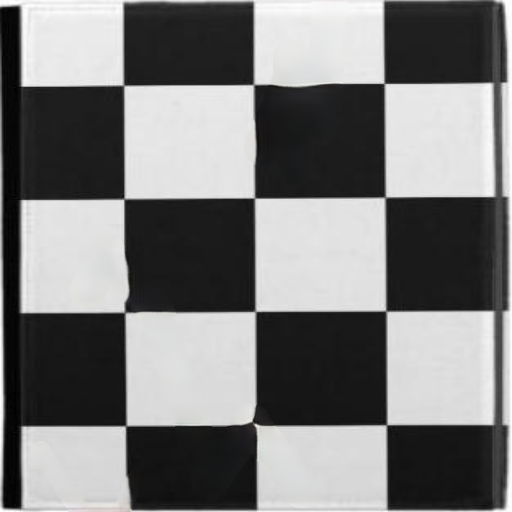} &
\includegraphics[width=0.13\linewidth]{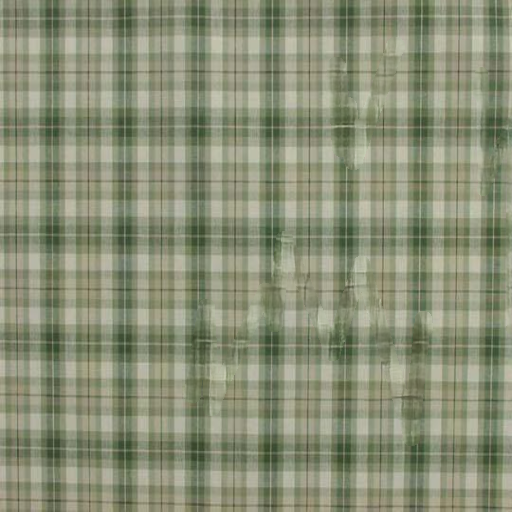}\\  

  \rotatebox{90}{ \small{ Ours }}&
\includegraphics[width=0.13\linewidth]{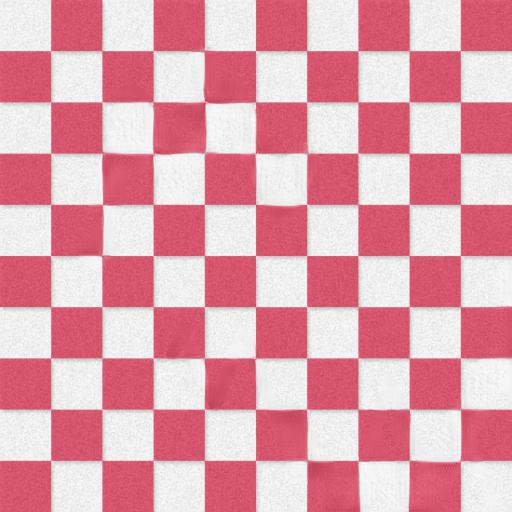} &  
\includegraphics[width=0.13\linewidth]{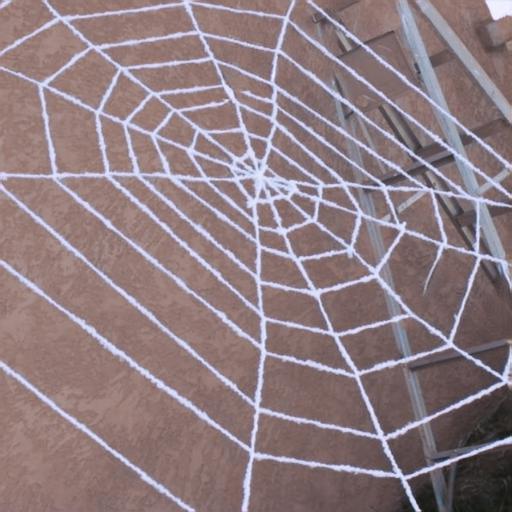} & 
\includegraphics[width=0.13\linewidth]{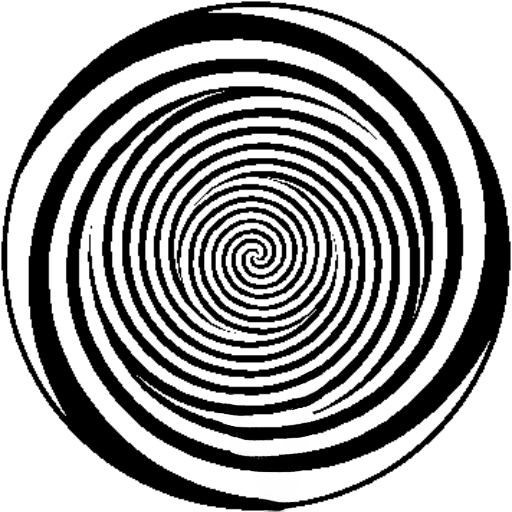} & 
\includegraphics[width=0.13\linewidth]{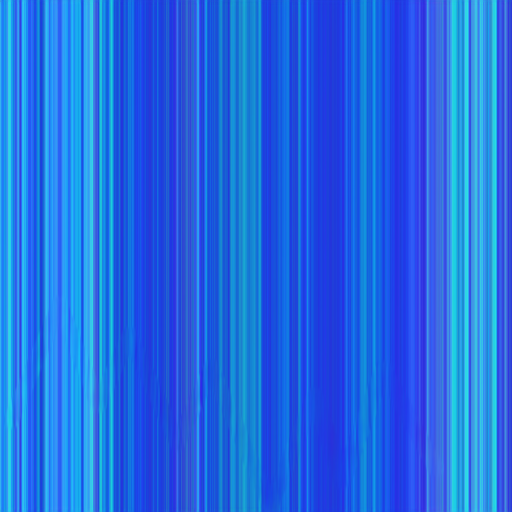}& 
\includegraphics[width=0.13\linewidth]{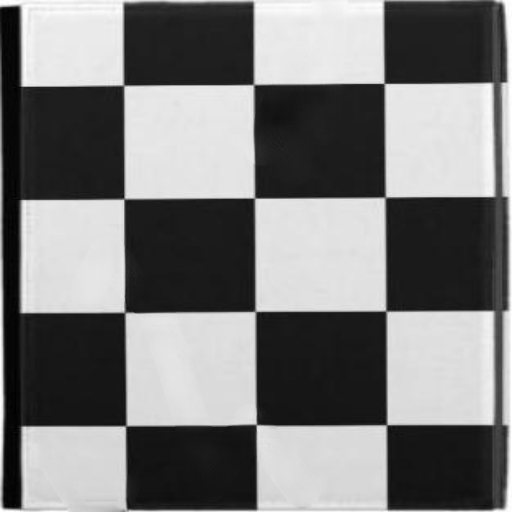}&
\includegraphics[width=0.13\linewidth]{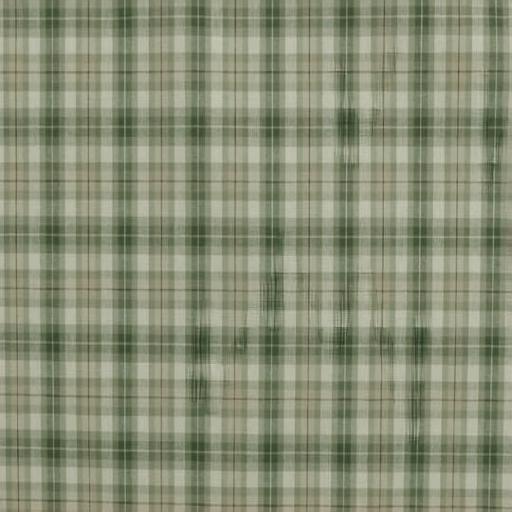}
\\   
    \end{tabular}
    \caption{Sampled results on DTD.}
    \label{fig:sample-dtd}
\end{figure*}

More visual results are shown in Fig.~\ref{fig:more}, Fig.~\ref{fig:sample-dtd}. 

\end{document}